%% file: thesis.tex
\documentclass{book}

\newcommand*{\noaddvspace}{\renewcommand*{\addvspace}[1]{}}
\addtocontents{lof}{\protect\noaddvspace}
\usepackage{amstext,amssymb,amsfonts,latexsym}
\usepackage{amsthm, amsmath, amssymb} 
\usepackage[ruled,vlined]{algorithm2e}
\usepackage[utf8]{inputenc}
\usepackage{style/bib}
\usepackage[UKenglish]{babel}
\usepackage{csquotes}
\usepackage{ccicons}
\usepackage{animate}
\usepackage{media9}
\usepackage{setspace}
\usepackage{todonotes}
\usepackage{pdfpages}
\usepackage{hyperref}
\usepackage{multirow}
\usepackage{lscape}
\usepackage{titlesec}
\usepackage{svg}
\usepackage{subcaption}
\usepackage{booktabs}

\hypersetup{
    pdfauthor = {Victor Zuanazzi},
    pdftitle = {Do not trust the neighbors! Adversarial Metric Learning for Self-Supervised Scene Flow Estimation},
    colorlinks,
    linkcolor={black},
    citecolor={black}
    }

\SetKwInput{KwInput}{Input}                
\SetKwInput{KwOutput}{Output}              

\DeclareMathOperator*{\tm}{\mathsf{tm}}
\DeclareMathOperator*{\mst}{\mathsf{mst}}

\DeclareMathOperator*{\ml2}{\mathsf{mL2}}

\DeclareMathOperator*{\cc}{\mathsf{cc}}

\DeclareMathOperator*{\KNN}{\mathsf{KNN}}
\DeclareMathOperator*{\knn}{\mathsf{knn}}

\usepackage[numbers,sort]{natbib}
\usepackage{microtype}

\usepackage{graphicx, color}
\usepackage[a4paper,margin=2cm]{geometry}

\newenvironment{itquote} 
{\begin{quote}\itshape}
{\end{quote}}

\begin{document}

\begin{titlepage}

\newcommand{\HRule}{\rule{\linewidth}{0.5mm}} 
\center 



\textsc{\Large MSc Artificial Intelligence}\\[0.2cm]

\textsc{\Large Master Thesis}\\[0.5cm] 


\HRule \\[0.4cm]

{\huge \bfseries Do not trust the neighbors! \\ Adversarial Metric Learning for Self-Supervised Scene Flow Estimation}\\[0.4cm] 

\HRule \\[0.5cm]


by\\[0.2cm]

\textsc{\Large Victor Zuanazzi}\\[0.2cm] 

12325724\\[1cm]


{\Large \today}\\[1cm] 

48 Credits\\ %

1st Semester 2020\\[1cm]%


\begin{minipage}[t]{0.4\textwidth}

\begin{flushleft} \large

\emph{Supervisor:} \\
MSc. \textsc{Joris van Vugt} \\
Dr. \textsc{Olaf Booij} \\
Dr. \textsc{Pascal Mettes} \\ 

\end{flushleft}

\end{minipage}

~

\begin{minipage}[t]{0.4\textwidth}

\begin{flushright} \large

\emph{Assessor:} \\

Prof. Dr. \textsc{Cees Snoek}\\

\end{flushright}

\end{minipage}\\[2cm]


\includegraphics[width=2.5cm]{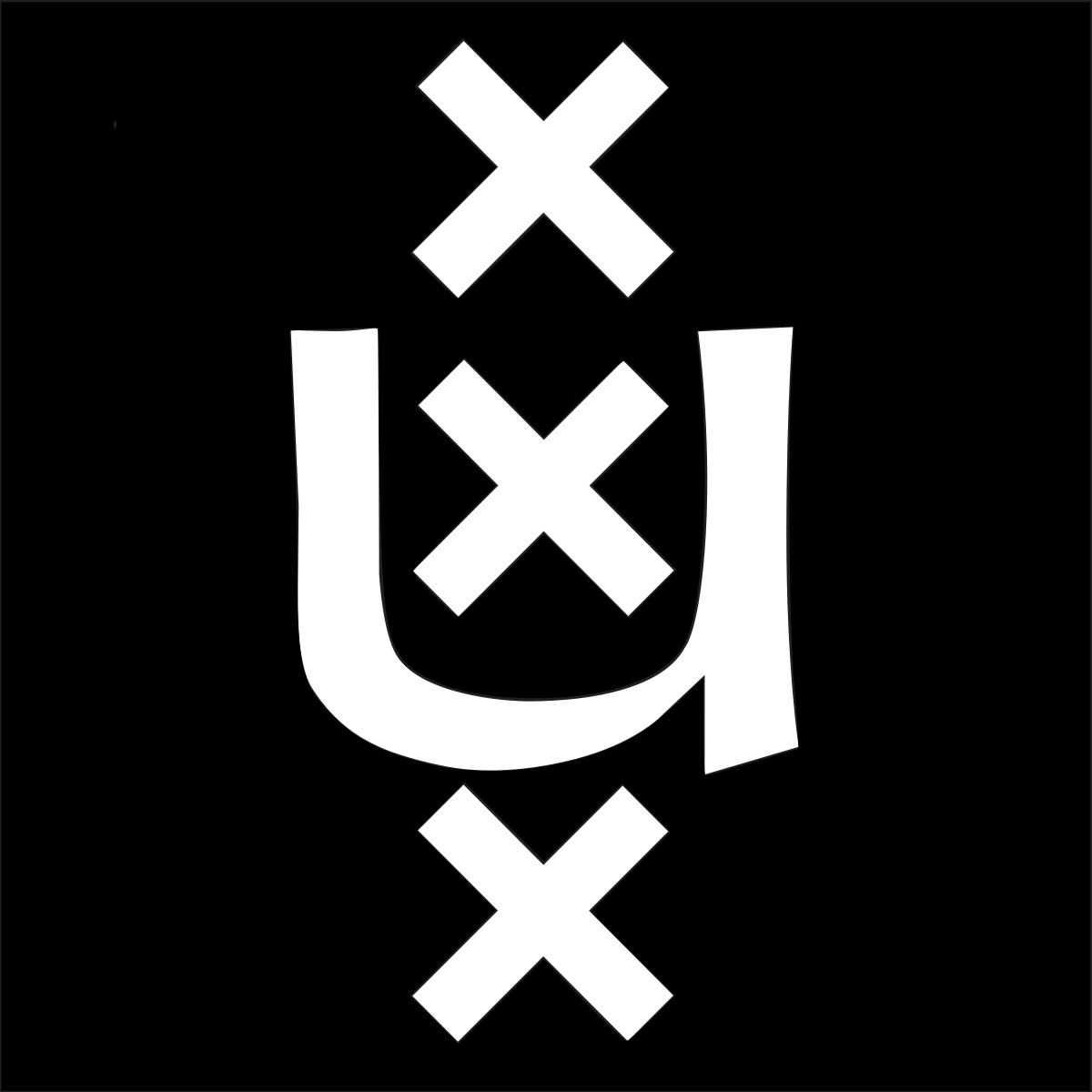}\\ 

\textsc{\large Informatics Institute}\\[1.0cm] %

\vfill 

\end{titlepage}


\renewcommand{\thepage}{\roman{page}}	

\setstretch{1.5}

\input{src/abstract.tex}
\input{src/thanks.tex}

\setcounter{tocdepth}{2}
\newpage
\tableofcontents

\newpage
\setcounter{page}{1}
\renewcommand{\thepage}{\arabic{page}}	


\input{src/introduction.tex}
\input{src/litreview.tex}
\input{src/mechanism.tex}
\input{src/methods.tex}
\input{src/dataset}
\input{src/results2.tex}

\input{src/conclusion.tex}


\addcontentsline{toc}{chapter}{Bibliography}
\bibliography{thesis.bib}

\addcontentsline{toc}{chapter}{List of Figures}
\listoffigures
\addcontentsline{toc}{chapter}{List of Tables}
\listoftables
\addcontentsline{toc}{chapter}{List of Algorithms}
\listofalgorithms

\appendix
\input{src/appendix.tex}

\end{document}

%% file: src/abstract.tex

\newpage
\chapter*{Abstract}

Scene flow is the task of estimating 3D motion vectors to individual points of a dynamic 3D scene. Motion vectors have shown to be beneficial for downstream tasks such as action classification and collision avoidance. However, data collected via LiDAR sensors and stereo cameras are computation and labor intensive to precisely annotate for scene flow. We address this annotation bottleneck on two ends. We propose a 3D scene flow benchmark and a novel self-supervised setup for training flow models. The benchmark consists of datasets designed to study individual aspects of flow estimation in progressive order of complexity, from a single object in motion to real-world scenes. 
Furthermore, we introduce Adversarial Metric Learning for self-supervised flow estimation. The flow model is fed with sequences of point clouds to perform flow estimation. A second model learns a latent metric to distinguish between the points translated by the flow estimations and the target point cloud. This latent metric is learned via a Multi-Scale Triplet loss, which uses intermediary feature vectors for the loss calculation. We use our proposed benchmark to draw insights about the performance of the baselines and of different models when trained using our setup. We find that our setup is able to keep motion coherence and preserve local geometries, which many self-supervised baselines fail to grasp. Dealing with occlusions, on the other hand, is still an open challenge.

%% file: src/thanks.tex

\chapter*{Acknowledgements}



I would like to express my deepest appreciation to  Joris van Vugt and Olaf Booij, my daily supervisors who definitely earned the title. Their experience, enthusiasm and guidance were constant throughout this work. I would also like to extend my deepest gratitude to my UvA supervisor, Pascal Mettes, who kept his promise on pushing for high academic standards. His out-of-the-box thinking and perspicacity were invaluable.

This thesis is a product of the collaboration between TomTom and the Universiteit van Amsterdam (UvA). I would like to thank both entities for investing in working together. In particular, I would like to thank my manager Michael Hofmann for the opportunity and his inputs in this work. On the UvA's side, I am grateful to Cees Snoek for welcoming and encoraging this type of colaboration.

In my time in TomTom, I had great pleasure in working with the Sugargliders, Alexander Korvyakov, Deyvid Kochanov, Fatimeh Karimi Nejadasl, Marius-Cosmin Clucerescu, Nicolau Leal Werneck, Prabu Dheenathayalan, and Ysbrand Galama. A team of friendly and knowledgeable people. It was also a pleasure collaborating with my fellow researchers David Maximilian Biertimpel, Elias Kassapis, Erik Stammes, and Melika Ayoughi. I would like to extend my gratitude to the whole Autonomous Driving team, who are too many to list here, for their interest, patience, and \textit{gezelligheid}. Special thanks go to Cedric Nugteren for facilitating the access to the servers where all experiments were performed.

Special thanks to Stijn Verdenius for his fresh perspective and feedback. Many thanks to Ana Laura V. Z. de Abreu, Hans Vanacker, Maria Luiza V. Z. de Abreu, and Mona Poulsen for their constant presence. I cannot begin to express my gratitude to Francisco J. de Abreu and Renata A. V. Z. de Abreu who believed in me before I believed in myself. Their presence, support, and love cannot be put into words. Finally, I am extremely grateful to Tess Vanacker, who supported me in so many ways it is not reasonable to enumerate nor fair to summarize. Het wonderlijke dwaallicht dat me naar dit continent lokte en me met warm enthousiasme begeleidde op het pad naar deze scriptie. 

Thank you. Dank u wel. Muitíssimo obrigado.


%% file: src/introduction.tex
\chapter{Introduction}

From cellphones with stereo cameras to cars equipped with LiDAR and Radar sensors, 3D data has become increasingly present in our society. This type of data has received attention from industry and academia in the research and development of several applications. One particularly evident example being self-driving features for the automotive industry. Whereas depth and motion information are equally important e.g. for collision avoidance features, most sensors have no means to collect the latter. This leaves us the task of estimating a 3D motion field. Such low-level understanding of a dynamic environment is called \emph{scene flow}.

Many applications could profit from having scene flow as an auxiliary input. Promising results have been shown for online and offline tasks such as semantic segmentation, object detection, and tracking \cite{Behl_2017_ICCV, Vogel_2013_ICCV, flowsegmentation}. The low level features  required to perform flow estimation may also be relevant to perform other tasks. The work of \cite{PointFlowNet} shows the benefit of jointly learning scene flow, rigid body motion, and 3D object detection. In map-making, scene flow is a powerful tool to filter out dynamic objects from a LiDAR scan \cite{FlowNet3D}. The increasing presence of machine learning models in society is also pushing companies to develop models with interpretable results. Scene flow has a very intuitive movement interpretation that can aid work on action classification, trajectory prediction, and 3D reconstruction \cite{FlowNet3D++, flow3dreconstruction, flow3dreconstruction2, actionflow}. 

\begin{figure}
    \centering
    \includegraphics[width = .3\linewidth]{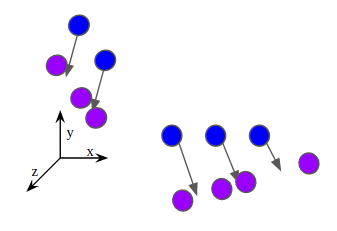}
    \caption{Illustration of the scene flow task. Blue points belong to a snapshot at time $t$ and purple points belong to a snapshot at time $t+1$. It is possible that a different number of points is recorded at different time steps and that there is no one-to-one correspondence between points of different frames. The task of scene flow is to estimate motion vectors for each point of a scene.}
    \label{img:sceneflow}
\end{figure}

Scene flow has gained an increasing interest in the research community \cite{FlowNet3D, FlowNet3D++, SelfSupervisedFlow, HPLFlowNet, PointPWCNet, LidarFlow}. In short, the task consists of estimating 3D motion vectors -- a flow field -- for every point in a frame given a sequence of frames, as illustrated by Figure~\ref{img:sceneflow}. It does not assume any knowledge of structure or motion of a scene. Quoting the definition from \cite{flyingthings3d}:

\begin{itquote}
    Estimating scene flow means providing the depth and 3D motion vectors of all visible points in a stereo video [a scene]. It is the “royal league” task when it comes to reconstruction and motion estimation and provides an important basis for numerous higher-level challenges such as advanced driver assistance and autonomous systems.
\end{itquote}

Researchers and engineers aiming to build applications that make use of a flow field are faced with a chicken-egg type of problem. Motion data is not readily available from sensors, so it must be estimated. However, techniques yielding sufficiently accurate estimations are still to be developed. At the time of writing and to the best of our knowledge, KITTI Scene Flow \cite{kitti} is the only dataset of real data containing flow annotations. Only 200 scenes were semi-automatically annotated, far too few for training large, sophisticated models. The computational cost and human labor necessary makes it practically infeasible to create a large scale dataset of real data with annotated flow targets.

Synthetic data has been used to fuel most of the development of scene flow so far \cite{FlowNet3D, FlowNet3D++, HPLFlowNet, PointPWCNet}. A common approach is to perform supervised training on a synthetic dataset such as FlyingThings3D \cite{flyingthings3d} followed by finetuning on a portion of KITTI Scene Flow \cite{kitti}. However, there are large non-annotated datasets of real data \cite{lyft2019, Pitropov2020CanadianAD, nuscenes2019} that have received seldom attention for the task of scene flow. Self-supervised training can be of great value for the scientific community, as well as for many applications that profit from scene flow estimation. 

The works of \cite{PointPWCNet, SelfSupervisedFlow} propose replacing the flow supervision with spatial and geometric self-supervision. As a proxy of motion, nearest neighbor based distances are used. The training is aided by auxiliary losses and regularization terms aiming to enforce geometric consistency. However, the aperture problem \cite{aperture} of computer vision makes nearest neighbor based distances rather unreliable for supervising scene flow estimation. We investigate and reflect on the effectiveness of their approaches. 

Whereas supervised scene flow estimation has examples of well-performing models in different benchmarks \cite{PointFlowNet, FlowNet3D, FlowNet3D++}, self-supervised scene flow has been less explored. The insights of this work may help the scientific community to close the performance gap between supervised and self-supervised flow estimation. The main contributions of this thesis are:

\begin{enumerate}
    \item We found that it is not fair to compare the results reported by \cite{FlowNet3D, FlowNet3D++, HPLFlowNet, PointPWCNet}. The authors worked under two different assumptions regarding the nature of data sampling. One of which makes the task of scene flow estimation artificially simpler than the other. We coined the \emph{Correspondence Mechanism} and the \emph{Re-sampling Mechanism} to distinguish between two relevant sets of assumptions on the data gathering. We show the fundamental differences between the two and argue the latter is most representative of data gathered by LiDAR and stereo cameras. 
    \item We propose a novel training setup -- \emph{Adversarial Metric Learning} -- for self-supervised scene flow estimation. A flow model is fed with sequences of point clouds to perform flow estimation. A second model learns a latent metric to distinguish between the points translated by the flow estimations and the target point cloud. This latent metric aims to replace nearest neighbor based distances. It is learned via a \emph{Multi-Scale Triplet loss}, which uses intermediary feature vectors for the loss calculation. 
    \item We propose a 3D scene flow benchmark, the \emph{Scene Flow Sandbox}. It consists of datasets designed to study individual aspects of flow estimation in progressive order of complexity, from a single object in motion to real-world scenes. We make extensive use of the benchmark to intuitively explain the failure modes of different models and methods.
\end{enumerate} 


\section{Problem Statement}
\label{sec:problem_statement}

The concept of flow field has its origins in field theory and fluid dynamics. It is defined as the distribution of density and velocity of fluid over space and time \cite{fluiddynamics}. In computer vision, a flow field has an analogous meaning. It may be defined as the velocity distribution of a recorded scene over space and time. Optical flow is possibly the most studied and well-established discipline for estimating a flow field. Given a sequence of frames from a video, the objective is to compute a flow field mapping each pixel at frame $t$ to its position in the image grid in the frame $t + 1$. We call it grid-based flow, \cite{optcflowdef} defines it as:

\begin{itquote}
    Optical flow is defined as the apparent motion of individual pixels on the image plane. It often serves as a good approximation of the true physical motion projected onto the image plane.
\end{itquote}

Flow estimation can also be performed in 3D data. It is called \emph{scene flow} and was first introduced by \cite{EarlySceneFlow}. The aim is to map each point from frame $t$ to its position in the frame $t+1$. In real applications that translates to estimating a velocity vector for each point. Figure~\ref{img:sceneflow} illustrates the task. Scene flow vectors have a real-world meaning, they are 3D velocity vectors. For a generic application, it is sensible to define the flow vectors in meters per frame. Most commercial sensors use a constant sampling rate, the conversion to meters per second, $\left[ \frac{m}{s} \right]$, is trivial:
\begin{equation}
    \vec{f} \left[ \frac{m}{s} \right] = \vec{f} \left[\frac{m}{\text{frame}} \right] \gamma \left[\frac{\text{frame}}{s} \right],
\end{equation}
where $\gamma$ is the sampling rate and $\vec{f}$ is the flow vector. 

We would like to stress that in optical flow the flow vectors are measured in pixels per frame. An object that is one meter distant from a camera and moves one meter to the side will translate a number of pixels. The same object ten meters further from the camera moving one meter to the side will translate a tenth of the number of pixels. Even though the motion in the 3D space was the same, the motion in the pixel space was not. Furthermore, optical flow vectors cannot be translated into 3D flow vectors without a depth map and camera-specific settings such as the focal length and pixel size. 


A sequence of measurements collected via a sensor is called a scene. For instance, raw LiDAR scans (or sweeps) contain probe id, azimuth angle, and depth (most sensors also record the reflectively of the surface). Stereo Cameras record a disparity map. For academic purposes, this work assumes a scene, independent of the sensor, can be converted to rectangular coordinates (XYZ) without loss of 3D spatial information on every recorded point. We focus on scenes made of frames of point clouds. 

We introduce the mathematical notation used in this work. Conceptually, point clouds are unordered sets of points. In other words, the points are not sorted in any particular order. However, for arithmetic convenience we treat point clouds as matrices, $C \in \mathbb{R}^{N \times 3}$ with $\vec{p}_i$ being the point in the $ith$ row of $C$. The matrix $F \in \mathbb{R}^{N \times 3}$ stores the corresponding flow vectors. The point $\vec{p}_i$ has the corresponding flow vector $\vec{f}_i$, such that:
\begin{equation}
    \hat{C} = C + F,
    \label{eq:hatc}
\end{equation}
is the translated point cloud $C$. It is assumed that $C_t$ is an instant scan of the scene at time $t$ and it is followed by $C_{t+1}$. In other words, time is quantized by the frame rate. We may also refer to $C_1$ and $C_2$ as to make explicit that we are dealing with a dynamic scene with only two consecutive frames. It is worth noting that $C$ and any random permutation of its points are equally representative of a scene. If $C$ is shuffled, then $F$ has to be shuffled in the same way so to keep Equation~\ref{eq:hatc} valid.

To assume the data is in point cloud format does not mean to disregard how the data is collected. If the data is gathered in such a way that $C_{t+1}= C_t + F_t$  we name it the Correspondence Mechanism. Else, the sensor performs a re-sampling of measurements at every time step, which we call the Re-sampling Mechanism. For the latter $C_{t+1}$ has no correspondent point at $C_t$. Whereas the first assumes fully observable scenes, the latter resembles data from LiDAR sensors the closest. The concepts are explained in further detail in Section~\ref{subsec:corrvssamp}.  

In this section, we introduced the concept of scene flow. The task of optical flow is popular in the computer vision community, we showed the parallels and differences between the two. We then introduced relevant concepts for understanding the task. This work focuses on scene flow performed on point clouds collected via the Re-sampling mechanism.


%% file: src/litreview.tex
multi21(1)
\chapter{Literature review}
\label{sec:litreivew}

In this section, we briefly describe the relevant research on scene flow estimation on point clouds. We start with traditional methods and move towards the state-of-the-art deep learning approaches. Each one has a different backbone for consuming 3D data. Lastly we review self-supervised setups for learning scene flow estimation. In Sections \ref{sec:relatedmech}, \ref{sec:relatedtarget} and \ref{sec:modelchoices} we make explicit how we differentiate from previous work.

The task of Scene Flow was introduced to the scientific community by \cite{EarlySceneFlow} on stereo videos. They proposed a variational approach to the task, inspired by the work of \cite{HornShunckFlow} on optical flow. For optical flow, the use of convolutions was an obvious choice and DeepFlow \cite{deepflow} was the first deep learning model applied to the task. In contrast, the choice of architecture for consuming 3D point clouds and outputting flow vectors is less obvious. We describe three particularly relevant methods. 

Deep neural network architectures were only proposed to tackle scene flow on point clouds after the introduction of methods for extracting features from point clouds. PointNet \cite{PointNet} was the first architecture to extract point-wise and global features from point clouds. The follow-up work PointNet++ \cite{PointNet++} introduced hierarchical feature extraction layers. PointConv \cite{pointconv} uses the density of the points in a point cloud to weight a discrete 3D convolution operation that approximates 3D continuous convolutions. PointConv can be seen as the point cloud version of convolutions in images. SplatNet \cite{splatnet} proposes projecting the 3D point cloud into latices \cite{splatnet}. The convolutions performed on 2D lattices are faster to compute and less memory intensive than in 3D point clouds. Those architectures serve as backbones for the following flow models. 

The first work to use deep learning techniques to perform scene flow estimation was \cite{FlowNet3D}. The authors used the building blocks of \cite{PointNet++} to design an architecture called FlowNet3D that consumes two point clouds and estimates the flow vectors. To show its potential, the authors train the model usilg FlyingThings3D \cite{flyingthings3d} and report results on KITTI Scene Flow \cite{kitti}. The follow-up work of \cite{FlowNet3D++} show the benefit of using point-to-plane, and cosine distance as auxiliary geometric losses.

PointPWC-net \cite{PointPWCNet} is inspired by PWC-Net \cite{PWCflow}. PWC-net \cite{PWCflow} is an architecture that uses three components for optical flow estimation: coarse to fine pyramids, warping layers, and cost volumes. PointPWC-net \cite{PointPWCNet} uses PointConv \cite{pointconv} as backbone for feature extraction on point clouds. In PointPWC-net the coarse pyramids are subsamples of the point clouds using furthest point sampling. The warping layer performs point-wise translation. The cost volume was designed to take temporal information into account. The authors report results on KITTI Scene Flow \cite{kitti} after supervised training on FlyingThing3D \cite{flyingthings3d}.

The authors of \cite{HPLFlowNet} propose HPLFlowNet that makes use of 2D permutohedral lattices for performing 3D scene flow estimation on large point clouds. Whereas the two aforementioned models become resource intensive when used with large point clouds, HPLFlowNet handles them in constant time. They report results using 50k points for KITTI Scene Flow \cite{kitti} scenes without the need for scaling up the hardware. 

The three aforementioned flow models were built on top of different operations for extracting features from point clouds. Their work focuses primary on supervised setups for learning flow, the end-point-error is minimized directly via L2 loss. The models are trained on FlyingThings3D \cite{flyingthings3d}, finetuned, and tested on KITTI Scene Flow \cite{kitti}. Yet, self-supervised setups have also received attention from the scientific community. 

The work of \cite{SelfSupervisedFlow} uses the FlowNet3D \cite{FlowNet3D} architecture to perform self-supervised flow estimation. The authors proposes to combine nearest neighbor loss and cycle consistency loss to train the network on. The flow model is trained in three steps. First, it is trained with supervision on FlyingThings3D \cite{flyingthings3d}, then trained on NuScenenes and Lyft \cite{nuscenes2019, lyft2019} using the self-supervised losses, and further finetuned - with supervision - on KITTI Scene Flow \cite{kitti}. 

The authors of PointPWC-net \cite{PointPWCNet} propose a self-supervised setup alongside the architecture. They make use of the chamfer distance between point clouds as the main loss. Two auxiliary losses are also used. They make use of a local flow consistency penalty. In a local neighborhood, we expect the flow vectors to be similar, thus differences are penalized. Lastly, authors make use of a Laplacian loss, the points present in a local neighborhood are used to approximate a local plane's normal vector, the L2 distance between the normals plane of $pc_2$ and $\hat{pc}_2$ is used. 

The work of \cite{SelfSupervisedFlow, PointPWCNet} replaced the flow supervision with spatial and geometric self-supervision. They use nearest-neighbor based distances to approximate motion. For keeping geometric coherence the authors use regularization terms aimed to minimize distortions caused by locally inconsistent flow vectors.

The relevant literature was described in this section. We listed three different approaches to supervised flow \cite{FlowNet3D, PointPWCNet, HPLFlowNet}, each uses a distinct method for extracting features from point clouds \cite{PointNet++, pointconv, splatnet}. With this foundation in place, we moved to explain how self-supervision has been tackled by \cite{SelfSupervisedFlow, PointPWCNet}. Both methods use proxies for spatio-temporal and geometric elements. The following sections describe the relationship the literature has with our work. The argumentation is weighted by concepts and experiments introduced in later chapters.


\section{Assuming the Correspondence Mechanism}
\label{sec:relatedmech}

In the process of searching for a self-supervised approach, we reproduced their results and performed experiments with numerous models and setups \cite{FlowNet3D, PointPWCNet, HPLFlowNet}. We noticed that not all authors had the same set of assumptions regarding the data used in their experiments. To fairly compare the different methods, we defined the Correspondence and the Re-sampling Mechanisms. Those are explained in detail in Chapter~\ref{subsec:corrvssamp} and their impact on the scene flow task is shown in Section~\ref{subsec:assumption}.

For now, it is sufficient to know the Re-sampling Mechanism is the most representative of data gathered by sensors such as LiDARs and stereo cameras. The sensor re-samples measurements at every time step. Thus $C_{t+1}$ has no correspondent point at $C_t$. On the other hand, the Correspondence Mechanism is most representative of data gathered by sensors that perform tracking, such as GPS. The data is gathered in such a way that points in $C_t$ have immediate correspondents in $C_{t+1}$. Mathematically this means: $C_{t+1}= C_t + F_t$. 

The works of \cite{HPLFlowNet, PointPWCNet} do not make explicit their assumed mechanism. To the best of our understanding, they assume the Correspondence Mechanism. We draw this conclusion from the publicly available code-bases \cite{hplflow_repo, pointpwc_repo}. The flow target is computed by point wise subtraction $F = C_{t+1} - C_t$. This operation is only valid if $\vec{p}_i^{(t + 1)} \in C_{t+1}$ corresponds to the point $\vec{p}_i^{(t)} \in C_{t}$. Note that data collected by LiDAR sensors, such as from KITTI Scene Flow \cite{kitti}, do not have this property. 

There are, however, a few considerations to be made. The data used as input of the models are samples drawn from the original point clouds. $N$ points are sampled without replacement from the original point clouds, $C'_{t+1} \subseteq C_{t+1}$, and $C'_{t} \subseteq C_{t}$. The flow vectors are selected such that $\vec{f}_j \in F'$ corresponds to point  $\vec{p}_j^{(t)} \in C'_{t}$. If $N$ is the number of available points, this operation is reduced to shuffling $C_t$ and $C_{t+1}$. The known correspondences are lost, but there are point correspondences still. We argue that this sampling operation does not adequately approximate the Re-sampling Mechanism. The field of view is effectively distorted and occlusions in $C_t$ are carried over to $C_{t+1}$ without regard to the actual trajectory of the different objects.

In summary, the works of \cite{HPLFlowNet, PointPWCNet} and \cite{FlowNet3D, FlowNet3D++} are tackling different problems by making a different assumption on the mechanism behind the data gathering. To fairly compare the different works, we found it useful to differentiate between mechanisms. We regard the Re-sampling Mechanism as the most representative for the scene flow task.


\section{Flow Targets Found in Self-Supervision}
\label{sec:relatedtarget}

We follow to explain our perspective on self-supervision and how we interpret the approaches proposed by \cite{SelfSupervisedFlow, PointPWCNet}. Many variants of self-supervision have been proposed in different fields of machine learning. All of them share one element in common. Annotated targets are not part of the training process, the data used as input to train the model is also used as a target in training time. For the task of flow estimation, the supervision signal comes from the sequential nature of data collation itself.

The works of \cite{SelfSupervisedFlow, PointPWCNet} have scientific value, among other contributions, they opened the field of scene flow estimation to self-supervised approaches. However, they include ground truth flow information in different ways in their proposed approaches. The losses proposed by \cite{SelfSupervisedFlow} unlock the use of large datasets that are not annotated for scene flow. However, the method requires a pre-trained flow estimator and further supervised fine-tuning. The self-supervised step has a limited contribution to the final performance of the model. In our experiments in Section~\ref{sec:nnexps} we show the main limitations of the approach when no supervised step is taken.

The self-supervised training proposed by \cite{PointPWCNet} does not require a pre-trained model. It does, however, assume the Correspondence Mechanism for training and evaluation. As explained in the Section~\ref{sec:relatedmech}, the ground truth flow vectors are implicitly incorporated in the input of the model for training and evaluation. We explain the Correspondence and Re-sampling Mechanisms and their immediate consequences to scene flow estimation in Section~\ref{subsec:corrvssamp}. To assume one or the other mechanism has a great impact on the performance of the model, as shown in Section~\ref{subsec:assumption}.

To the best of our knowledge, a self-supervised training setup that does not require ground truth flow vectors in any way is still to be introduced. The authors of \cite{SelfSupervisedFlow, PointPWCNet} make relevant scientific contributions to the field of scene flow. Nevertheless, we contest their claim on performing self-supervised scene flow estimation. 

\section{Modeling Choices}
\label{sec:modelchoices}

In scene flow, we are interested in estimating the flow field of a dynamic scene. Most of the previous work has approached the task in a supervised paradigm. In this section, we draw parallels between the supervised and self-supervised methods previously proposed. We then briefly motivate a generative modeling approach to flow estimation.

In order to draw a parallel between the supervised and self-supervised methods introduced in the Chapter~\ref{sec:litreivew}, we will review some basic concepts of machine learning theory. The supervised approaches of \cite{FlowNet3D, FlowNet3D++, HPLFlowNet, PointPWCNet} aim to model a discriminative function, those are the neural network architectures introduced by each one. A discriminative function $h_\Theta(\cdot)$ maps an input $(C_1, C_2)$ directly to an output $\hat{F}$ \cite{bishop2007}, $h_\Theta(C_1, C_2) = \hat{F}$, where $\Theta$ are learnable parameters.

Scene flow estimation is a multivariate regression task. We may optimize the parameters of our discriminative function by minimizing the mean squared error (MSE) between target and estimation, $||\hat{F} -  F||_2^2$, \cite{bishop2007}. Convergence to the ground truth flow vectors is expected\footnote{The discriminative functions used by \cite{FlowNet3D, FlowNet3D++, HPLFlowNet, PointPWCNet} are not linear, thus convergence to a local minimum is possible.} if the flow targets are off by a normally distributed noise. In practice, better results are achieve when the L2 loss, $||\hat{F} -  F||_2$, is minimized instead. This allows us to relax the assumption regarding the distribution of the noise, from a normal distribution to just an unknown unimodal distribution. In summary, the supervised solutions \cite{FlowNet3D, FlowNet3D++, HPLFlowNet, PointPWCNet} make use of a discriminative function that is optimized under the assumption the noise on the flow targets is unimodal.

The self-supervised approaches proposed by \cite{SelfSupervisedFlow, PointPWCNet} also aim to use a discriminative function. Namely, FlowNet3D and PointPWC-net respectively. The authors of \cite{SelfSupervisedFlow, PointPWCNet} also aim to optimize those discriminative functions via minimization of the L2 loss. However, the flow supervision is replaced by a nearest neighbor based self-supervision. We point out the following fundamental issues with this setup.

Let $\hat{p}_i = \vec{p}_i^{(1)} + \hat{f}_i$ be the estimated location of the point $\vec{p}_i^{(1)} \in C_1$ when translated by the estimated flow vector $\hat{f}_i$. And let $\vec{p}_j^{(2)} \in C_2$ be the nearest neighbor to $\hat{p}_i$. We enumerate two fundamental issues in using $||\hat{p}_i - \vec{p}_j^{(2)}||_2$ as a loss. First, there are no guaranties $\vec{p}_j^{(2)}$ is the nearest neighbor of $\vec{p}_i^{(1)} + \vec{f}_i$, where $\vec{f}_i$ is the ground truth unknown vector. Second, the error around the target is not expected to be unimodal any longer. If $\vec{p}_i^{(1)}$ is translated by a slightly different estimated flow vector it may have a different nearest neighbor $\vec{p}_k^{(2)} \in C_2$. In summary, the nearest neighbor assignment is ill-equipped for optimizing a discriminative function using an L2 loss. 

We use this last insight to propose a generative modeling approach for scene flow estimation.  Generative models aim to - implicitly or explicitly - approximate the probability distribution of inputs as well as of the outputs \cite{bishop2007}. Generative Adversarial Nets (GANs) \cite{gan} introduced adversarial learning to the field of machine learning. Adversarial training has been successfully applied to many tasks other than image generation \cite{elgan, Yan_2019, Zhang2018SegGANSS, Ganbler}. To the best of our knowledge, we are the first to introduce adversarial training to the task of scene flow on point clouds. In the Chapter~\ref{sec:method} we explain our proposed approach in detail. 

%% file: src/mechanism.tex
\chapter{Correspondence vs. Re-Sampling}
\label{subsec:corrvssamp}

\begin{figure}
    \centering
    \includegraphics[width = .75\linewidth]{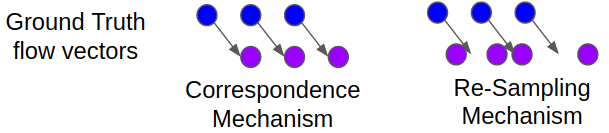}
    \caption{Illustrative example of the Correspondence and the Re-sampling Mechanisms. The blue points belong to $C_1$ and the purple points belong to $C_2$. When correspondence is assumed, each point at $C_2$ has a corresponding point at $C_1$. When Re-Sampling is assumed this correspondence is not present any longer, and the number of points may be different at each time step.}
    \label{img:link}
\end{figure}

We found that previous works in the literature make different assumptions on how the data is gathered. In this section, we define the \emph{Correspondence Mechanism} or the \emph{Re-sampling Mechanism}\footnote{In the field of fluid dynamics, the Correspondence Mechanism is called the Lagrangian specification of the flow field. That is, the flow field is defined by individual particles. The Re-sampling Mechanism is called the Eulerian specification of the flow field. The flow field is defined by a volume through which the fluid flows\cite{fluiddynamics}.}, as illustrated by Figure~\ref{img:link}.

\begin{figure}
    \centering
    \includegraphics[width = .75\linewidth]{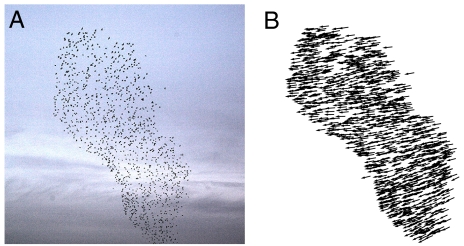}
    \caption{Snapshot of a flock (A) and the velocity vector of individual birds (B). Figure taken from \cite{flock}.}
    \label{img:flock}
\end{figure}

Two similar tasks will be used to guide the reader through the concepts. One is performing scene flow estimation from LiDAR sweeps. A LiDAR is mounted on a car and driven on a trajectory scanning the environment around it, the data is stored as a sequence of point clouds, the points are sampled from surfaces of moving and static objects. For each point cloud, we want to estimate the velocity vector of every point. The other task is performing bird tracking in a moving flock \cite{flock}, Figure~\ref{img:flock} illustrates the task. Each bird has its own GPS tracker. The data is also stored as a sequence of point clouds, each point belongs to a particular bird in the flock. Though, we may or may not have a reference (such as an id.) to the individual birds. We want to estimate the velocity of each bird at each time step.

We call it the Correspondence Mechanism when $C_2 = C_1 + F$ is implied. It may assume Point Clouds are - unordered - sets of points. It is not necessarily known which point in $C_1$ corresponds to which point in $C_2$, but the correspondence exists. This mechanism results in the following implications for the task of scene flow estimation: 

\begin{enumerate}
    \item Every point in $C_1$ can be traced to $C_2$ deterministically. Let $\vec{p}_i^{(1)} \in C_i$ and its correspondent $\vec{p}_j^{(2)} \in C_2$ then p$(\vec{p}_i^{(1)} + \vec{f}_i^{(1)}) - \vec{p}_j^{(2)} = \vec{0}$.
    \item It is agnostic or ignorant to occlusions. For instance, if an object is occluding another object in $C_1$, then this occlusion is translated to $C_2$ even if the objects have different motion trajectories. 
    \item The field of view changes to adapt to the motion. As points do not disappear nor do new points appear, it effectively implicates that the field of view has to adapt to include all moved points. 
\end{enumerate}

From the listed implications we understand that assuming the Correspondence Mechanism is reasonable in situations where points have individual importance. For bird tracking, the position of each bird is known at each time step. Thus the velocity vector is just the difference between the positions. The concept of occlusion and field of view are ill-defined in this situation. One bird cannot occlude the GPS signal of another bird, nor will they ever leave the range of GPS coverage. Other examples of fields that may benefit from the Correspondence Mechanism are particle tracking in fluid mechanics and motion capture in computer vision.  

Particularly in the context of scene flow, the Correspondence Mechanism can be used for synthetic data. For instance, point clouds of ShapeNet \cite{shapenet} have no self-occlusion, by applying a deterministic transformation to the point cloud we will have $C_2 = C_1 + F$. However, this assumption does not hold when we consider the real 3D data collected by a LiDAR, where $C_2 \neq C_1 + F$ as illustrated by Figure~\ref{img:link}. The data collection has no way of accounting for points that leave the field of view or that are occluded by a change in the scene. What we call the Re-sampling Mechanism, is that $C_1$ and $C_2$ are independent samples of points from a dynamic scene taken at different time steps. Note that the individual points are not particularly relevant, instead, we are interested in the surfaces and objects they were sampled from. The implications, in the same order as for the Correspondence Mechanism, are the following.

\begin{enumerate}
    \item There is no deterministic function that maps $C_1$ to $C_2$. The flow vector translates the points in $C_1$ to their would-be positions in the time $C_2$ is sampled. No point in $C_2$ has a respective point in $C_1$.
    \item Occlusions and dis-occlusions. Points can only be sampled at the surface of the object that faces the sensor. Thus every object has self-occlusion and objects may occlude each other. Those occlusions change as the objects move.
    \item The field of view is only dependent on the trajectory of the sensor and not dependent on the motion of the objects it records. Objects may leave and enter the field of view.
\end{enumerate}


\begin{figure}
    \centering
    \includegraphics[width = .75\linewidth]{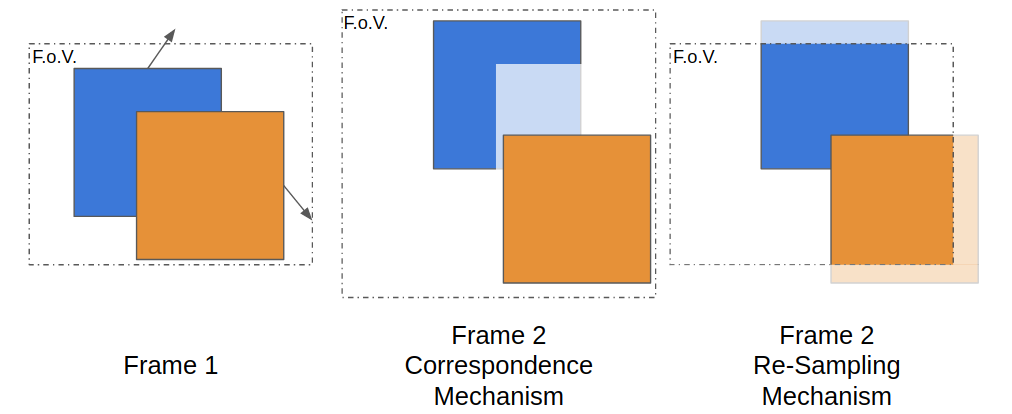}
    \caption{Illustrative example of the Correspondence and the Re-sampling Mechanism. The shaded colors represent the areas of the objects that are not visible when using one or other assumption. F.o.V stands for Field of View.}
    \label{img:occex}
\end{figure}

Figure~\ref{img:occex} aims to illustrate the implications of the occlusions and the field of view. The orange square partially occludes the blue square at frame 1. The Correspondence Mechanism implies that the blue square is just its visible part. At frame 2 the field of view is adjusted to fit the entire objects, but the blue square is kept partially occluded even though no object is performing the occlusion. The implications of the Re-sampling Mechanism are rather the opposite. The points of the blue square that were occluded at frame 1 are now visible, but both squares are partially visible due to the fixed field of view.

Just as assuming the Correspondence Mechanism is ill-suited for scene flow estimation on LiDAR sweeps, the Re-sampling Mechanism is ill-suited for bird tracking. The assumptions we defined belong to distinct scientific niches. If a scientific work wants to claim relevance to applications that use LiDAR scans or stereo cameras, our findings favor assuming the Re-sampling Mechanism in the evaluation method. We refer the reader to Section~\ref{subsec:assumption} for further insights.


%% file: src/methods.tex
\chapter{Method}
\label{sec:method}

In scene flow, we are interested in estimating the flow field of a dynamic scene. Given a sequence of point clouds, we want to estimate the flow vectors for the individual points. We aim to tackle the task from a self-supervised perspective. Which means we only have access to sequences of point clouds at training time. The ground truth flow vectors are not accessible. In this chapter, we first list the issues related to point cloud metrics. We then explain our proposed approach and how we tackle each issue. The self-supervised setup is explained in detail in Section~\ref{sec:aml}. The relevant losses are explained in Sections \ref{sec:cclos} and \ref{subsec:mstl}.

Previous work \cite{SelfSupervisedFlow, PointPWCNet} on self-supervised scene flow estimation replaces the flow target with a nearest neighbor based distance. In Section~\ref{sec:modelchoices} we discussed the limitations of the design choice from a machine learning perspective. We are interested in devising a loss metric between the estimated point cloud and the target one, which takes the following issues into account.

\begin{enumerate}
    \item Computational complexity: losses on the point cloud space have computational and space complexity that is at least quadratic in the number of points. Hardware limits are reached rather fast when dealing with real data. Just for comparison, \cite{PointPWCNet} reports all their self-supervised results using 8192 points. Yet, a LiDAR sensor such as Velodyne HDL-64E (used to gather data for KITTI \cite{kitti, Menze2018JPRS}) records up to 110k points per sweep. Computing nearest neighbors or chamfer distances on such point clouds is memory intensive and slow. Instead, it would be interesting to learn a metric of the distance between point clouds that is linear in the number of points. 
    \item The Re-sampling Mechanism: as explained in Section~\ref{subsec:corrvssamp}, the points of a point cloud are assumed to be samples from a surface. However, the nearest neighbor-based performs assignments between points, thus it is rather sensitive to the set of sampled points. It would be beneficial to perform the loss calculations on a space that takes this sampling into consideration. 
    \item Partial observability: nearest neighbor-based losses perform assignments between points, which makes those losses sensitive to occlusions in a scene. It would be useful to have a loss that treats points of fully visible objects differently than points of occluded objects.
\end{enumerate}

We aim to tackle those three issues by learning a metric between the estimated and the target point cloud. In our setup, we use a Flow Extractor to perform scene flow estimation and a Cloud Embedder to perform metric learning. The two models are trained in an adversarial fashion. An intuitive explanation of how our setup tackles each issue follows.

\begin{enumerate}
    \item Computational complexity: the loss calculation happens in a latent space, thus it is linear in the number of points in the scenes. The forward pass of the models is dependent on the number of points and hardware limitations are not entirely ruled out. In practice, our approach makes significantly more efficient use of hardware than nearest neighbor-based losses.
    \item The Re-sampling Mechanism: two point clouds sampled from the same scene may look different. Yet, the Cloud Embedder learns to map both of them to the same neighborhood of the latent space. The Re-sampling Mechanism is taken into account in the design of the training setup. 
    \item Partial Observability: the Cloud Embedder may learn to ignore points that are not relevant for the loss calculations. The model extracts features for individual points, some of those features may encode information about occlusions. We note however that such property is not enforced during training.
\end{enumerate}

Having motivated our proposed approach we proceed to explain each one of its components in further detail. The self-supervised setup is explained in Section~\ref{sec:aml}. The relevant losses are explained in Section~\ref{sec:cclos} and Section~\ref{subsec:mstl}.


\section{Adversarial Metric Learning}
\label{sec:aml}

In this section, we introduce a novel the self-supervised training setup \emph{Adversarial Metric Learning} proposed for scene flow estimation. We first explain the models used in the setup. Then we provide a step by step description of the training. Lastly, we provide our intuition on the setup and the training of the two models.

As previously motivated, we devise an adversarial setup where we aim to simultaneously learn flow estimation and a metric between the estimated and the target point clouds. The flow estimation is performed by a learnable model we call the \emph{Flow Extractor}. The metric is learned by a fully differentiable model we call the \emph{Cloud Embedder}. Analogous as for image generation using GANs \cite{gan} the Flow Extractor takes the role of the Generator and the Cloud Embedder takes the role of the Discriminator.

The Flow Extractor can be any model that performs flow estimation on sequences of point clouds and that can be trained via gradient descent techniques such as FlowNet3D \cite{FlowNet3D} and PointPWC-net \cite{PointPWCNet}. 

The Cloud Embedder receives as input a point cloud and outputs a fixed-size vector. Note that number or order of points in the input does not impact the outputed vector. The model should learn to map similar point clouds to the same neighborhood of the latent space and dissimilar point clouds to distant regions. In practice, the Cloud Embedder is a neural network. We use architectures based on \cite{PointNet++, PointNet}, but others are possible. The Cloud Embedder is thus trained via metric learning. More specifically, we make use of the Multi-Scale Triplet loss, the loss is explained in detail in the Section~\ref{subsec:mstl}.

\begin{figure}
    \centering
    \includegraphics[width = 1\linewidth]{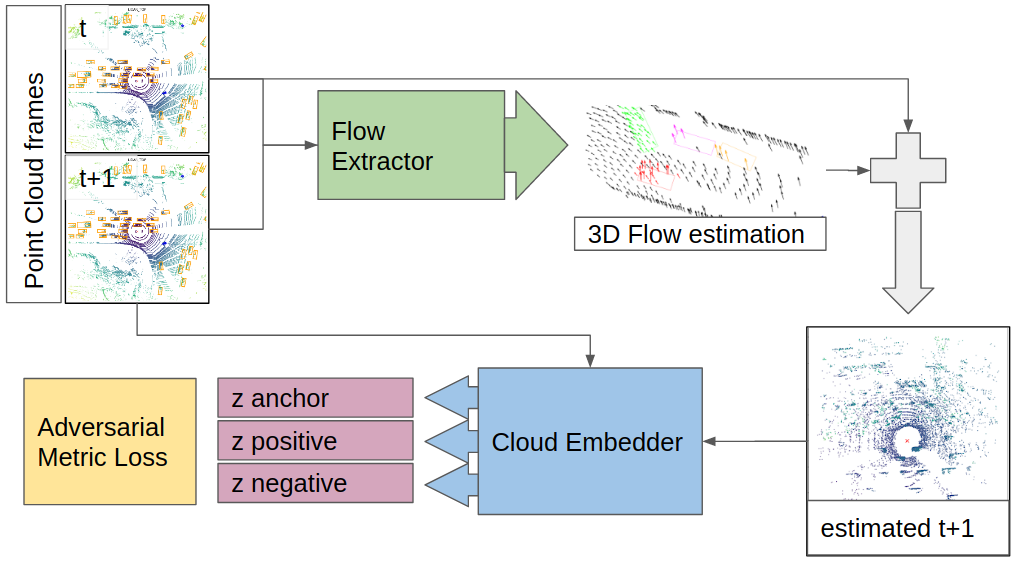}
    \caption{Illustration of the setup for Adversarial Metric Training}
    \label{img:setup}
\end{figure}

The Cloud Embedder is responsible for providing the Flow Extractor with feedback to improve its flow estimations. The Flow Extractor has to improve its flow estimations so the predicted point cloud resembles the target the most. The Cloud Embedder, on its turn, has to learn to differentiate between the target and the predicted point cloud. The setup is illustrated on Figure~\ref{img:setup}, the training can be summarized in the following steps.

\begin{enumerate}
    \item The Flow Extractor $\Phi(\cdot, \cdot)$ performs forward and backward flow estimation:
        \begin{align}
            \hat{F}_\rightarrow &= \Phi(C_t, C_{t+1}),   \\
            \hat{C}_{t+1} &= C_t + \hat{F}_\rightarrow, \\
            \hat{F}_\leftarrow &= \Phi(\hat{C}_{t+1}, C_t),
        \end{align} where $\hat{F}_\rightarrow$ is the estimated forward flow, $\hat{F}_\leftarrow$ is the estimated backward flow, and $\hat{C}_{t+1}$ is the estimated point cloud at time $t+1$.
    \item $C_{t+1}'$ and $C_{t+1}''$ are sampled from $C_{t+1}$ without overlap. The Cloud Embedder $\Psi(\cdot)$ maps the point clouds to latent representatios:
    \begin{align}
            \vec{z}_a &= \Psi(C_{t+1}'), \\
            \vec{z}_p &= \Psi(C_{t+1}''), \\
            \vec{z}_n &= \Psi(\hat{C}_{t+1}),
        \end{align} where $\vec{z}_a$ is the anchor, $\vec{z}_p$ is the positive example and $\vec{z}_n$ is the negative example used in the triplet loss.
    \item Multi-Scale Triplet loss, $\mathcal{L}_{\mst}(\cdot, \cdot, \cdot)$, defined by Equation~\ref{eq:mst}, is used to train the Cloud Embedder : 
        \begin{equation}
            \mathcal{L}_{\Psi} = \mathcal{L}_{\mst}(\vec{z}_a, \vec{z}_p, \vec{z}_n), 
            \label{eq:ce}
         \end{equation} 
    \item The Flow Extractor is trained using two losses. The euclidean norm given by $\mathcal{L}_{\ml2}(\cdot, \cdot)$ defined by Equation~\ref{eq:ml2}, and Cycle Consistency loss $\mathcal{L}_{\cc}(\cdot, \cdot)$ defined by Equation~\ref{eq:ccl}:
         \begin{equation}
            \mathcal{L}_{\Phi} = \mathcal{L}_{\ml2}(\vec{z}_a, \vec{z}_n) + \gamma_{\cc} \mathcal{L}_{\cc}(\hat{F}_\rightarrow, \hat{F}_\leftarrow),
            \label{eq:fe}
         \end{equation}
        where $\gamma_{\cc}$ is a scaling hyper parameter. 
\end{enumerate}

\begin{figure}
    \begin{subfigure}{0.40\textwidth}
        \centering
        \includegraphics[width = 0.95\linewidth]{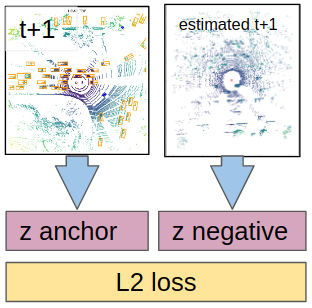}
        \caption{Flow Extractor is trained to minimize the L2 distance between $C_{t+1}$ and $\hat{C}_{t+1}$ in the latent space. \textcolor{white}{Flow Extractor is trained to minimize the L2 distance between $C_{t+1}$ and $\hat{C}_{t+1}$ in the latent space. Flow Extractor is trained to minimize the L2 space.}}
        \label{img:FEloss}
    \end{subfigure}
    \begin{subfigure}{0.60\textwidth}
        \centering
        \includegraphics[width = .95\linewidth]{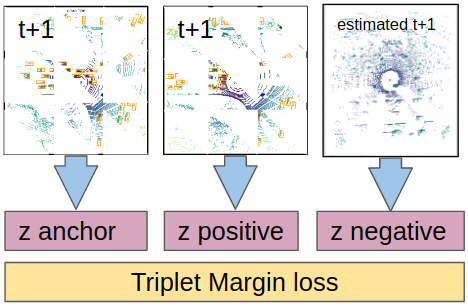} \caption{Cloud Embedder is trained to minimize the triplet margin between $pc_{t+1}$ and $\hat{pc}_{t+1}$. Two non-overlapping sub-sets of $pc_{t+1}$ are randomly sampled. The sampling is illustrated as masked areas of the point clouds, however the actual sampling is agnostic to any aspects of the point cloud. That means that all regions are, in expectation, equally represented in both sub-sets.}
        \label{img:CEloss}
    \end{subfigure}
    \caption{Illustration of the inputs for the loss calculation for the Flow Extractor and for the Cloud Embedder.}
    \label{img:losses}
\end{figure}

The Flow Extractor aims to predict flow vectors that approximate the target point cloud as much as possible. The L2 norm between $\vec{z_a}$ and $\vec{z_n}$ is the quantity to be minimized, Figure~\ref{img:FEloss} illustrates it. We have to select a triplet of examples for training the Cloud Embedder. The negative example is $\hat{C}_{t+1}$. The anchor and the positive example are random non-overlapping sub-samples of $C_{t+1}$, as Figure~\ref{img:CEloss} illustrates.

\begin{figure}
    \centering
    \includegraphics[width = 1.\linewidth]{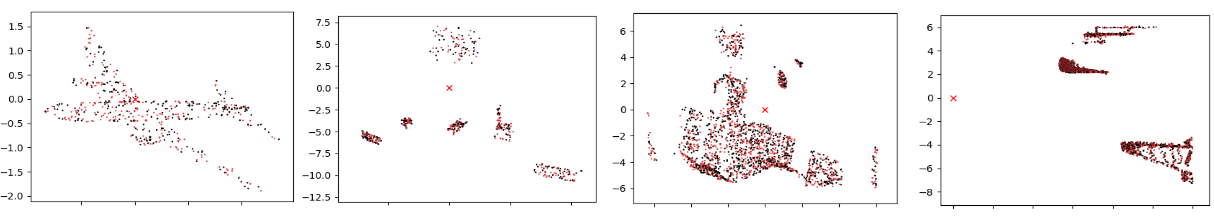}
    \caption{Examples of sub-sampled point clouds visualized by the black and red dots. From left to right the datasets are Single ShapeNet, Multi ShapeNet, FlyingThings3D, KITTI. The points of different colors have uniform coverage of the scene and are expected to represent each object equally well.}
    \label{img:halfcloud}
\end{figure}

We want to call the attention of the reader that the sub-sampling of $C_{t+1}$ is done such that both resulting point clouds are expected representative of $C_{t+1}$, Figure~\ref{img:halfcloud} illustrates the subsamples. Both samples come from the same underlying distribution. Thus it should be fairly simple for the Cloud Embedder to map $\vec{z}_a$ and $\vec{z}_p$ close together and far from $\vec{z}_n$. The latter is expected to come from a distant distribution at the beginning of the training. When the Nash Equilibrium of the adversarial training is reached we expect the Flow Extractor to have learned the real flow.

\section{Cycle Consistency Loss}
\label{sec:cclos}

If we give the sequence $[\hat{C}_{2}, C_{1}]$ to the Flow Extractor, it should estimate a backward flow $\hat{F}_\leftarrow$ that is opposite to the forward flow $\hat{F}_\rightarrow$. It pushes the forward flow vectors to cancel the backward flow vectors, it is an incentive for the network to learn a geometrically consistent flow. Cycle Consistency is used in \cite{FlowNet3D, SelfSupervisedFlow} as average L2 norms. 

When L2 is used to enforce cycle consistency, there is a strong incentive for the network to minimize its loss by predicting zero flow vectors. The cosine similarity is an alternative loss that penalizes the incorrect directions but not the incorrect norms of the flow vectors. Intuitively, using cosine similarity in combination with a norm-aware loss may diminish the risk of converging to the local minimum of zero flow vectors. The calculation is as follows:
\begin{equation}
    \mathcal{L}_{\cc}(\hat{F}_\rightarrow, \hat{F}_\leftarrow) = \frac{1}{N} \sum_{i = 0}^N ||\vec{f}_i^{(\rightarrow)} - (-\vec{f}_i^{(\leftarrow)})||_2 + \frac{\vec{f}_i^{(\rightarrow)} \cdot \vec{f}_i^{(\leftarrow)}}{||\vec{f}_i^{(\rightarrow)}||_2\cdot||\vec{f}_i^{(\leftarrow)}||_2},
    \label{eq:ccl}
\end{equation}
where $\vec{f}_i^{(\rightarrow)} \in \hat{F}_\rightarrow$, and $\hat{F}_\rightarrow\in \mathrm{R}^{N \times 3}$ is the estimated forward flow, $\vec{f}_i^{(\leftarrow)} \in \hat{F}_\leftarrow$ and $\hat{F}_\leftarrow \in \mathrm{R}^{N \times 3}$ is the estimated backward flow and $N$ is the number of points. 

We expect cycle consistency loss to be beneficial for enforcing the preservation of local geometries and locally consistent flow. It is also cheap to compute, computational complexity is linear in the number of points.


\section{Multi-Scale Triplet Loss}
\label{subsec:mstl}

Triplet Margin Loss was introduced by \cite{TripletMargin} expanding on the concepts introduced by \cite{TripletMetricLearning}. In short, similar inputs should be mapped to a neighborhood of the latent space and a dissimilar input should be mapped further away. The triplets of inputs are called anchor $x_a$, positive example $x_p$ and negative example $x_n$. The mapping function, $h(.)$, may have learnable parameters:
\begin{align}
    d_p &= ||h(x_a) - h(x_p)||, \\
    d_n &= ||h(x_a) - h(x_n)||, \\
    \mathcal{L}_{\tm}(d_p, d_n) &= \max(0, m + d_p - d_n),
\end{align}
where $(d_p, d_n)$ are the positive and negative distances and $m$ is the margin usually set to $1.0$. In our work, $h(.)$ is the Cloud Embedder $\Psi(\cdot)$. It is a neural network that performs lossy compression from the point cloud to the latent space.
We point to the following limiting factors:

\begin{enumerate}
    \item The latent vector has a fixed number of dimensions. It is effectively a bottleneck where the most relevant information has to be encoded. Whereas a rigid body transformation can be encoded into six dimensions, a complex scene may require an impractically large vector in order to store all necessary information.
    \item Unless enforced, there are no guarantees that by end of the training the bottleneck is used to its maximum information capacity. 
\end{enumerate}

In this work we focus on the first point, we aim to make the information bottleneck less restrictive. However, the second factor is as relevant to improve performance. One LiDAR sweep collects tens of thousands of points, compacting it into a single vector is open research in itself. 

One option to increase the information capacity is to simply increase the number of dimensions of the latent vector. There are, however, hardware limitations to this increase. An alternative is to use intermediary feature vectors from the mapping function as latent vectors. By doing so we increase the amount of information used to compare point clouds without using more memory. Feature maps can be reduced to feature vectors by a pooling operation, such as max or mean pooling. 

We formalize it mathematically. Given a function $h(\cdot)$ and an input $x$ the function performs a hierarchical mapping $h(x) = (\vec{z}^{(0)}, \vec{z}^{(1)}, \dots, \vec{z}^{(L)})$ where $\vec{z}^{(0)}$ corresponds to the activations of the deepest layer and $\vec{z}^{(L)}$ of the most shallow layer:
\begin{align}
    d_p^{(l)} & = ||\vec{z}_a^{(l)} - \vec{z}_p^{(l)}||, \\
    d_n^{(l)} & = ||\vec{z}_a^{(l)} - \vec{z}_n^{(l)}||,
\end{align} 
\begin{equation}    
    \mathcal{L}_{\mst}(x_a, x_p, x_n) = \sum_{l=0}^L \gamma_l \max(0, m + d_p^{(l)} - d_n^{(l)}),
    \label{eq:mst}
\end{equation} 
\begin{equation}
    \mathcal{L}_{\ml2}(x_a, x_n) = \sum_{l=0}^L \gamma_l d_n^{(l)},
    \label{eq:ml2}
\end{equation}
where $\gamma_l$ is a layer scaling factor. 

\begin{figure}
    \centering
    \includegraphics[width = 1\linewidth]{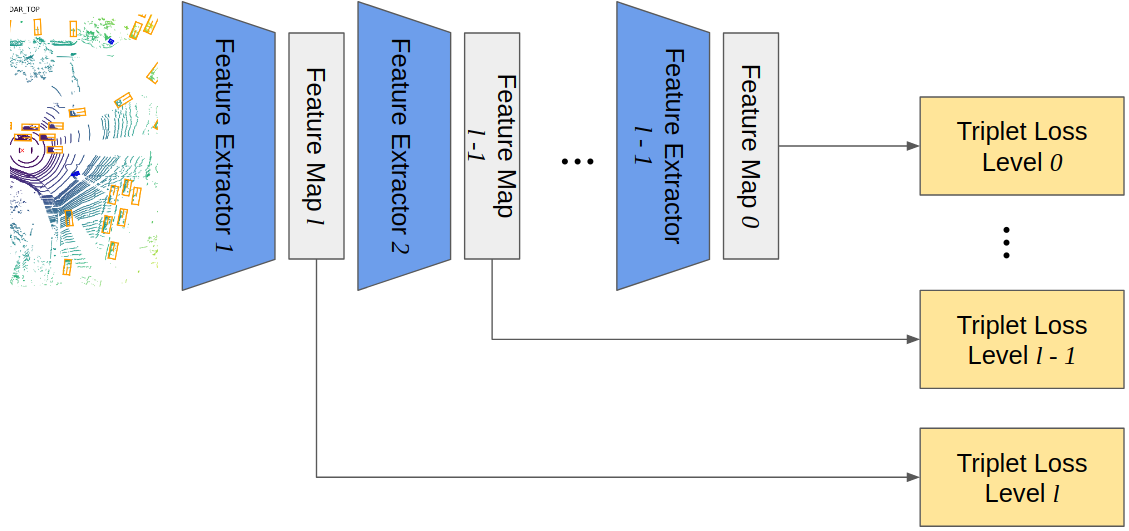}
    \caption{Illustration of the Multi-Scale Triplet Loss. The Feature Extractors can be any differentiable model, in our work we use PointNet++ \cite{PointNet++} Feature Extractors.}
    \label{img:mst}
\end{figure}

We build further on the concept of triplet margin and propose the \emph{Multi-Scale Triplet Loss}. Which uses intermediary feature vectors for metric learning, as illustrated by Figure~\ref{img:mst}. We expect that the Multi-Scale Triplet loss to enlarge the information capacity of the bottleneck and to simplify the optimization process. Previous work \cite{resnet, unet, msggan} has shown that intermediary activations improve the gradient feedback shallow layers receive and improve the trainability of deep architectures. Appendix~\ref{apx:mstl} shows some experiments on clustering and dimensionality reduction. To the best of our knowledge, we are the first to introduce this concept.

%% file: src/dataset.tex
\chapter{The Scene Flow Sandbox}
\label{sec:datasets}

In research, isolated insights often compound into a discovery. We introduce the Scene Flow Sandbox, our proposed benchmark. First, we give a global overview of the benchmark. Then, we briefly explain the aspects of scene flow each dataset focuses on. A detailed explanation of the individual datasets is then given in the following sections. The sandbox is an environment designed to make scene flow experimentation intuitive and insightful. 

The benchmark consists of five datasets designed to study individual aspects of flow estimation in progressive order of complexity, from a single object in motion to real-world scenes. The first three are synthetic datasets, each one incorporates the aspects of real data we are interested in studying. The last two datasets use data collected by LiDAR scans. Flow targets are available in one of them for evaluation purposes. Failures observed on the synthetic datasets are expected to surface on real data. We are mostly interested in exploring the aspects summarized in Table~\ref{tab:datasets} and explained in the following paragraphs.

\begin{table}
    \centering
    \begin{tabular}{@{}lllll@{}}
    \toprule
    Name &
      \begin{tabular}[c]{@{}l@{}}Number of \\ Objects\end{tabular} &
      \begin{tabular}[c]{@{}l@{}}Point \\ Correspondences\end{tabular} &
      Observability &
      Data Generation \\ \midrule
    Single ShapeNet & 1       & $\sim10\%$      & Full    & Synthetic   \\
    Multi ShapeNet  & 2 to 20 & $\sim$0.1\% to 1\% & Full    & Synthetic   \\
    FlyingThings3D  & Many    & None            & Partial & Synthetic   \\
    KITTI           & Many    & None            & Partial & LiDAR scans \\
    Lyft            & Many    & None            & Partial & LiDAR scans \\ \bottomrule
    \end{tabular}
    \caption{Summarization of the differences between datasets. }
    \label{tab:datasets}
\end{table}

The first aspect we are interested in studying is related to the number of objects in a scene. The simplest scene is made of one single object. The geometry of the object does not suffer major changes between two frames. The flow is locally coherent, points belonging to a small neighborhood have flow vectors that are similar in direction and magnitude. Notice that one transformation matrix should be able to fully describe the motion of the scene. A single transformation matrix has no means of describing the motion of a scene that contains two objects with independent trajectories. Even though the geometry of each object must still be kept consistent, the geometry of the scene may drastically change. A well-performing model must estimate locally coherent flow vectors of each object, but independent flow vectors for different objects.

The second aspect regards how the level of observability, or visibility, of a scene impacts its complexity. In a fully visible scene, the model is aware of all the parts of the objects at all moments. In other words, the objects do not occlude each other nor leave the field of view. Partial observability means objects do self-occlude, they may occlude each other, and parts of objects may leave and enter the field of view. Intuitively, it is easier to estimate the motion of fully visible objects, than that of partially occluded objects. We are interested in studying how, and if, a model learns to good motion priors for an occluded region. 

The first two aspects are interesting for spotting failure modes of flow estimators. For instance, a model may capture the motion of a fully visible scene well but fails to capture the motion of a partially visible scene. The failure mode is singled out and attributed to the level of observability. The following two aspects, however, are relevant to understand the level of complexity of the scene but don't necessarily help isolate a problematic factor. 

The third aspect regards point correspondences. We argued in Section~\ref{subsec:corrvssamp} that the Re-sampling mechanism is most representative for data gathered by LiDAR. Each synthetic dataset approximates the Re-sampling mechanism to the limits of its intended complexity. This means the same point may be present in two consecutive frames. The lower this probability, the more complex the scene is.

The fourth and last aspect we take into account is the inherent differences between synthetic and real data collected by LiDAR. The synthetic scenes can be engineered to study particular aspects of scene flow, that is not possible to do with real scenes. There is no control on the number of objects or the type of occlusions. 

The four aforementioned aspects are used in different combinations in five datasets. From the least to the most complex, the sandbox was tailored to facilitate insights when performing experiments. The following sections explain each dataset in more detail, in which the complexity is added in steps, from the motion of a single object to real scenes.


\section{Fully Visible Scene Single Object}
\label{subsec:shapenet}

\begin{figure}
    \begin{subfigure}{0.49\textwidth}
        \includegraphics[width=0.99\linewidth]{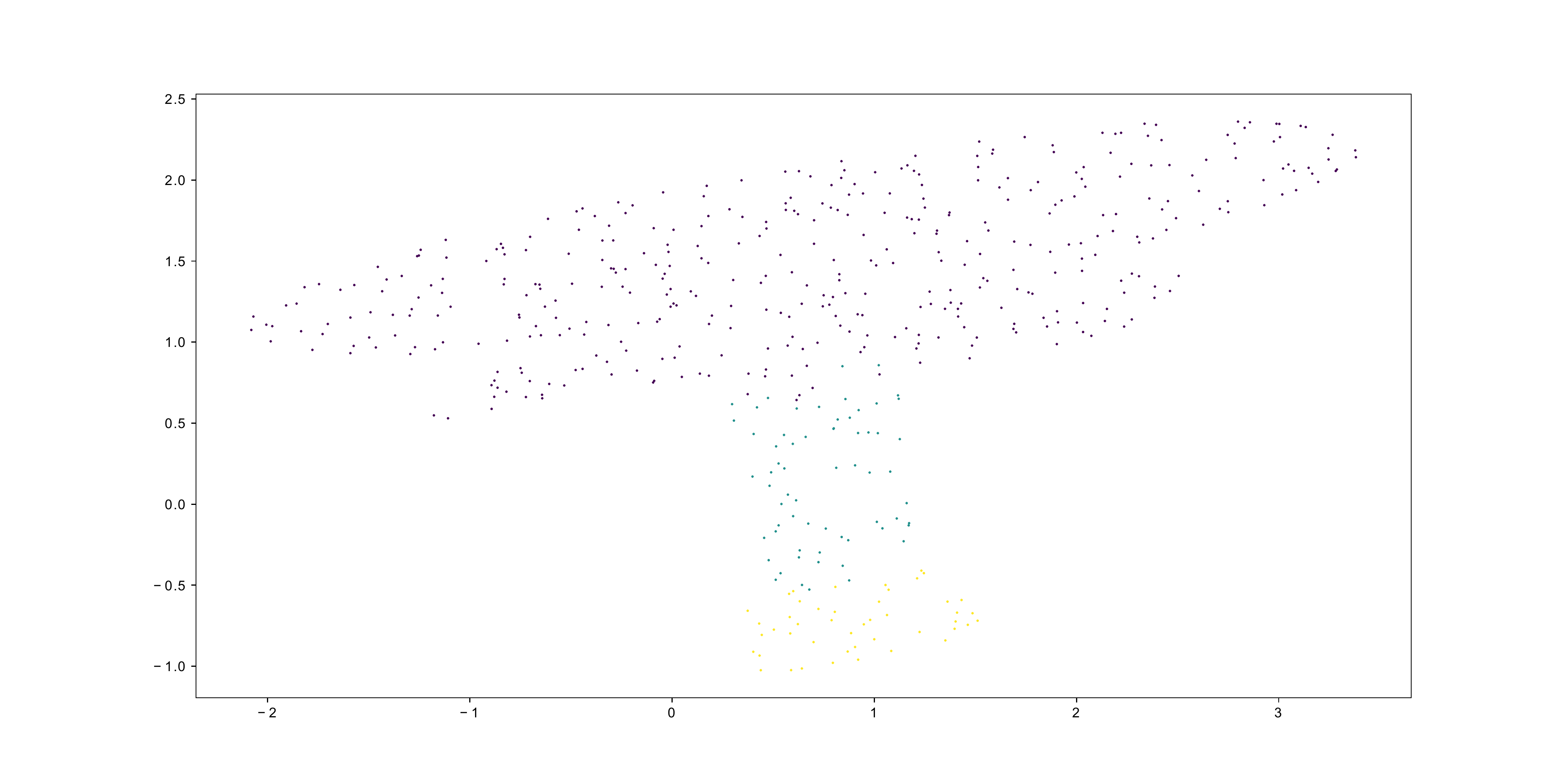}
        \caption{Scene with table.}
        \label{img:ss1}
    \end{subfigure}
    \begin{subfigure}{0.49\textwidth}
        \includegraphics[width=0.99\linewidth]{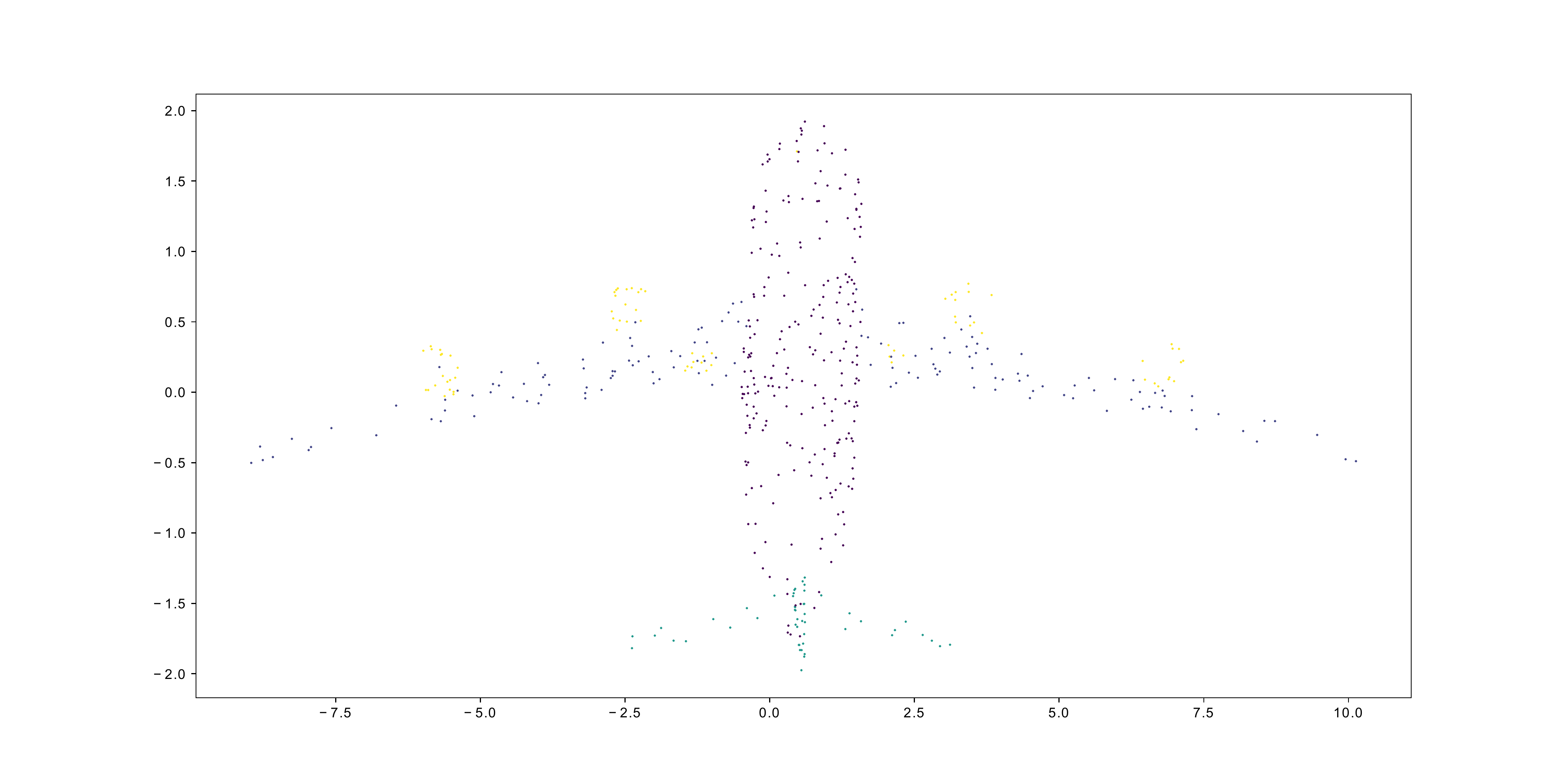}
        \caption{Scene with airplane.}
        \label{img:ss2}
    \end{subfigure}
    \caption{Examples of frames from Single ShapeNet. Colors indicate the sections of the objects.}
    \label{img:singleshapenet}
\end{figure}

We start with fully visible scenes containing only one moving object in it. We call it \emph{Single ShapeNet}. It has one point cloud taken from the surface of an object from ShapeNet \cite{shapenet}. The point cloud is transformed over frames using a transformation matrix. The details are explained in Algorithm~\ref{alg:shapenet} in Appendix~\ref{apx:algos}. The movement of the object can be encoded in nine dimensions: stretch, translation, and rotation relative to the $X, Y, Z$ axis. The aim is to assess if models can learn flow from sequences of point clouds. Figure~\ref{img:singleshapenet} shows two examples.

From Algorithm~\ref{alg:shapenet}, it is evident that at each time step a sub-set of the object point cloud is sampled at random. The scene is fully visible because there are no self-occlusions. The model is aware of all the parts of the object. Correspondences may be present because the same point might be sampled in two consecutive frames. The number of points sampled in the experiments is 512. That is one order of magnitude lower than the total number of points available, which makes potential point correspondences rather low. About 10\% of the points of one frame are expected to appear in the following frame. In general we can say that $C_t \neq C_{t-1} + F_{t-1}$. 

Single ShapeNet is the simplest dataset we use in our experiments. We aim to observe if a model is able of capturing motion from a dynamic scene. The flow estimations should be locally coherent and keep geometric structures. 


\section{Fully Visible Scene Multi Object}

\begin{figure}
    \begin{subfigure}{0.49\textwidth}
        \includegraphics[width=0.99\linewidth]{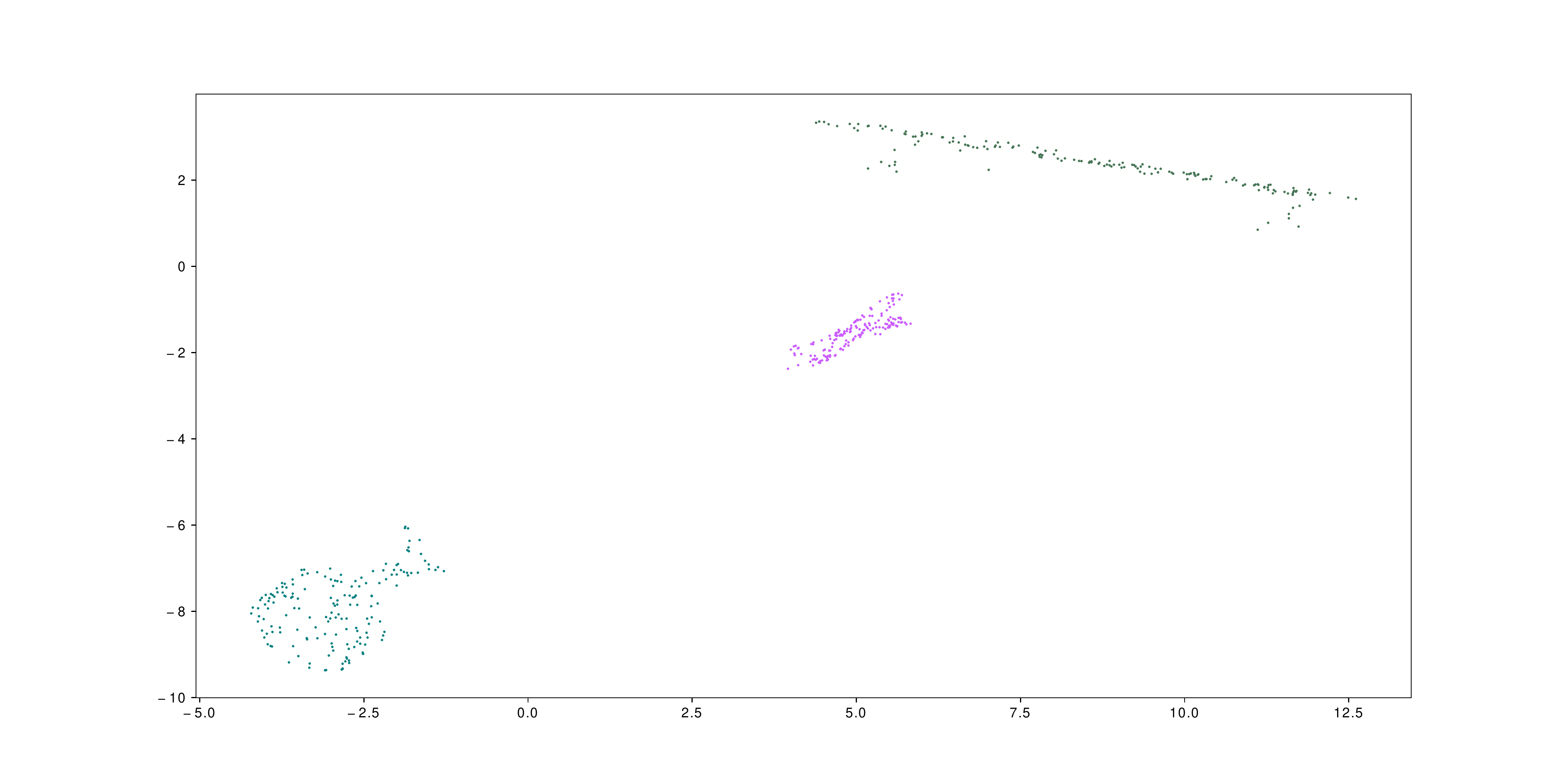}
        \caption{Scene with airplane (pink) and table (gray) and vase (green).}
        \label{img:ms1}
    \end{subfigure}
    \begin{subfigure}{0.49\textwidth}
        \includegraphics[width=0.99\linewidth]{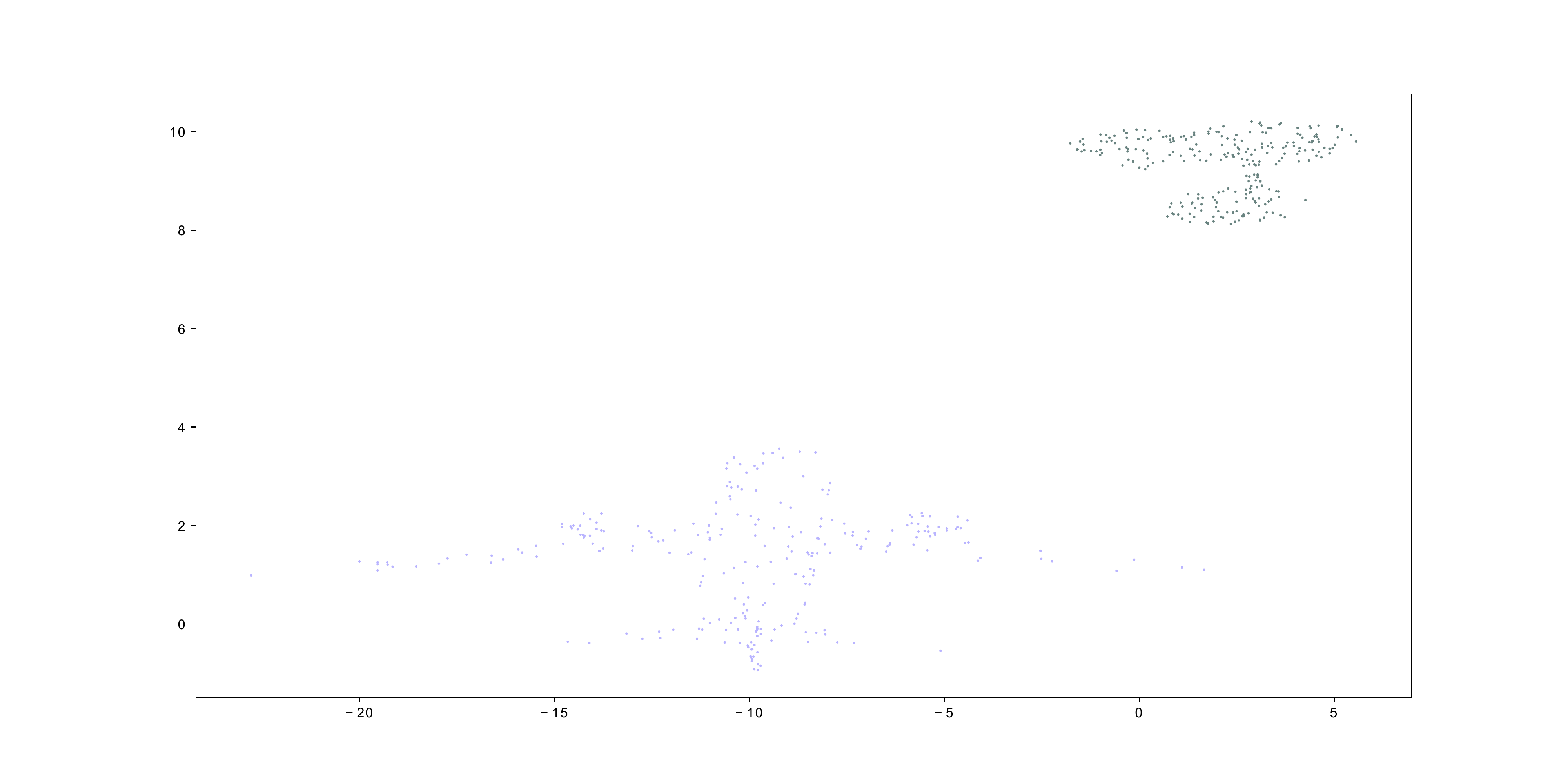}
        \caption{Scene with airplane (light purple) and table (gray).}
        \label{img:ms2}
    \end{subfigure}
    \caption{Examples of frames from Multi ShapeNet.}
    \label{img:multishapenet}
\end{figure}

The second dataset steps up the complexity by having multiple objects in a fully visible scene. We call it \emph{Multi ShapeNet}. It has scenes made of point clouds taken from the surface of a random number of objects from ShapeNet \cite{shapenet}. Each object is transformed by an independent transformation matrix, as shown in Algorithm~\ref{alg:multishapenet} in Appendix~\ref{apx:algos}. Just as in Single ShapeNet, there are no self-occlusions, no inter-object occlusions and points cannot leave the field of view. Figure~\ref{img:multishapenet} shows two examples.

In Section~\ref{subsec:shapenet}, it was explained the correspondences may occur even though the point clouds are independently sampled at each time frame. For the Multi ShapeNet, those correspondences are less present. Out of the 512 points, between 0.1\% and 1\% of the points in one frame are expected to be re-sampled in the next frame. It is possible, however, that an object is represented by only a few points or no points at all.

A well-performed flow estimation will grasp locally coherent and globally independent motion. That is, one object should not suffer major distortions in its trajectory, and different objects have independent trajectories. 


\section{Partially Observable Scene Multi Object}

\begin{figure}
    \begin{subfigure}{0.24\textwidth}
        \includegraphics[width=0.99\linewidth]{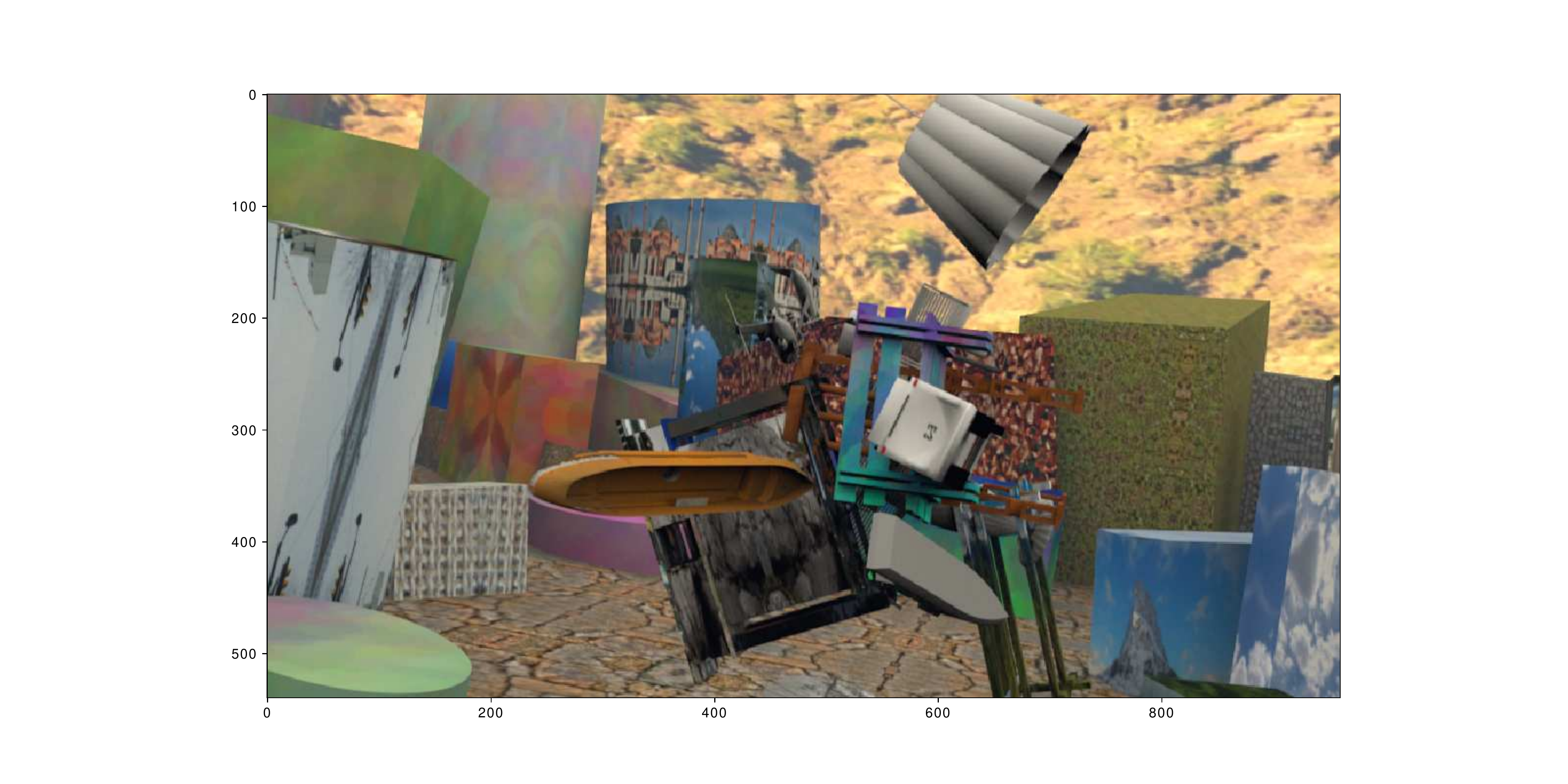}
        \caption{Image of scene \ref{img:xyz1}}
        \label{img:rgb1}
    \end{subfigure}
    \begin{subfigure}{0.24\textwidth}
        \includegraphics[width=0.99\linewidth]{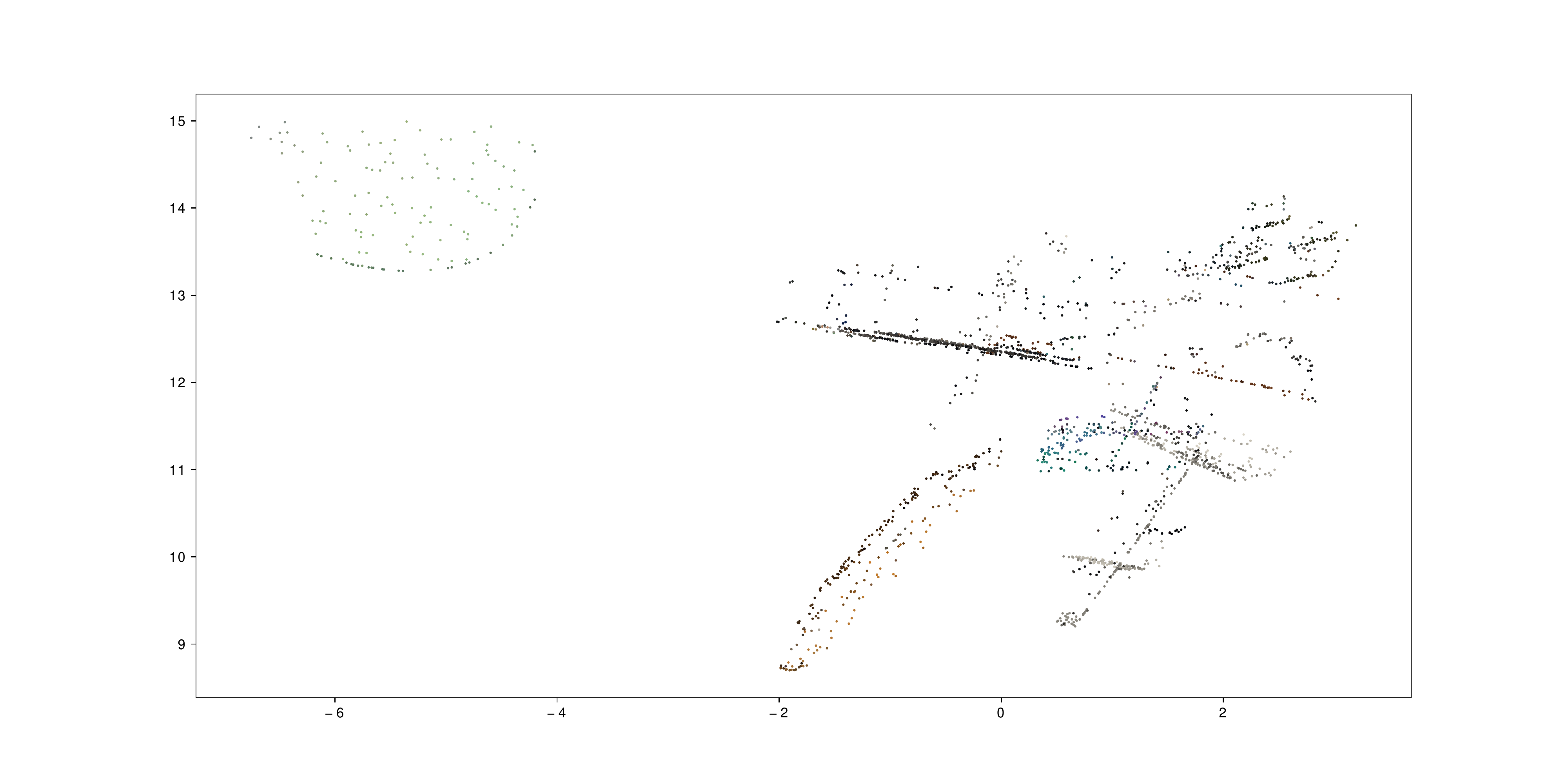}
        \caption{Point cloud of scene \ref{img:rgb1}}
        \label{img:xyz1}
    \end{subfigure}
    \begin{subfigure}{0.24\textwidth}
        \includegraphics[width=0.99\linewidth]{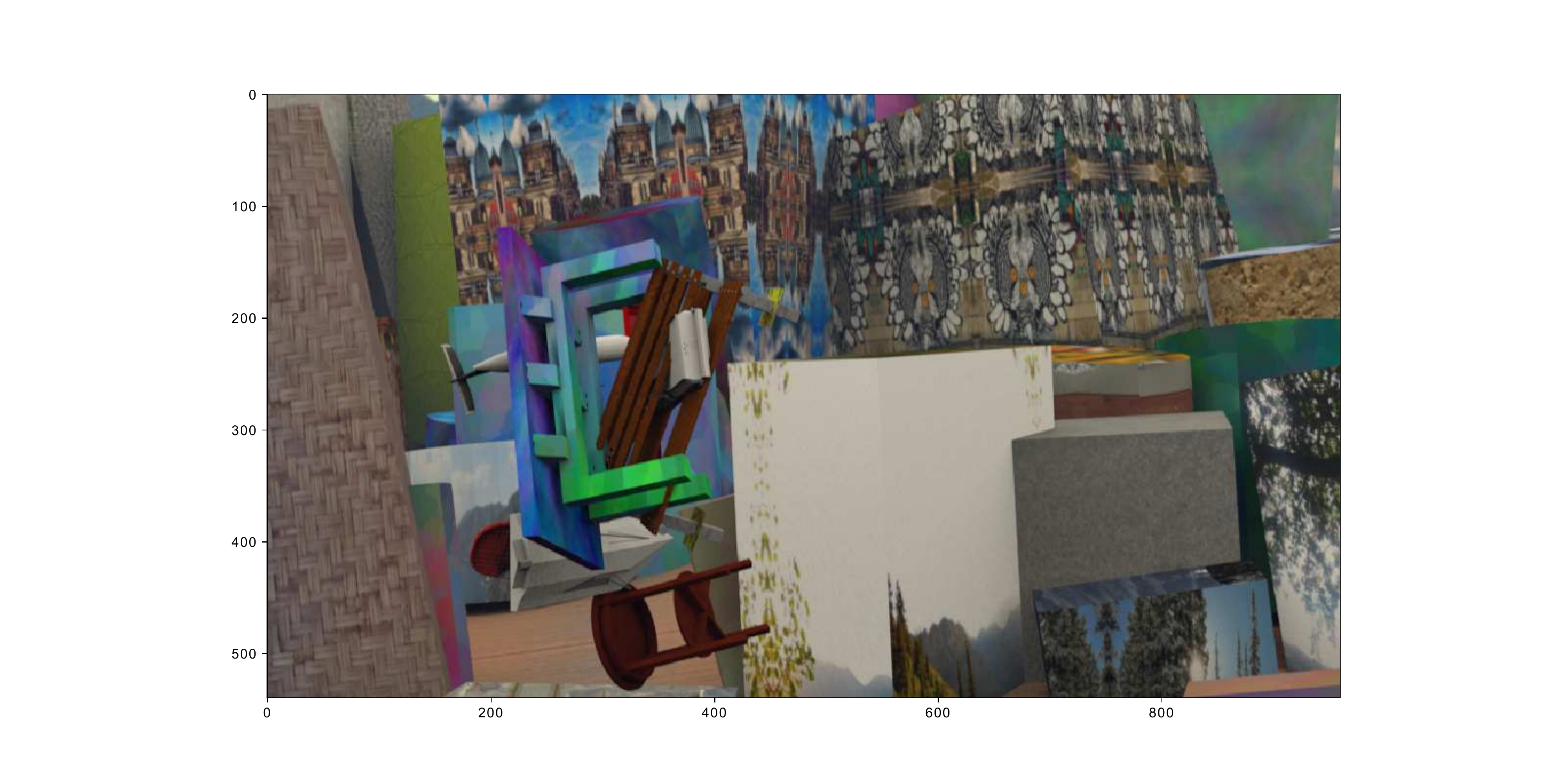}
        \caption{Image of scene \ref{img:xyz2}}
        \label{img:rgb2}
    \end{subfigure}
    \begin{subfigure}{0.24\textwidth}
        \includegraphics[width=0.99\linewidth]{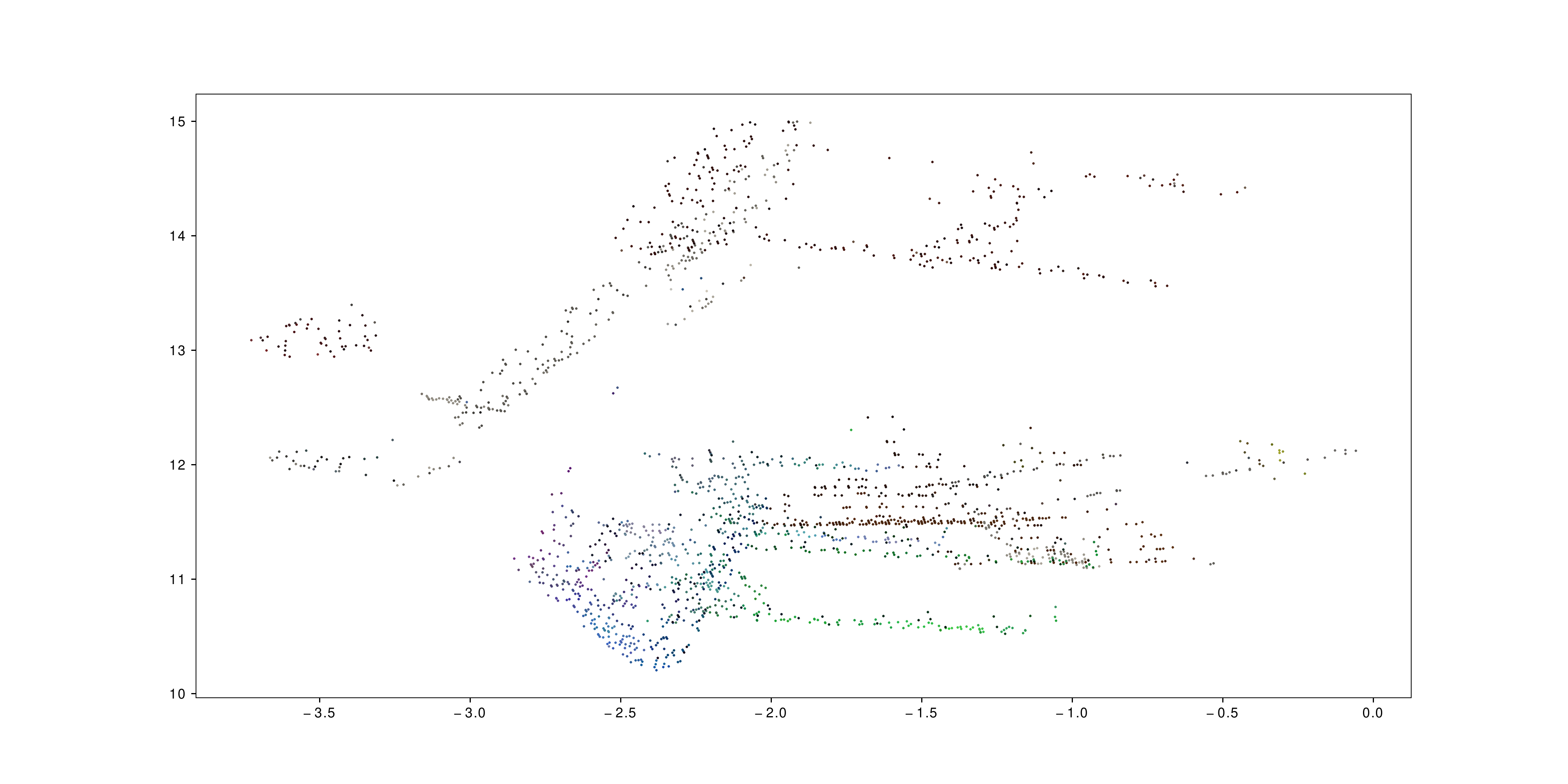}
        \caption{Point cloud of scene \ref{img:rgb2}}
        \label{img:xyz2}
    \end{subfigure}
    \caption{Examples of frames from FlowNet3D. The points of \ref{img:xyz1} and \ref{img:xyz2} were colored with the RGB from \ref{img:rgb1} and \ref{img:rgb2}.}
    \label{img:flownet3d}
\end{figure}

The third and last synthetic dataset has partially observable scenes with multiple objects. Scenes from FlyingThings3D \cite{flyingthings3d} are converted into point cloud format. The original dataset was designed to emulate stereo cameras. Figure~\ref{img:flownet3d} shows the image and the point cloud version of two scenes.

FlyingThings3D makes another step in complexity. Partially observability means objects do self-occlude, they may occlude each other, and parts of objects may leave and enter the field of view. It is even possible for objects to completely disappear from one frame to the next as they leave the field of view or are occluded by other objects. Correspondences are not present in this dataset. Each RGBD frame is independently converted to a point cloud frame with 8192 points. The conversion from RGBD images to point clouds is made explicit in Algorithm~\ref{alg:ft3d} in Appendix~\ref{apx:algos}. Unless explicitly stated, the experiments make use of uniformly sampled 2048 points per frame.

With this dataset we are interested in studying how, and if, a model learns to guess the motion of an occluded object. The objects interact with each other and the movement of a visible object may give cues for estimating the movement of an occluded one. 


\section{Real Scene With Flow Annotations}
\label{subsec:kitti}

The previous three synthetic datasets can be used for quick experimentation and understanding of the difficulties encountered by a proposed model or training method. However, the final validation must happen on real data. We call it \emph{KITTI} after KITTI Scene Flow \cite{kitti}. To the best of our knowledge, it is the only dataset that provides flow annotations from real-world data. There are 200 annotated scenes available in total. Which is rather limiting for training large deep learning models.

The scenes of KITTI Scene Flow were captured by a LiDAR sensor mounted on top of a standard station wagon. The vehicle drove around the mid-size city of Karlsruhe (Germany), in rural areas and on highways. Up to 15 cars and 30 pedestrians are visible per image \cite{Geiger2012CVPR, Geiger2013IJRR}. The authors use the LiDAR data to create disparity maps projected onto the front camera. Even though the 3D data is collected by a LiDAR, the authors post-process it using camera inputs to an RGBD format. We refer the reader to the paper \cite{kitti} for further details of how the data has been selected and processed. 

\begin{figure}
    \begin{subfigure}{0.24\textwidth}
        \includegraphics[width=0.99\linewidth]{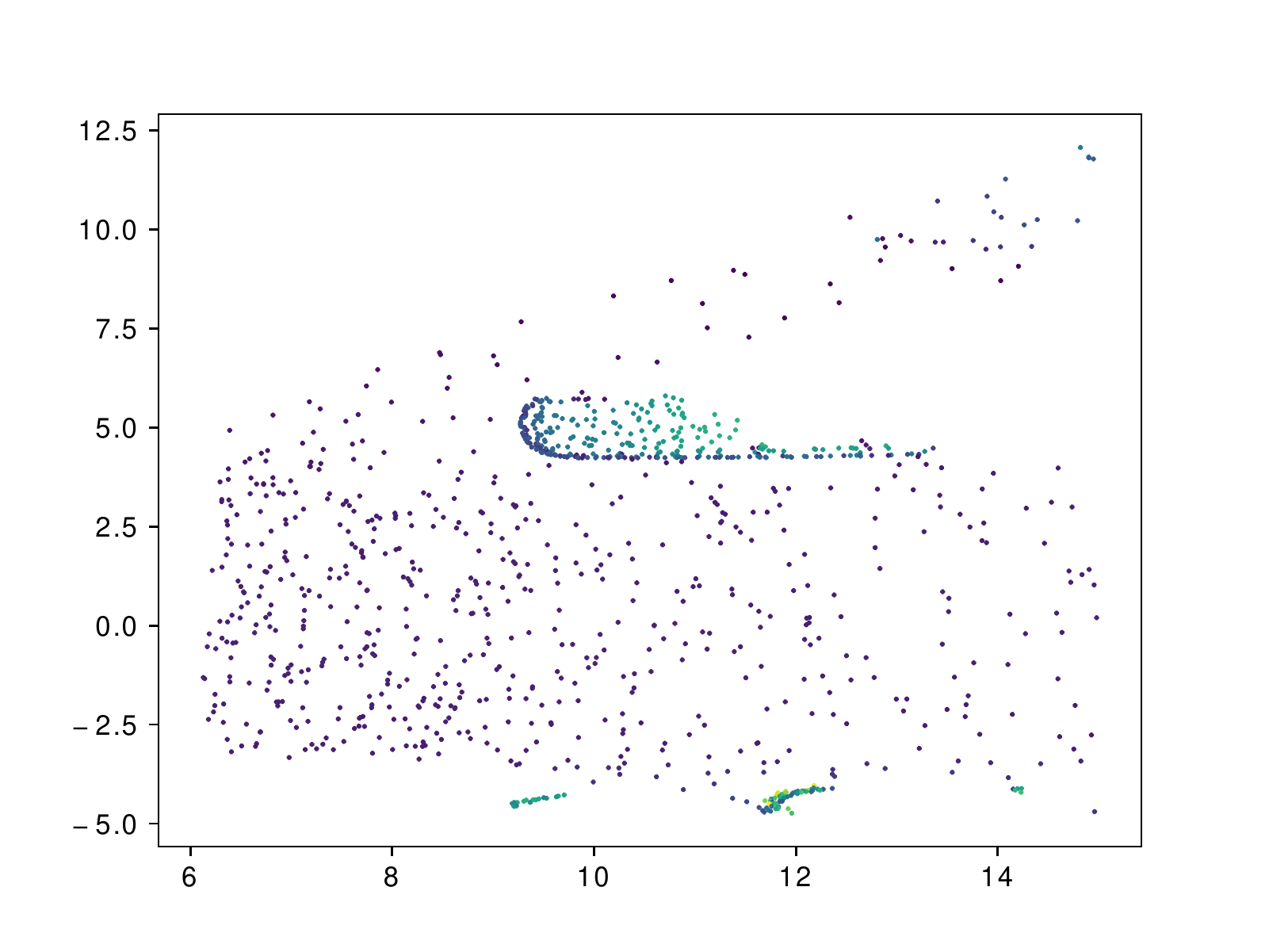}
        \caption{Example of KITTI with ground.}
        \label{img:g1}
    \end{subfigure}
    \begin{subfigure}{0.24\textwidth}
        \includegraphics[width=0.99\linewidth]{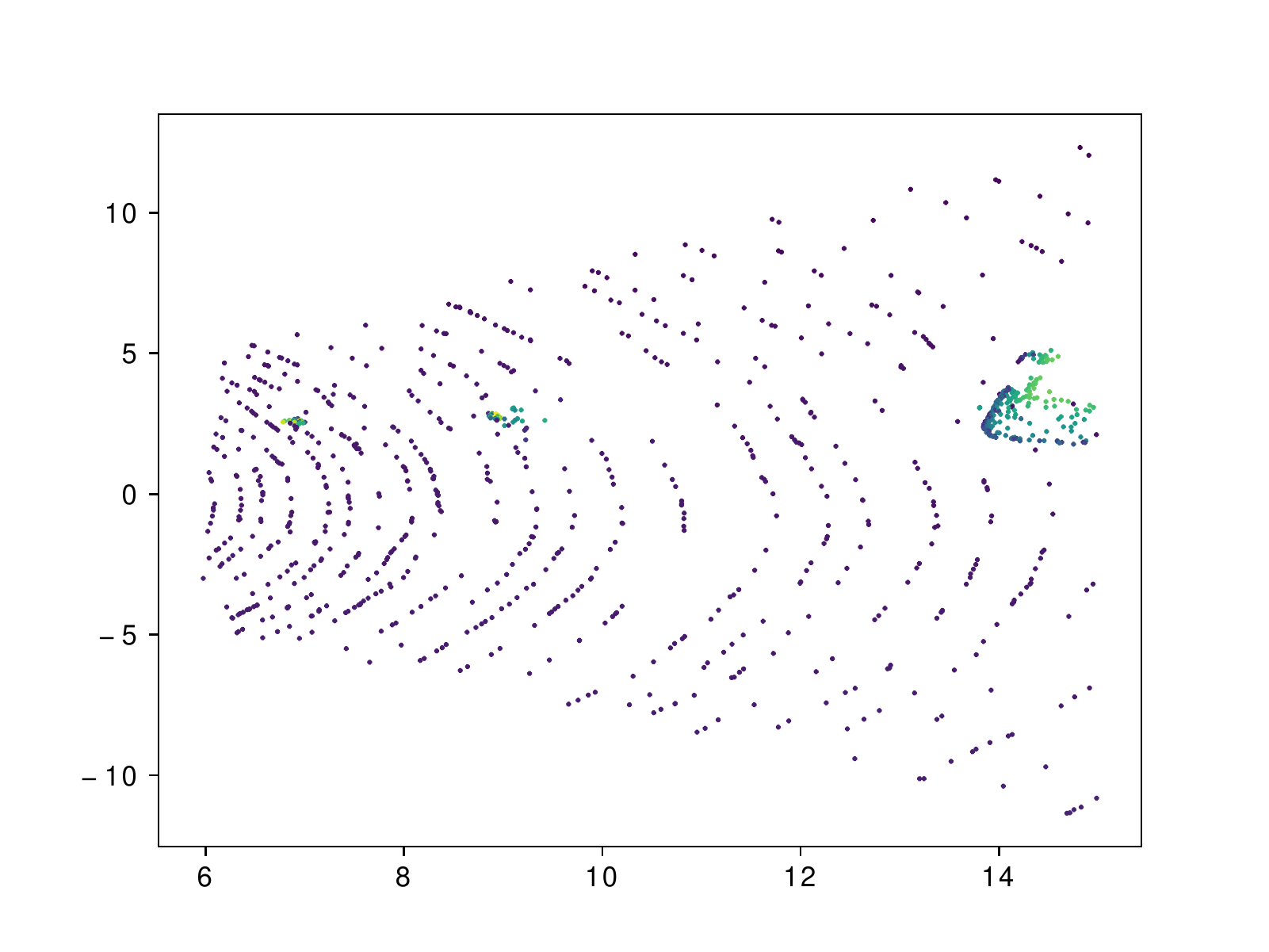}
        \caption{Example of KITTI with ground}
        \label{img:g2}
    \end{subfigure}
    \begin{subfigure}{0.24\textwidth}
        \includegraphics[width=0.99\linewidth]{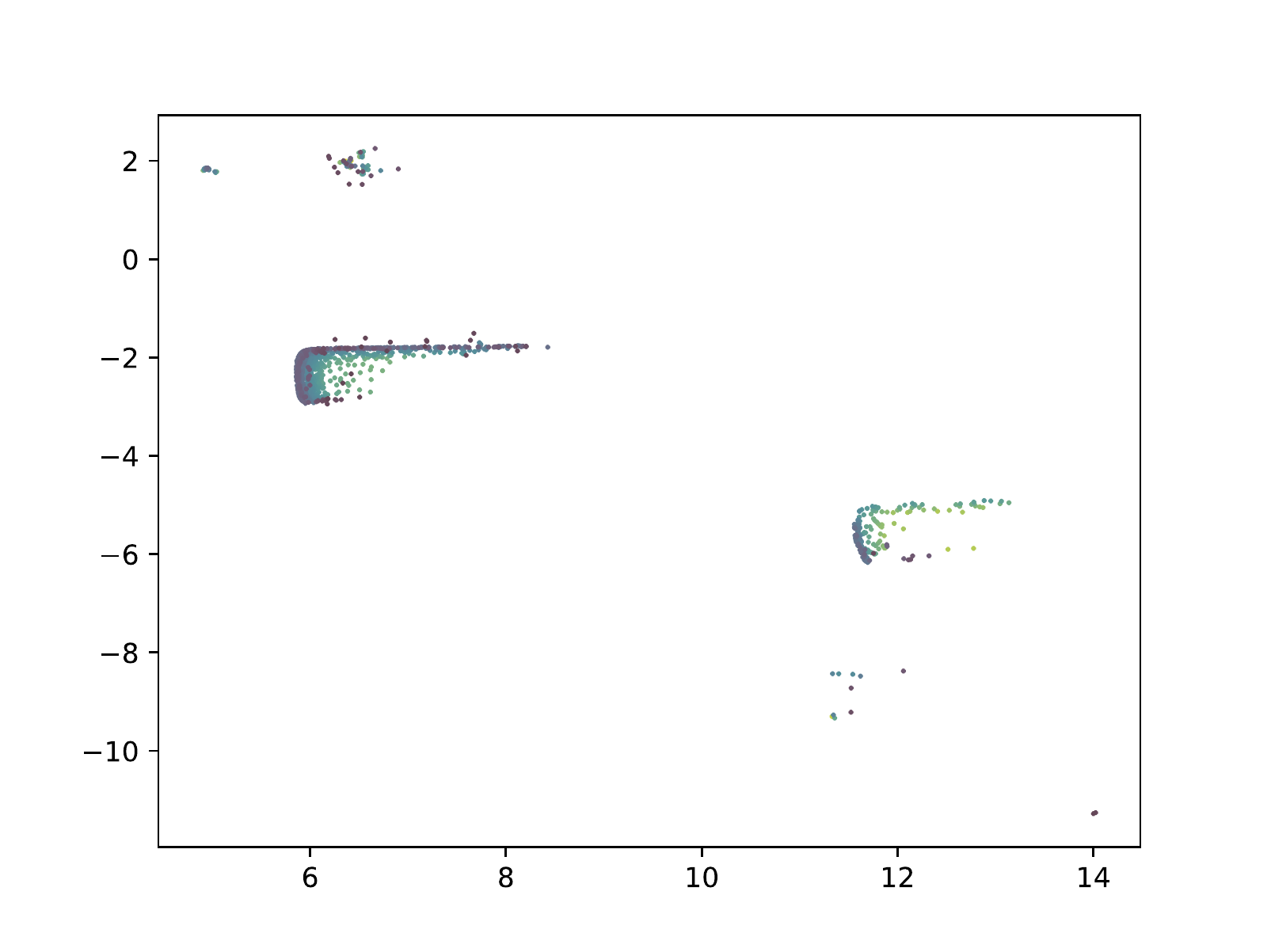}
        \caption{Example of KITTI without ground}
        \label{img:ng1}
    \end{subfigure}
    \begin{subfigure}{0.24\textwidth}
        \includegraphics[width=0.99\linewidth]{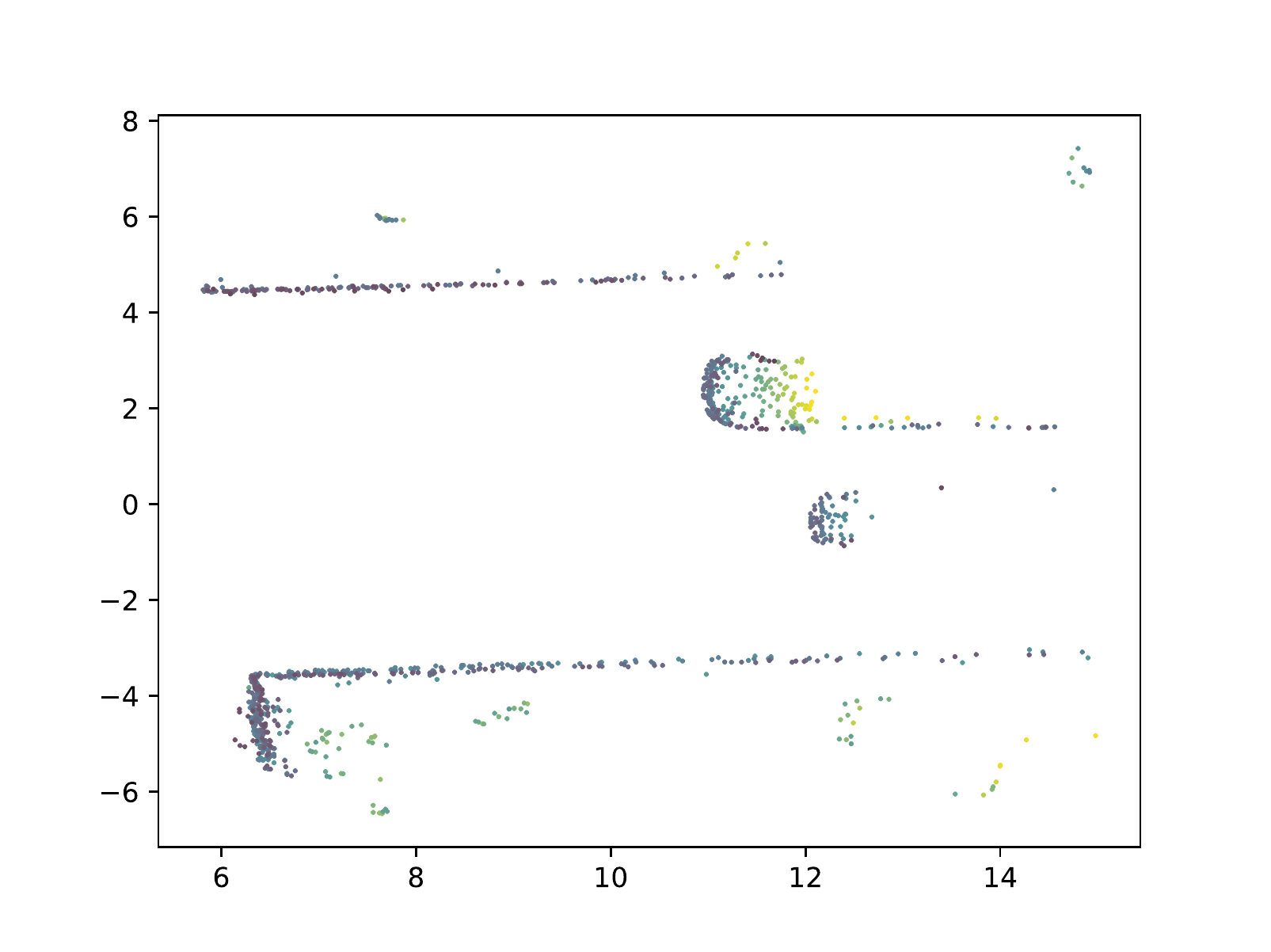}
        \caption{Example of KITTI without ground}
        \label{img:ng2}
    \end{subfigure}
    \caption{Examples of frames from KITTI. Height is used to color the points.}
    \label{img:kitti}
\end{figure}

In our experiments, we found it insightful to report results on post-processed scenes for ground removal. When the version of KITTI with no ground is used, it is explicitly mentioned. The post-processing was performed by the previous work \cite{FlowNet3D}. The algorithm was not made available, however, 150 scenes can be downloaded from the repository \cite{FlowNet3D_repo}. Figure~\ref{img:kitti} shows two examples of each version.

Each frame contains between 50k and 100k points. We sample 4096 points in a cube with sides of 30 meters centered at the LiDAR. We keep a held-out test set of 50 scenes, in line with previous research \cite{FlowNet3D, HPLFlowNet}. 

The main goal of KITTI is to evaluate a model or a method against real data. Failure modes present in synthetic datasets are also expected to be visible on real data as well. However, a well-performing model on synthetic datasets may still fall short on this real dataset. Insights drawn from simpler datasets can be used to make informative changes in a model or training method for further improvements in real test cases.


\section{Real Scenes Without Flow Annotations}

The missing piece in the benchmark is a large dataset of real data for self-supervised training. We call it \emph{Lyft}. It is a modified version of the original Lyft \cite{lyft2019}, which was not annotated for scene flow. Similarly to KITTI, the data was gathered by a LiDAR mounted on top of a car driven on urban and rural areas. A total of 22680 scenes are available, we judge that is enough data for training sophisticated models. Figure~\ref{img:multishapenet} shows two examples.

\begin{figure}
    \begin{subfigure}{0.49\textwidth}
        \includegraphics[width=0.99\linewidth]{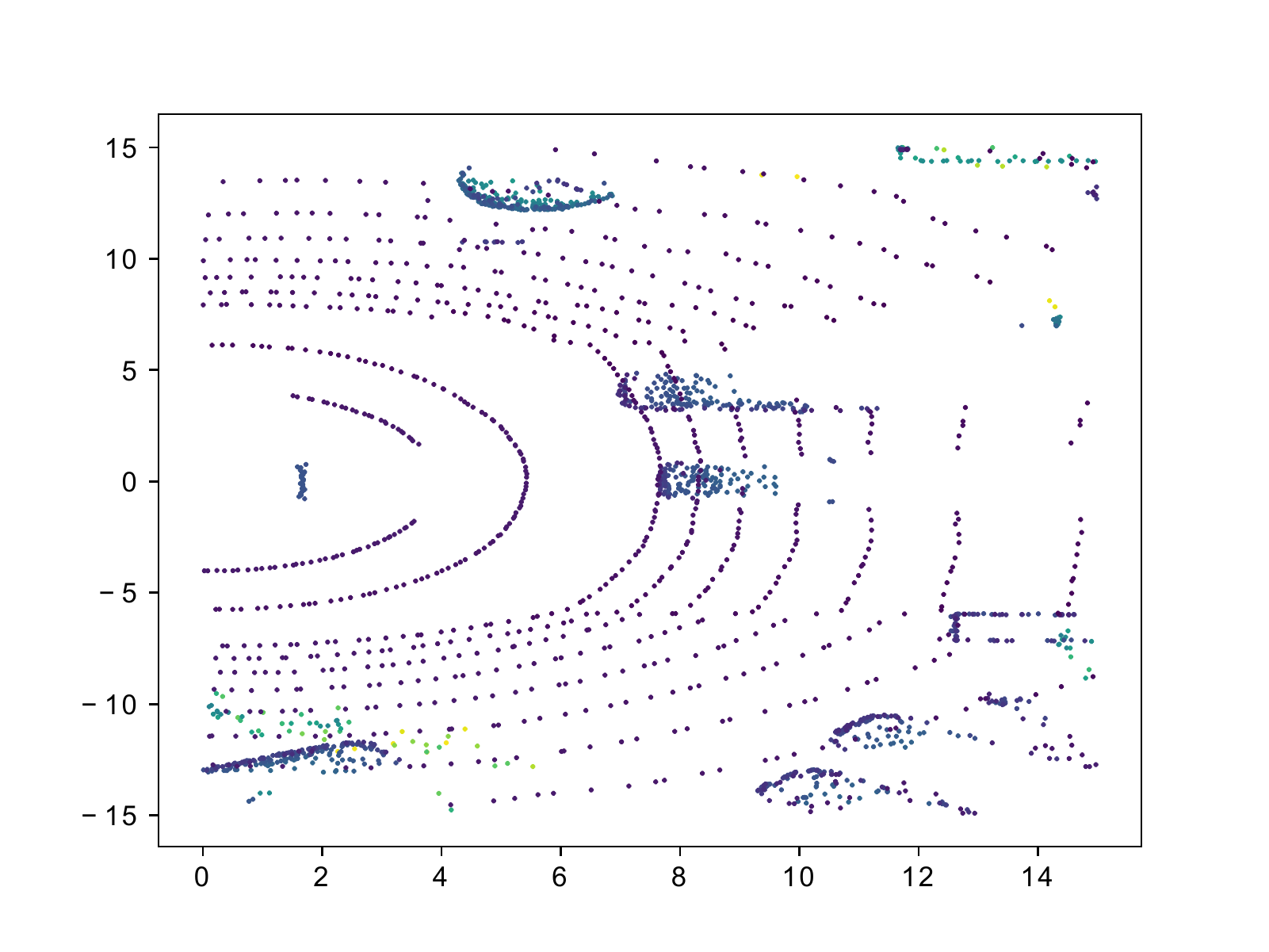}
        \caption{}
        \label{img:ly1}
    \end{subfigure}
    \begin{subfigure}{0.49\textwidth}
        \includegraphics[width=0.99\linewidth]{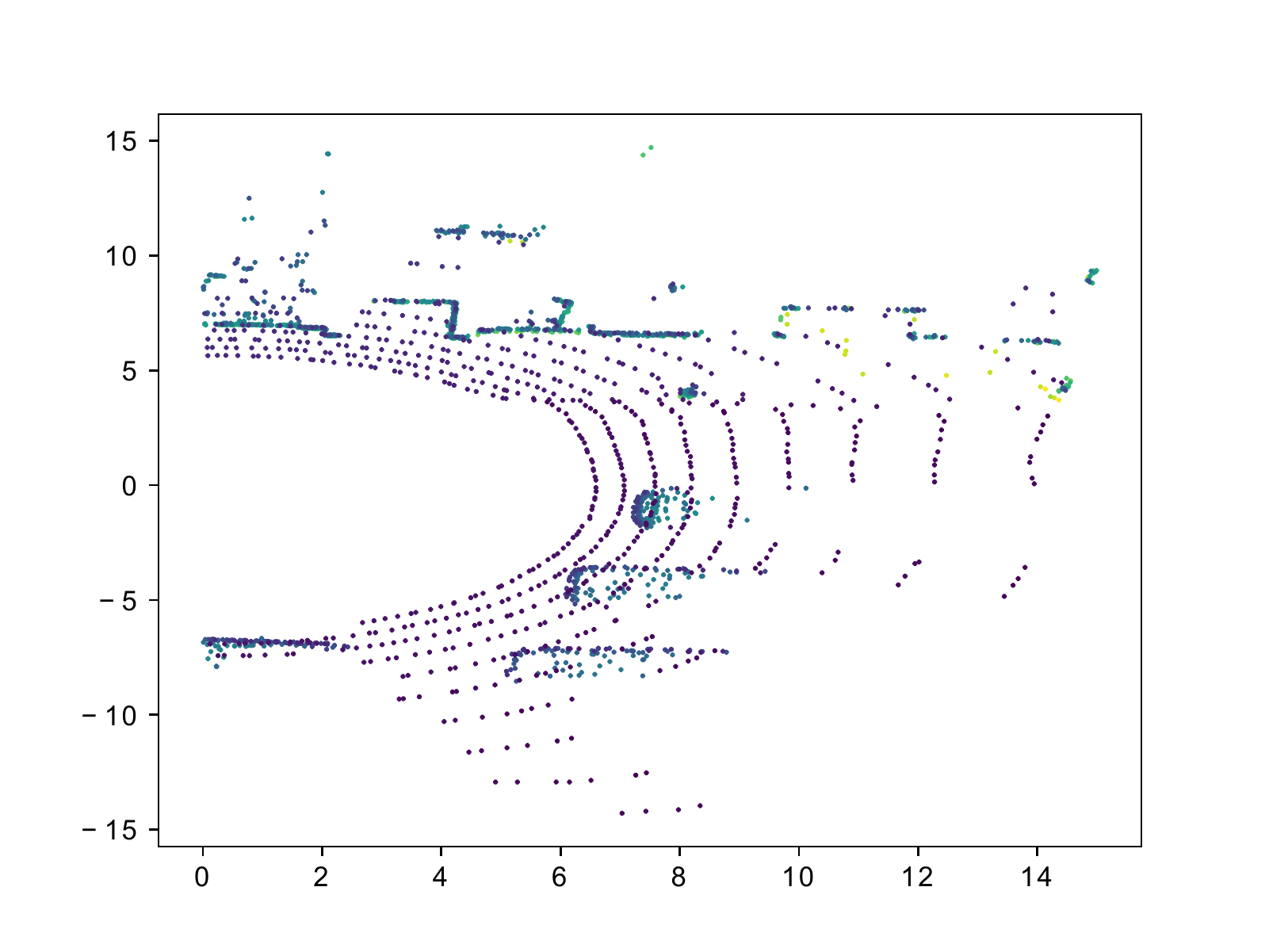}
        \caption{}
        \label{img:ly2}
    \end{subfigure}
    \caption{Examples of frames from Lyft. Height is used to color the points.}
    \label{img:lyft}
\end{figure}

We use the data collected by the LiDAR on top of the vehicle. The data collected by the lateral LiDARs and cameras are ignored. The points are sampled from a cube with sides of 30 meters and that is in front of the car ($x > 0$). This selection aims to reduce the domain shift from KITTI, used for evaluation, and Lyft, used for training. 

In our sandbox, the last step is to train a model using Lyft and test it on KITTI. The first step is to identify failure modes on the three synthetic datasets, where rapid prototyping and testing cycles are possible. When we are confident the training setup performs to our expectations on synthetic data, we may leap into training and testing it on real data. Lyft and KITTI complement each other in this last step, KITTI provides the flow annotations for evaluation and Lyft provides plenty of data for training.

%% file: src/results2.tex
\chapter{Experiments and Analysis}
\label{sec:experiments}

In this chapter, we use the Scene Flow Sandbox to gain an intuitive understanding of the performance of different setups and flow models. This chapter is organized as follows. First, we explain the evaluation metrics and introduce zEPE to make the comparison between the different datasets of our sandbox intuitive. We then compare how the Adversarial Metric Learning compares to other methods when used on real data. Following, we aim to build an intuition based on the Scene Flow Sandbox and on the types of insights we can draw from it. We identify recurrent failure modes of flow models and showcase how to bridge quantitative and qualitative results. Thirdly, we explore the use of Adversarial Metric Learning for scene flow estimation using three different flow models. We also perform an ablation study on its key components. Next, we investigate the usefulness of using nearest neighbors, as proposed by \cite{SelfSupervisedFlow}. Then, we move to investigate the impact of the Correspondence and Re-sampling mechanisms on the works of \cite{HPLFlowNet, PointPWCNet}. By the end of this section, we hope the reader will have a clear understanding of the advantages and limitations of the different approaches, including our own.


\section{Metrics}
\label{sec:metrics}

Flow estimation is a regression task. For each point in the scene, we are interested in regressing its flow vector as close to the ground truth as possible. The quality of the estimations are measured using the following metrics:

\emph{End Point Error} (EPE): the average Euclidean distance between the target and the estimated flow vectors. It is the main metric we are interested in improving. 

\emph{Accuracies}: the percentage of flow predictions that are below a threshold. The threshold has two criteria, achieving one of them is sufficient. In alignment with previous work we use the following \cite{FlowNet3D, FlowNet3D++, PointPWCNet, HPLFlowNet, SelfSupervisedFlow}:
    \begin{itemize}
        \item Acc 01: the prediction error is smaller than $0.1$ meter or $10\%$ of the norm of the target.  
        \item Acc 005: the prediction error is smaller than $0.05 $ meter or $5\%$ of the norm of the target.
    \end{itemize} 

\emph{zEPE}: we introduce the zEPE, it is the end point error normalized by the mean flow norm of the dataset. Table~\ref{tab:meannorm} shows the average of the flow norms for each dataset in the sandbox.

\begin{table}
    \centering
    \begin{tabular}{@{}ll@{}}
        \toprule
        Dataset         & Mean Norm [$m$] \\ \midrule
        Single ShapeNet & 0.4004          \\
        Multi ShapeNet  & 0.8827          \\
        FlyingThings3D  & 0.7595          \\
        KITTI           & 1.2514          \\
        KITTI No Ground & 0.9170          \\ \bottomrule
    \end{tabular}
    \caption{Mean norm of the flow vector of each dataset in the benchmark. A hypothetical model with zEPE = 1 on all datasets performs as well on Single ShapeNet as on Kitti, even though the EPE on those datasets is rather different. In a scene in KITTI most points belong to the ground, the mean norm is skwed to the average ego motion of the car. When the ground is removed, the mean norm is closer to the average motion of the dynamic objects in the scene.}
    \label{tab:meannorm}
\end{table}

The first three metrics - EPE and accuracies - are used to compare how different models perform on a fixed dataset. However, they offer little insight when the aim is to study how one specific model performs across different datasets. The zEPE was introduced to allow for a straight forward comparison of how a model performs in each dataset of our sandbox. Together, all four metrics are used for the quantitative evaluation of the experiments. 

\section{Scene Flow On Real Data}
\label{sec:bait}

The primary motivation of this work is to perform scene flow estimation on real data. In this section we compare our Adversarial Metric Learning method to the self-supervised methods proposed by \cite{SelfSupervisedFlow, PointPWCNet} and to a supervised baseline \cite{FlowNet3D}. 

We compare four different methods. We aim to make the results comparable, but without diverging significantly from what was originally proposed by \cite{SelfSupervisedFlow, PointPWCNet, FlowNet3D}. The supervised baseline was trained on FlyingThings3D and finetuned on KITTI with and without ground, as proposed by \cite{FlowNet3D}. The work of \cite{SelfSupervisedFlow, PointPWCNet} relied on the ground truth flow in different ways, as explained in Section~\ref{sec:relatedmech}. We show results that do not make use of target flow vectors at training time. The work of \cite{SelfSupervisedFlow} proposed a three step training, supervised training a FlowNet3D using FlyingThings3D, then self-supervised on a large non-annotated dataset such as \cite{lyft2019, nuscenes2019} and finally finetune the model on KITTI Scene Flow \cite{kitti}. We trained FlowNet3D from scratch on Lyft, using the losses proposed by \cite{SelfSupervisedFlow} and did not perform any finetuning step. The work of \cite{PointPWCNet} assumed the Correspondence Mechanism. They trained the PointPWC-net on a modified FlyingThings3D and reported results on KITTI. We trained the PointPWC-net on FlyingThings3D using the Re-sampling mechanism. All methods were tested on the test split of KITTI, with and without ground. We refer the reader to Appendix~\ref{apx:expsetup} for further experimental details.

Table~\ref{tab:sumresults} summarizes the results. We notice that the supervised training followed by finetuning is still state-of-the-art. Our method performs best among the self-supervised baselines. The gap between supervised and self-supervised, however, is still to be closed.

\begin{table}
    \centering
    \begin{tabular}{lllllll}
        \toprule
        Training Method & Flow Extractor                      & Dataset          & EPE    & zEPE    & Acc 01  & Acc 005 \\ \midrule
        Supervised \cite{FlowNet3D}      & FlowNet3D                           & KITTI            & \underline{0.1729} & \underline{0.1381} & \underline{57.68}\% & \underline{22.73}\% \\
        Self-Supervised \cite{SelfSupervisedFlow} & FlowNet3D & KITTI            & 1.0903 & 0.8712 & \textbf{9.81}\%  & \textbf{3.08}\% \\ 
        Self-Supervised \cite{PointPWCNet} & PointPWC-net     & KITTI            & 2.5717 & 2.0551 & 0.00\%  & 0.00\% \\
        Adversarial Metric & FlowNet3D                        & KITTI            & \textbf{0.9673} & \textbf{0.7729} & 3.01\%  & 0.76\% \\
        Adversarial Metric & PointPWC-net                     & KITTI            & 1.0497 & 0.8388 & 3.41\%  & 1.02\% \\ \midrule
        Supervised \cite{FlowNet3D}      & FlowNet3D                           & KITTI No Ground  & \underline{0.1880} & \underline{0.2050} & \underline{52.12}\% & \underline{22.81}\% \\
        Self-Supervised \cite{SelfSupervisedFlow} & FlowNet3D & KITTI No Ground  & 0.7002 & 0.7635 & 5.05\%  &  1.43\% \\ 
        Self-Supervised \cite{PointPWCNet} & PointPWC-net     & KITTI No Ground  & 1.4671 & 1.5998 & 0.03\%  & 0.00\% \\
        Adversarial Metric & FlowNet3D                        & KITTI No Ground  & 0.6733 & 0.7342 & \textbf{5.82}\%  & 1.03\% \\
        Adversarial Metric & PointPWC-net                     & KITTI No Ground  & \textbf{0.5542} & \textbf{0.6043} & 5.58\%  & \textbf{1.45}\% \\ \bottomrule
    \end{tabular}
    \caption{Comparison of different methods performing flow estimation on KITTI. The best metrics of self-supervised methods are reported in bold. The underlined metrics indicate the best overall performance, regardless of the training method.}
    \label{tab:sumresults}
\end{table}



The superior performance of the supervised method is not surprising. The flow models were trained to minimize the EPE directly. The self-supervised models had inferior performance, but are possible to train on datasets in which the flow targets are not available. In general, the self-supervised models perform better when the ground is removed. The ground is a large plane object that gives little information about motion. Even though, the ground was kept in the training data of the models trained on Lyft. 

Our Adversarial Metric Learning approach is the least impacted by the presence of the ground. We attribute that to the metric learned by the Cloud Embedder. As opposed to the nearest neighbor based distances, the Cloud Embedder may give different importance to points belonging to different objects. It may learn to regard points belonging to the floor as less informative than points belonging to moving objects. However, we have little more than the quantitative results to conjecture about the advantages and limitations of our proposed approach. 

At this stage, it is not straight forward to study the results of Table~\ref{tab:sumresults}. We may speculate about the limitations of each method, but we lack the tools to understand each one in-depth. The Scene Flow Sandbox was developed with this aim in mind. In the following sections we explore the benefits of the sandbox and use it to understand the limitations of our proposed approach and what could be the root causes.


\section{Exploring the Scene Flow Sandbox}
\label{subsec:baselines}

In this section, we show the usefulness of our Scene Flow Sandbox. It helps to bridge the qualitative and quantitative analysis. The insights yield the identification of five failure modes that will be used in the analysis of the following sections. 

The sandbox is useful as long as it facilitates drawing insights from flow estimations. To show the usefulness of our sandbox, we define five baselines for flow estimation. The first three are non-learning baselines used to expose recurrent failure modes. The last two baselines are supervised models later used to showcase insights taken from the failure modes on synthetic and real data.

\begin{itemize}
    \item Zero: estimate zero flow vectors regardless of the scene. A model that performs worse than just estimating zero flow vectors is worse than no flow estimation. 
    \item Average: the point clouds are reduced to their centroids, the estimated flow is the distance between the centroids. $\vec{f} = \frac{1}{|C_2|} \sum_{\vec{p} \in C_2} \vec{p} - \frac{1}{|C_1|} \sum_{\vec{p} \in C_1} \vec{p}$. Where $C_1$ and $C_2$ are consecutive point clouds and $|C_i|$ is the number of points in the point cloud. 
    \item KNN: the flow vector is the average distance between the point in $C_1$ and its k-nearest neighbors in $C_2$. In the experiments, we set $k = 1$ to highlight the variance of this approach.
    \item Segmenter: flow model trained with supervision. It is a PointNet Segmenter \cite{PointNet} used as a pointwise regressor. It was modified to receive sequences of point clouds. A temporal dimension is added to each point, corresponding to the frame they belong to. Then the consecutive point clouds are concatenated into one point cloud $C_{in} = [C_1, C_2]$. The forward pass of the Segmenter is performed on the entire point cloud $C_{in}$ and only flow corresponding to $C_1$ is outputted. 
    \item FlowNet3D: flow model from \cite{FlowNet3D} trained with supervision.
\end{itemize}

The non-learning baselines do not require training. We report their results on synthetic datasets and perform a failure mode analysis. The Segmenter and FlowNet3D were trained and tested on all synthetic datasets. We performed domain adaptation as described by \cite{FlowNet3D}, FlowNet3D was trained using FlyingThings3D and finetuned and tested on different data splits of KITTI. We refer the reader to Appendix~\ref{apx:expsetup} for more details of the experimental setup.

Table~\ref{tab:nlbaselines} shows how the different baselines perform on the synthetic datasets. The non-learning baselines perform rather poorly, the supervised baselines perform better, with FlowNet3D having the best scores of all. 

\begin{table}
    \centering
    \begin{tabular}{llllll} 
    \toprule
    Dataset         & Flow Estimation & EPE    & zEPE    & Acc 01  & Acc 005 \\ \midrule
    Single ShapeNet & Zero            & 0.4004 & 1.0     & 2.49\%  & 0.28\%  \\ 
    Single ShapeNet & Average         & 0.3191 & 0.7969  & 16.97\% & 5.05\%  \\
    Single ShapeNet & KNN             & 0.3097 & 0.7735  & 11.80\%  & 0.54\%  \\ 
    Single ShapeNet & Segmenter       & 0.1014 & 0.2532 & 55.24\% & 15.97\% \\
    Single ShapeNet & FlowNet3D       & 0.0567 & 0.1416 & 88.50\% & 54.02\% \\ \midrule
    Multi ShapeNet  & Zero            & 0.8827 & 1.0     & 0.22\%  & 0.02\%  \\
    Multi ShapeNet  & Average         & 0.8245 & 0.9341  & 0.18\%  & 0.02\%  \\
    Multi ShapeNet  & KNN             & 0.6455 & 0.7312 & 3.79\%  & 1.42\%  \\
    Multi ShapeNet  &  Segmenter      & 0.4948 & 0.5605 & 2.36\%  & 0.34\%  \\
    Multi ShapeNet  &  FlowNet3D      & 0.1524 & 0.1726 & 47.23\% & 15.02\% \\ \midrule
    FlyingThings3D  & Zero            & 0.7595 & 1.0     & 5.00\%  & 1.3\%   \\
    FlyingThings3D  & Average         & 1.0924 & 1.4404  & 0.95\%  & 0.12\%  \\
    FlyingThings3D  & KNN             & 0.7010 & 0.9249  & 5.92\%  & 1.16\%  \\ 
    FlyingThings3D  &  Segmenter      & 0.6786 & 0.8935 & 3.36\%  & 0.44\%  \\
    FlyingThings3D  &  FlowNet3D      & 0.1838 & 0.2420 & 61.58\% & 31.94\% \\ \bottomrule 
    \end{tabular}
    \caption{Performance of the baselines on the synthetic datasets.}
    \label{tab:nlbaselines}
\end{table}

\begin{figure}
    \centering
    \includegraphics[width = 1\linewidth]{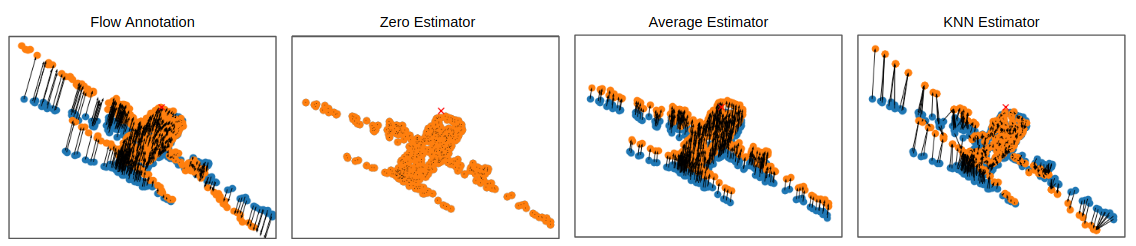}
    \caption{Example taken from Single ShapeNet. The 2D projection shows the Ground Truth flow vectors and the estimations made by the different baselines. The blue points correspond to time $t$ and orange points to time $t+1$.}
    \label{img:nlshapenet}
\end{figure}

\begin{figure}
    \centering
    \includegraphics[width = 1\linewidth]{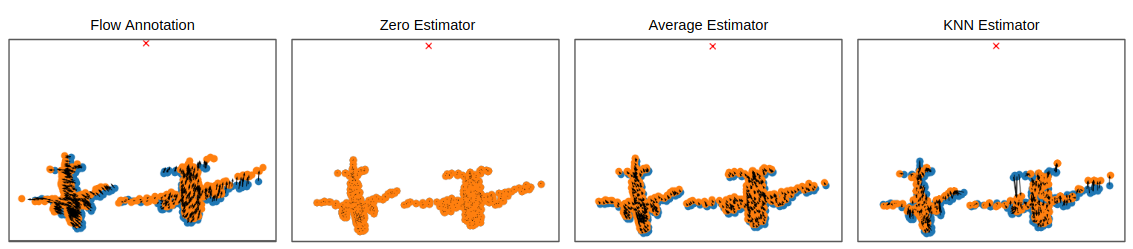}
    \caption{Example taken from Multi ShapeNet. The 2D projection shows the Ground Truth flow vectors and the estimations made by the different baselines. The blue points correspond to time $t$ and orange points to time $t+1$.}
    \label{img:nlmultshapenet}
\end{figure}

\begin{figure}
    \centering
    \includegraphics[width = 1\linewidth]{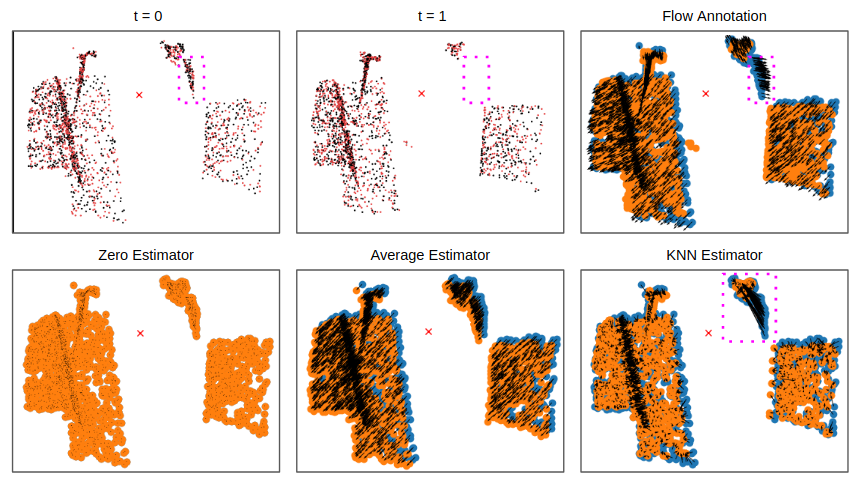}
    \caption{Example taken from FlyingThings3D. The 2D projection is chosen to make occlusions evident. The pink dotted line highlights an object that was present in time $t=0$ and was occluded at time $t=1$. The blue points correspond to time $t$ and orange points to time $t+1$.}
    \label{img:nlfly}
\end{figure}

There are several failure modes behind the metrics shown in Table~\ref{tab:nlbaselines}. We use our sandbox and the non-learning baselines to identify five of them:

\begin{enumerate}
    \item Failure to capture motion. Models that show this failure mode are too simplistic for the considered scene. Examples of this failure mode are shown in Figure~\ref{img:nlshapenet}, the Zero estimator does not have any tools to capture motion. In fact, any model with a zEPE close to 1 performs about as poorly as the Zero estimator. The EPEs shown in Table~\ref{tab:nlbaselines} are the average flow norms shown in Table~\ref{tab:meannorm}. A zEPE close to one is a strong indicator the model does not have the means to capture the motion of the scene.
    \item Underestimate flow norms. A model does estimate motion, but it is too simplistic to grasp the full translation, rotation, and scaling of objects. The Average estimator displays such failure mode in its estimations. For instance, in Figure~\ref{img:nlshapenet} the rotation of the airplane is completely ignored.  
    \item Locally inconsistent flow vectors. Noisy estimates may mean the model does not coherently capture how an object moves. Locally inconsistent flow vectors tend to deform the object. That can be observed in the flow estimations of the KNN estimator. Rather inconsistent flow vectors are visible in Figure~\ref{img:nlshapenet}, however, the global motion of the airplane is somewhat kept. 
    \item Failure to capture locally coherent and globally independent motion. Points that belong to one object will display a coherent movement, whereas points that belong to different objects move independently. For instance, Figure~\ref{img:nlmultshapenet} shows two airplanes from Multi ShapeNet moving in different directions. Both estimators Average and KNN fail to capture different aspects of this scene. The KNN estimator fails to estimate locally consistent flow vectors, but it does capture globally independent motion. The opposite happens with the Average estimator. Its estimations are locally consistent, though it has no means to estimate globally independent motion.
    \item Confusion caused by partially observable scenes. The model fails to learn priors and features that compensate for self-occlusions, occlusions, and objects leaving the field of view. An example of such failure mode is seen in Figure~\ref{img:nlfly}. The pink dotted line highlights an object that was visible at time $t=0$ and was occluded at time $t=1$. The KNN estimator performs point assignments, so it estimates that all points were merged to the nearest object. 
\end{enumerate}

Our sandbox helps us bridge the qualitative and quantitative analysis. The identification of the five aforementioned failure modes was fairly simple when done in a dataset that was just simple enough to highlight the failure mode. Single ShapeNet, the simplest dataset in our sandbox, was already enough to help us spot two failure modes. Such an analysis would have been much harder if we were only using FlyingThings3D. Now that we have criteria against which we can objectively analyze our models, we shall move to the supervised baselines. 

\begin{figure}
    \begin{subfigure}{0.5\textwidth}
        \includegraphics[width = 1\linewidth]{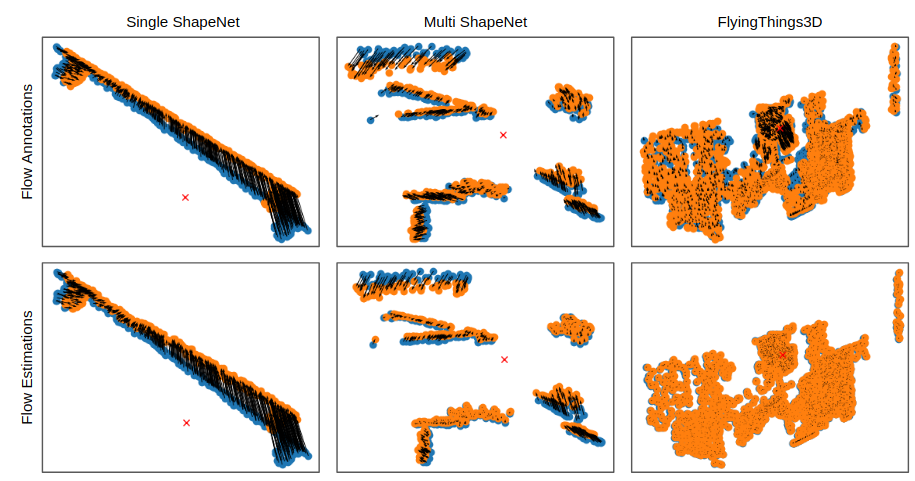}
        \caption{Qualitative results from Segmenter.}
        \label{img:supseg}
    \end{subfigure}
    \begin{subfigure}{0.5\textwidth}
        \includegraphics[width = 1\linewidth]{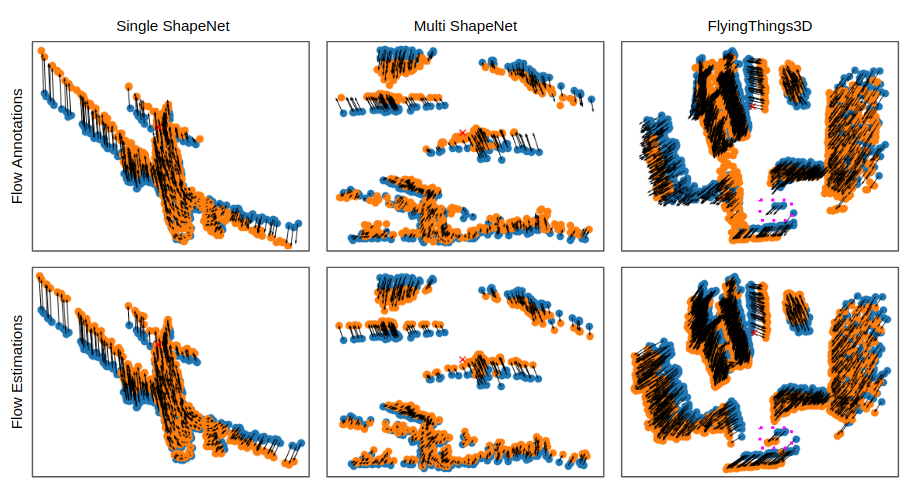}
        \caption{Qualitative results from FlowNet3D.}
        \label{img:supfn3d}
    \end{subfigure}
    \caption{Examples of Segmenter and FlowNet3D on the synthetic datasets.}
    \label{img:supbaselines}
\end{figure}

We use the failure modes to draw links between the quantitative results of Table~\ref{tab:nlbaselines} and the qualitative results in Figure~\ref{img:supbaselines}.

It is noticeable in Figure~\ref{img:supseg} how the performance of the Segmenter degrades with the increase in the complexity of the scene. The zEPEs of the model for the different datasets confirm this finding. The model performs rather well on Single ShapeNet. On Multi ShapeNet we identify the failure mode, the model underestimates the norm of the flow vectors, as previously noticed with the Average estimator. For FlyingThings3D, this underestimation is even more extreme. We attribute that to the lack of complexity of the model, which made it biased towards small flow vectors. A model with high bias is usually under parametrized for the task \cite{bishop2007}. We need a model that is sophisticated enough to capture the complexity of scenes from Multi ShapeNet and FlyingThings3D.

FlowNet3D has enough modeling capabilities for the synthetic datasets, as seen in Figure~\ref{img:supfn3d} and confirmed by Table~\ref{tab:nlbaselines}. We do not spot any of the failure modes on the displayed examples. The pink dotted line shows one example of occlusion. The object is present on the first frame (blue dots) but is occluded on the second frame (no orange dots in the expected position). The model, however, makes a rather accurate guess of the movement of the object. 

\begin{figure}
    \begin{subfigure}{0.5\textwidth}
        \includegraphics[width=0.99\linewidth]{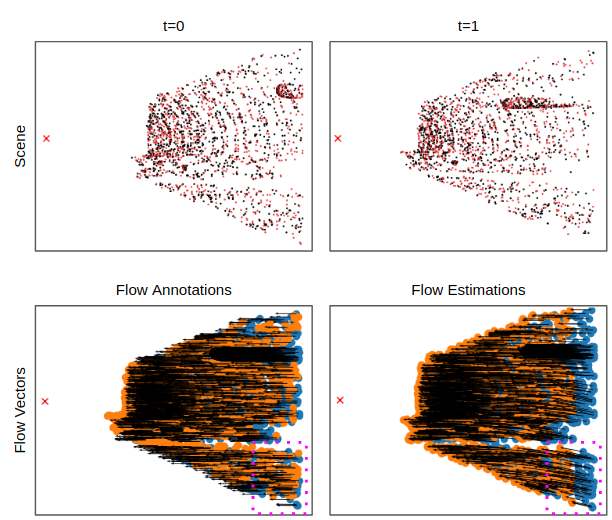}
        \caption[width=0.9\linewidth]{Qualiative results from FlowNet3D trained on FlyingThings3D, finetuneed and tested on KITTI with ground. The pink doted line shows a region where the FlowNet3D produces locally inconsistent vectors that deform the object.}
        \label{img:fn3dkitti}
    \end{subfigure}
        \begin{subfigure}{0.5\textwidth}
        \includegraphics[width=0.99\linewidth]{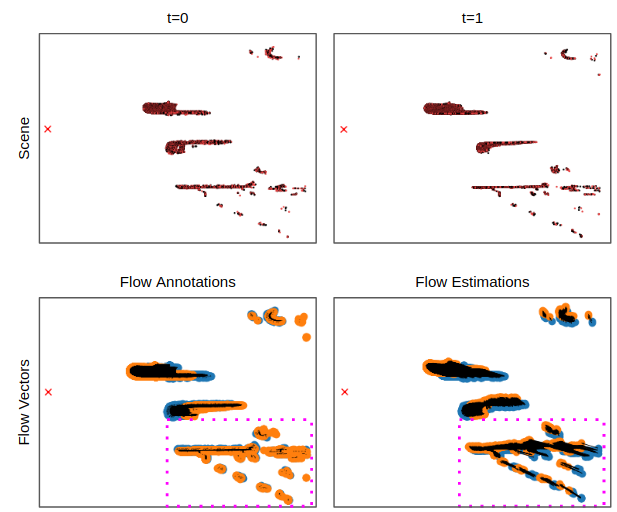}
        \caption[width=0.9\linewidth]{Qualiative results from FlowNet3D trained on FlyingThings3D, finetuneed and tested on KITTI without ground. The pink doted line highlights the failure mode of the scene. }
        \label{img:fn3dkitting}
    \end{subfigure}
    \caption{Qualitative results from FlowNet3D after domain adaptation.}
    \label{img:fn3dkittis}
\end{figure}

Finally, we evaluate how FlowNet3D performs on real data. Table~\ref{tab:supervisedkitti} shows the results. The finetuned FlowNet3D performs quite well on both versions of KITTI. Figure~\ref{img:fn3dkittis} shows examples from the two versions of KITTI. The pink dotted lines mark failure mode in the flow estimation. 


\begin{table}
    \centering
    \begin{tabular}{llllll}
        \toprule
        Dataset  & EPE    & zEPE & Acc 01  & Acc 005 \\ \midrule
        KITTI    & 0.1729 & 0.1381 & 57.68\% & 22.73\% \\
        \begin{tabular}[c]{@{}l@{}}KITTI \\ No Groud\end{tabular}  & 0.1880 & 0.2050 & 52.12\% & 22.81\% \\ \bottomrule
    \end{tabular}
    \caption{FlowNet3D pretrained on FlyingThings3D, finetuned and evaluated on KITTI.}
    \label{tab:supervisedkitti}
\end{table}

Understanding the failure modes on simple synthetic data helps us interpret the results from real data. Figure~\ref{img:fn3dkitti} shows the top projection of a scene from KITTI. Most points belong to static structures such as the ground and rail-guards. The motion of all those points is given by the ego-motion of the car. The flow annotations indeed show most points moving in the same direction. FlowNet3D displays locally inconsistent vectors that deform the ground marked by the pink dotted lines. A large part of the ground leaves the field of view in-between frames and another similar part of the ground enters the field of view. This pattern probably confuses the model.

A different failure mode is encountered when the ground is removed. Figure~\ref{img:fn3dkitting} shows a case where FlowNet3D failed to grasp the individual movement of different objects. The failure mode was seen before with the Average estimator, the flow vectors are similar in direction and magnitude with little consideration to the different objects they belong to.

This section was aimed to show the usefulness of the sandbox in debugging flow models. We used qualitative results to explain quantitative ones, this process was facilitated by the identification of five failure modes. If a model displays a failure mode in a simple dataset, it will likely display the same failure mode on more complex datasets. Thus, the sandbox allows for fast iteration on simple datasets before experimenting on more complex ones. 


\section{Adversarial Metric Learning}
\label{subsec:aml}

In Section~\ref{sec:modelchoices} we argued in favor of a generative modeling approach to scene flow estimation. We proposed tackling the task with adversarial training. In this section, we show the performed experiments and the results. We continue to use the sandbox to interpret the metrics.

In our approach, we replace the KNN-based loss by a distance metric between point clouds that is learned. The Cloud Embedder is a learnable module that maps a point cloud to a latent vector of fixed size. During training we aim to minimize the losses described by Equation~\ref{eq:ce} and Equation~\ref{eq:fe} simultaneously. Our setup is not dependent on any particular model, we performed experiments using three different architectures, the Segmenter, FlowNet3D, and PointPWC-net. The latter outputs a pyramid of flow estimations, from coarse to fine. To keep our experiments consistent, we only used the finest output. Further experimental details are documented in Appendix~\ref{apx:expsetup}. 

The results are summarized in Table~\ref{tab:advresults}. FlowNet3D performs the best on Single ShapeNet, but PointPWC-net has the lead on Multi ShapeNet and FlyingThings3D. 

The Segmenter did learn flow estimation for the simple scenes of Single ShapeNet. However, its zEPE close to $1$ on more complex datasets shows it has not learned useful features for flow estimation on Multi ShapeNet and FlyingThings3D. Just as it was the case for supervised training, the training setup is not necessarily the problem, but rather the level of complexity of the model. The Segmenter is ill-equipped to learn flow from complex scenes. Therefore we will focus on the failure mode analysis on FlowNet3D and PointPWC-net.

\begin{table}
    \centering
    \begin{tabular}{llllll}
    \toprule
    Flow Extractor     & Dataset            & EPE    & zEPE    & Acc 01  & Acc 005 \\ \midrule
    Segmenter          & Single ShapeNet    & 0.1600 & 0.3996 & 27.58\% & 4.71\%  \\
    Segmenter          & Multi ShapeNet     & 0.8098 & 0.9131 & 0.40\% & 0.00\%  \\
    Segmenter          & FlyingThings3D     & 0.7548 & 0.9938 & 4.85\% & 0.01\%  \\ \midrule
    FlowNet3D          & Single ShapeNet    & \textbf{0.1287} & \textbf{0.3214} & \textbf{50.03}\% & \textbf{16.64}\% \\
    FlowNet3D          & Multi ShapeNet     & 0.3497 & 0.3961 & 10.27\% & 1.78\% \\
    FlowNet3D          & FlyingThings3D     & 0.5629 & 0.7411 & 5.52\%  & 0.89\% \\ \midrule
    PointPWC-net       & Single ShapeNet    & 0.1824 & 0.4555 & 28.27\% & 7.55\% \\
    PointPWC-net       & Multi ShapeNet     & \textbf{0.2920} & \textbf{0.3308} & \textbf{15.31}\% & \textbf{3.00}\% \\
    PointPWC-net       & FlyingThings3D     & \textbf{0.5270} & \textbf{0.6939} & \textbf{7.63}\%  & \textbf{1.30}\% \\ \bottomrule
    \end{tabular}
    \caption{Flow Extractor trained with Adversarial Metric Learning and Cycle Consistency. Bold number show the best performing model per dataset.}
    \label{tab:advresults}
\end{table}

\begin{figure}
    \begin{subfigure}{0.5\textwidth}
        \includegraphics[width=0.99\linewidth]{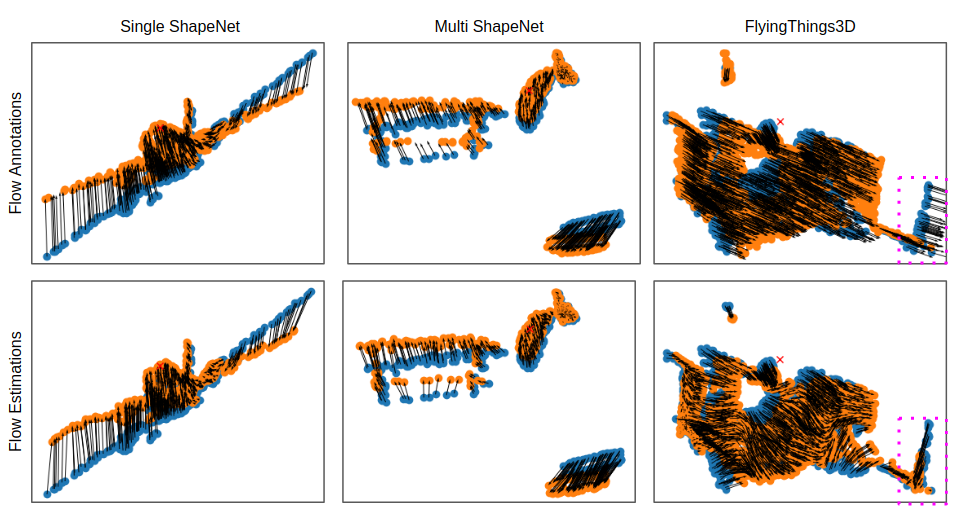}
        \caption{FlowNet3D used as flow extractor. The pink doted line shows an example of occlusion.}
        \label{img:fn3dmst}
    \end{subfigure}
    \begin{subfigure}{0.5\textwidth}
        \includegraphics[width = .99\linewidth]{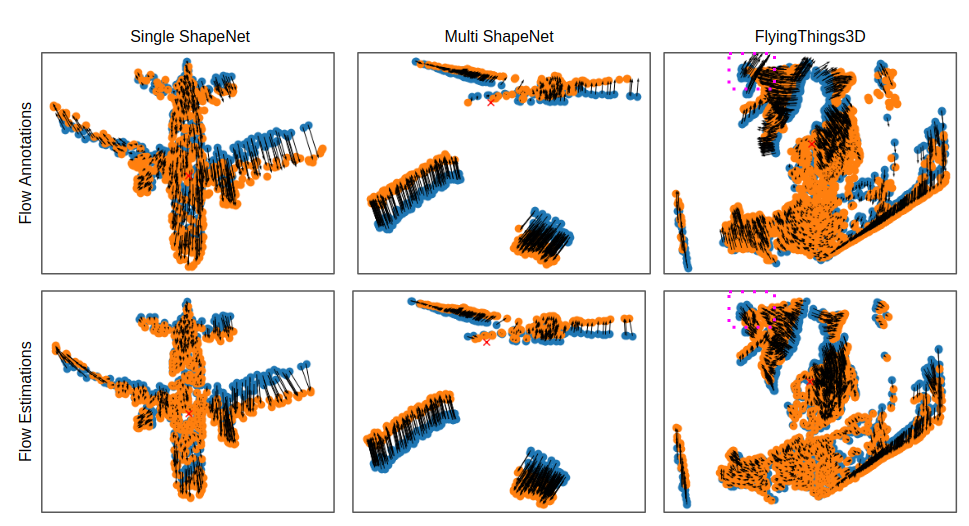}
        \caption{PointPWC-net used as flow extractor.}
        \label{img:ppwcmst}
    \end{subfigure}
    \caption{Qualitative results from Adversarial Metric Learning on Single ShapeNet, Multi ShapeNet and FlyingThings3D. The pink doted lines shows examples of occlusions that are not correctly handled by the flow extractors.}
    \label{img:mstresults}
\end{figure}

Figure~\ref{img:fn3dmst} shows qualitative results for FlowNet3D. In all the three datasets, we notice that FlowNet3D  has problems with locally consistent flow vectors - second failure mode. It is visible for Single ShapeNet and Multi ShapeNet, that the macro geometry of the objects is kept, only with localized deformations. It indicates that the model has learned to grasps how different objects have independent movement, even with slightly locally inconsistent flow vectors. If we move our analysis to FlyingThings3D, we will notice the flow estimations deviate visibly from the ground truth. The dotted pink line shows an object leaving the field of view. The failure mode we observed with KNN, that the points are assigned to the nearest object, is also present here. The rather high zEPE on FlyingThing3D and this insight point that occlusions are particularly problematic to learn. 

Figure~\ref{img:ppwcmst} shows qualitative results for PointPWC-net. On Single ShapeNet the flow vectors are locally consistent, but the macro geometry is not well kept. There are visible sections of the airplane that have slightly different directions or magnitude of movement. It is as if the architecture fragments the object into sections and estimates the motion of each one independently. This behavior may explain why the zEPE of this model is lower for Multi ShapeNet than for Single ShapeNet. The objects at Multi ShapeNet enjoy better local and macro flow consistency. PointPWC-net shows more consistent results, with the ground truth of FlyingThings3D. The magnitude of the predictions is consistently smaller, but the rough directions are kept. The pink dotted lines show an object that left the field of view. Just as FlowNet3D, its flow estimations point to the nearest object. 

Neither architecture overcame the difficulties added by occlusions and by objects leaving the field of view. In fact, the failure mode is very similar to what we observed with the KNN estimator. That was not in line with our expectations. We expected the models to guess a would-be location for that object without merging it into another one close by. In Figure~\ref{img:supfn3d} we highlighted that FlowNet3D is able to guess the motion of an occluded object. It is possible that our setup does not allow for the Flow Extractor to guess the motion of occluded objects. In our analysis, we found that occlusions are particularly critical. We conjecture that the Cloud Embedder uses missing objects as an easy proxy to distinguish between the real point cloud and the estimated one. The Flow Extractor, however, has no tool at its disposal to make objects appear or disappear, it just estimates flow vectors for every point in the scene. Thus, the Flow Extractor is not capable of fooling the Cloud Embedder in scenes where occlusions are present. 

So far we have focused our analysis on the synthetic datasets of the sandbox. We are also interested to evaluate our setup on real data. The training was done using scenes from Lyft and the evaluation performed using both versions of KITTI. We refer the reader to Appendix~\ref{apx:expsetup} for more experimental details.

The results are summarized by Table~\ref{tab:advresultskitti}. Both models show poor quantitative results on KITTI with ground and slightly better results on KITTI without ground. 

\begin{figure}
    \begin{subfigure}{0.5\textwidth}
        \includegraphics[width=0.99\linewidth]{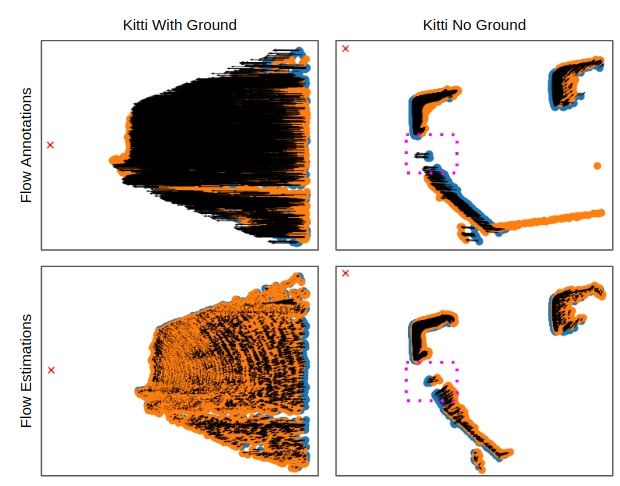}
        \caption{FlowNet3D used as flow extractor.}
        \label{img:fn3dmstkitti}
    \end{subfigure}
    \begin{subfigure}{0.5\textwidth}
        \includegraphics[width = .99\linewidth]{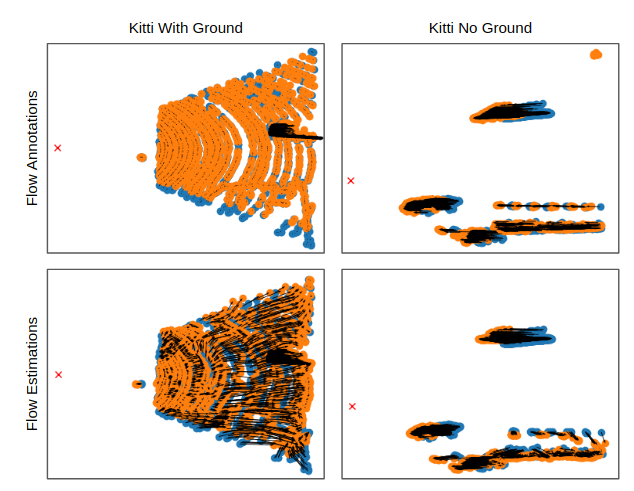}
        \caption{PointPWC-net used as flow extractor.}
        \label{img:ppwcmstkitti}
    \end{subfigure}
    \caption{Qualitative results from Adversarial Metric Learning on KITTI with and without ground.}
    \label{img:mstresultskitti}
\end{figure}

The results of Table~\ref{tab:advresultskitti} are not surprising. In Figure~\ref{img:mstresultskitti} it is observable that the severe occlusions present in the scenes seem to pose as a major barrier for further improvements for both models. That is in line with the failure modes observed on FlyingThings3D.

FlowNet3D performs slightly better on KITTI without ground. The $\sim 5\%$ difference on zEPE on Table~\ref{tab:advresultskitti} translates to estimations that are visually closer to the flow targets, as shown in Figure~\ref{img:fn3dmstkitti}. The flow estimations for KITTI with ground display a strong local and global inconsistency. Whereas the estimations for KITTI without ground are locally consistent. In Figure~\ref{img:fn3dmstkitti} the pink dotted line shows a case where FlowNet3D guessed the movement of an occluded object. The estimated flow is contrary to the real movement of the object. Still, this indicates that FlowNet3D was able to use the movement of different objects as cues to guess the flow vectors of the occluded object. 

PointPWC-net performs significantly better on KITTI without ground than on KITTI with ground, $\sim 30\%$ lower zEPE on Table~\ref{tab:advresultskitti}. In Figure~\ref{img:ppwcmstkitti} the scene of KITTI with ground has one moving object (possibly a car) and all the rest is stationary. PointPWC-net displays the third failure mode, it does not capture the independent motion of the different objects in the scene. Namely the pavement and the car. The flow predictions also lack local consistency. That is improved when the evaluation is performed on KITTI without ground. The flow vectors are locally consistent and the motion of different objects is seemingly independent.

\begin{table}
    \centering
    \begin{tabular}{llllll}
        \toprule
        Flow Extractor & Dataset            & EPE    & zEPE    & Acc 01  & Acc 005 \\ \midrule
        FlowNet3D      & KITTI              & 0.9673 & 0.7729 & 3.01\% & 0.76\% \\
        FlowNet3D      & KITTI No Ground    & 0.6733 & 0.7342 & 5.82\% & 1.03\% \\
        PointPWC-net   & KITTI              & 1.0497 & 0.8388 & 3.41\% & 1.02\% \\
        PointPWC-net   & KITTI No Ground    & 0.5542 & 0.6043 & 5.58\% & 1.45\% \\ \bottomrule
    \end{tabular}
    \caption{Flow Extractor trained with Adversarial Metric Learning and Cycle Consistency. Models trained on Lyft and tested on both versions of KITTI.}
    \label{tab:advresultskitti}
\end{table}

Overall, Adversarial Metric learning can be used to train different Flow Extractors on performing scene flow estimation. The Flow Extractors, however, are not expected to perform well in partially observable scenes. The result makes sense in hindsight. The Flow Extractor is encouraged to approximate the target point cloud as much as possible. However, it has no means to perform occlusions, thus mapping the points to a nearby object may approximate the target point cloud better, but it results in poor flow estimations. The Cloud Embedder, on the other hand, can easily distinguish between the target and the predicted point cloud by simply spotting missing objects. Further improvements require handling partially observable scenes, which we leave for future work. In the following sections we perform ablation studies on two components we found critical to our proposed approach.


\subsection{Contribution of Cycle Consistency Loss}

The Cycle Consistency Loss compares the forward and backward flow estimations. The loss is minimized when the backward flow cancels the forward flow. Yet, it does not necessarily mean the estimations are correct. For instance, estimating zero flow vectors will minimize the Cycle Consistency loss, but do not necessarily correspond to the motion of the scene. Intuitively, however, it may induce the model to perform estimations that are locally and globally consistent. 
We performed experiments using the different auxiliary Cycle Consistency losses to our Adversarial Metric Learning setup. 

We considered five alternatives to Cycle Consistency loss, MSE, L2, cosine similarity (Cos), and the combinations MSE with cosine similarity, and L2 with cosine similarity. We compare the five alternatives to not using the Cycle Consistency (None). We report results on Single and Multi ShapeNet.

\begin{table}
    \centering
    \begin{tabular}{llllll}
        \toprule
        Cycle Consistency & Dataset             & EPE    & zEPE   & Acc 01  & Acc 005 \\ \midrule
        None              & Single ShapeNet     & 0.2391 & 0.5971 & 15.14\% & 2.27\%  \\
        Cos               & Single ShapeNet     & 0.2187 & 0.5462 & 20.24\% & 4.75\%  \\
        MSE               & Single ShapeNet     & 0.2271 & 0.5671 & 18.38\% & 3.81\%  \\
        L2                & Single ShapeNet     & 0.3912 & 0.9770 & 2.73\% & 0.35\%  \\
        Cos + MSE         & Single ShapeNet     & 0.2167 & 0.5412 & 20.48\% & 4.84\%  \\
        Cos + L2          & Single ShapeNet     & \textbf{0.1287} & \textbf{0.3214} & \textbf{50.03}\% & \textbf{16.64}\% \\ \midrule
        None              & Multi ShapeNet      & 0.4920 & 0.5573 & 2.85\%  & 0.39\%  \\
        Cos               & Multi ShapeNet      & 0.4302 & 0.4873 & 5.44\%  & 0.83\%  \\
        MSE               & Multi ShapeNet      & 0.4405 & 0.4990 & 4.37\%  & 0.63\%  \\
        L2                & Multi ShapeNet      & 0.3786 & 0.4289 & 12.67\%  & 3.02\%  \\
        Cos + MSE         & Multi ShapeNet      & 0.4200 & 0.4758 & 6.56\%  & 1.08\%  \\
        Cos + L2          & Multi ShapeNet      & \textbf{0.3497} & \textbf{0.3961} & \textbf{10.27}\% & \textbf{1.78}\% \\ \bottomrule
    \end{tabular}
    \caption{Ablation studies on FlowNet3D trained with different auxiliary cycle consistency losses. The impact of different losses is quite expressive.}
    \label{tab:ablation_cycleloss}
\end{table}

Table~\ref{tab:ablation_cycleloss} shows the results of the ablation study. We see that the best results were achieved by using cosine similarity and L2 norm. On the other side of the spectrum, the worst results happened by not using Cycle Consistency.

This result is in line with previous supervised learning approaches, where the L2 norm is the most common loss used. When compared to the MSE loss, we see the L2 loss improves the results by a large margin. We attribute to that the fact that L2 loss penalizes outliers less. 

Unstable optimization is a common problem in adversarial training. Instead of reaching the Nash equilibrium, the networks are trapped in a local minimum that keeps them from improving. Enforcing cycle consistency improves the overall metrics, but it also played an important role in keeping the training stable. We stopped experiencing training collapse after introducing the Cycle Consistency loss.

The Cycle Consistency loss shows two-fold benefits in our experiments. It induces the model to perform locally consistent flow and it improves the stability of the adversarial setup. 


\subsection{Contribution of Multi-Scale Triplet Loss}

For training the Flow Extractor and the Cloud Embedder, we make use of the Multi-Scale Triplet Loss. The Cloud Embedder maps an entire point Cloud into a latent vector. The hypothesis was that the Cloud Embedder would have to distinguish between the ground truth $C_2$ and the estimated point cloud $C_1 + F$. The feedback to the Flow Extractor regards the positions of the points of $C_1 + F$, as well as global and local geometry. 

In the experiments, we use FlowNet3D as the Flow Extractor. All hyper-parameters are fixed, except for the scaling factor. We compare the use of the last feature vector (scaling is set to zero) against three methods for scaling the intermediary feature vectors.  For all three methods, the scaling of the last feature vector is $1$. In all three the shallowest layer is the most down-scaled. The results are reported on Multi ShapeNet and FlyingThings3D. \textbf{}

\begin{table}
    \centering
    \begin{tabular}{lllll}
        \toprule
        Dataset        & Scaling factor             & EPE    & Acc 01  & Acc 005 \\ \midrule
        Multi ShapeNet & Zero                       & 0.4043 & 5.60\%  & 0.76\%  \\
        Multi ShapeNet & $\frac{1}{\sqrt{l + 1}}$   & \textbf{0.3497} & \textbf{15.27}\% & \textbf{1.76}\%  \\
        Multi ShapeNet & $\frac{1}{l + 1}$          & 0.3850 & 7.57\%  & 1.12\%  \\
        Multi ShapeNet & $\frac{1}{(l + 1)^2}$      & 0.4137 & 7.33\%  & 1.23\%  \\ \midrule
        FlyingThings3D & Zero                       & 0.6500 & 4.41\%  & 0.5\%   \\
        FlyingThings3D & $\frac{1}{\sqrt{l + 1}}$   & 0.6514 & 5.81\%  & 0.9\%   \\
        FlyingThings3D & $\frac{1}{l + 1}$          & \textbf{0.5629} & 5.52\%  & 0.89\%  \\
        FlyingThings3D & $\frac{1}{(l + 1)^2}$      & 0.6395 & \textbf{6.54}\%  & \textbf{1.12}\%  \\ \bottomrule
    \end{tabular}
    \caption{FlowNet3D trained on Adversarial Metric using different scaling factors for the Multi-Scale Triplet loss. The scaling factors are functions that use the level $l$ of the activations. The activations are organized from last to first. Zero only used the last activation.}
    \label{tab:mstloss}
\end{table}

Table~\ref{tab:mstloss} shows the results. The use of the correct scaling method brings at least $10\%$ improvement when compared to the second-best. The combination of scaling and dataset is also important. Those results are in line with our intuition. Multi ShapeNet has simpler scenes than FlyingThings3D. The first layers of the Cloud Embedder extract features that are more descriptive of scenes of Multi ShapeNet, than of FlyingThings3D. Thus, the contribution to the loss of the first layers may be higher for the first dataset than for the latter.

The Multi-Scale Triplet loss is the main loss used in our setup. We leave for future work to explore the benefits of this loss in setups where the triplet loss is used. Appendix~\ref{apx:mstl} shows additional experiments that we performed while exploring it. By using multiple feature vectors we allow our latent space to be more information-rich. Which, in turn, improves the feedback given to the Flow Extractor.


\section{Why Not the Nearest Neighbors?}
\label{sec:nnexps}

We have argued against the use of nearest neighbor-based distances to supervise the training of the flow extractor. We back those arguments with the experiments of this section. We aim to learn scene flow with the self-supervision of the nearest neighbor loss, similarly as proposed by \cite{SelfSupervisedFlow}. We then use our sandbox to identify the failure modes of the training setup. We proceed to conjecturing whether or not a KNN-based loss is an adequate choice for the task.

We use a similar setup as the one proposed by \cite{SelfSupervisedFlow}. The main difference is we do not use a pretrained model. For each point in $C_1$ we calculate the mean distance of the k-nearest neighbor in $C_2$. The average of distances is used as loss to train the Flow Extractor. We use Cycle Consistency loss as defined by Equation~\ref{eq:ccl} to introduce the local geometric consistency. For more details of the setup we refer the reader to Appendix~\ref{apx:expsetup}. The loss calculation is given by:
\begin{align}
    \mathcal{L} &= \gamma_{\knn} \mathcal{L}_{\knn} + \gamma_{\cc} \mathcal{L}_{\cc}, \\
     \mathcal{L}_{\knn}(\hat{C}, C, k) &= \frac{1}{|\hat{C}|} \sum_{\vec{p} \in \hat{C}} \frac{1}{k} \sum_{\vec{p}_n \in \KNN(\vec{p}, C)} || \vec{p} - \vec{p}_n ||_2
\end{align}
where $\KNN(\vec{p}, C)$ is a function that returns the k closest points to $\vec{p}$ in $C$. $|\hat{C}|$ is the number of points in $\hat{C}$.

Table~\ref{tab:knnloss} shows the results of training a Segmenter and a FlowNet3D using the aforementioned training scheme. The models perform similarly across datasets, all zEPEs are just bellow 1. 

\begin{table}
    \centering
    \begin{tabular}{llllll}
        \toprule
        Flow Extractor & Dataset        & EPE    & zEPE   & Acc 01 & Acc 005 \\ \midrule
        Segmenter      & Single ShapeNet & 0.3786 & 0.9455 & 2.98\% & 0.43\%  \\
        Segmenter      & Multi ShapeNet  & 0.8444 & 0.9595  & 0.34\% & 0.03\% \\
        Segmenter      & FlyingThings3D  & 0.7559 & 0.9953 & 4.82\% & 0.98\%  \\ \midrule
        FlowNet3D      & Single ShapeNet & 0.3911 & 0.9768 & 2.76\% & 0.3\%  \\
        FlowNet3D      & Multi ShapeNet  & 0.8242 & 0.9337  & 0.44\% & 0.03\% \\
        FlowNet3D      & FlyingThings3D  & 0.7538 & 0.9925 & 5.1\%  & 1.3\%  \\ \bottomrule
    \end{tabular}
    \caption{Flow Extractors trained with nearest neighbor loss and cycle consistency loss. }
    \label{tab:knnloss}
\end{table}

\begin{figure}
    \begin{subfigure}{0.5\textwidth}
        \includegraphics[width=0.99\linewidth]{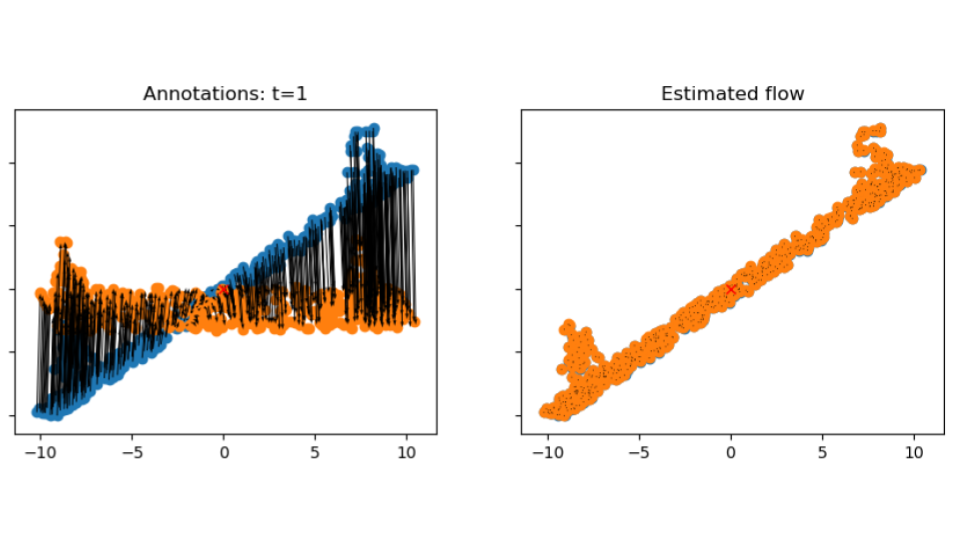}
        \caption{2D view of a table from Single ShapeNet. Flow Extractor used is FlowNet3D. Blue Points belong to frame 1, orange points belong to frame 2 and the black arrows are the flow vectors.}
        \label{img:knnresuts}
    \end{subfigure}
    \begin{subfigure}{0.5\textwidth}
        \includegraphics[width = .99\linewidth]{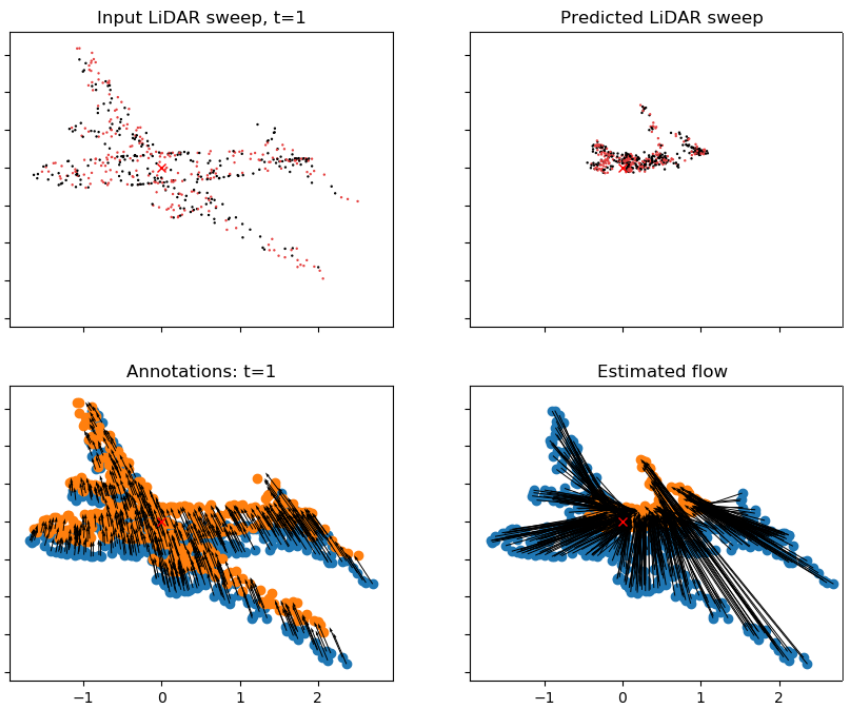}
        \caption{2D projection of an Airplane from Single ShapeNet. Flow Extractor cluster points in order to minimize the nearest neighbor distance. The top left image shows the target point cloud, the top right image shows the predicted one. The flow collapses most points around the centroid of the object. On bottom left the ground truth flow vectors are shown, on the bottom right the estimated flow vectors are displayed.}
        \label{img:knncolapse}
    \end{subfigure}
    \caption{Qualitative results from FlowNet3D on Single ShapeNet.}
    \label{img:knnproblems}
\end{figure}

We show results for all three synthetic dataset for the sake of completion. However, experiments with Single ShapeNet would have sufficed. We identify the first failure mode in Figure~\ref{img:knnresuts}, the models failed to capture motion. By comparing Tables \ref{tab:knnloss} and \ref{tab:nlbaselines} we conclude the issue does not lay on the flow models themselves, but rather in the training setup. 

It could be argued that the Cycle Consistency is spoiling training by favoring zero flow vectors. However, the results reported on Table~\ref{tab:knnloss} were the best we found in our experiments. For instance, by lowering the contribution of the Cycle Consistency term, $\gamma_{cc}$ the failure to capture motion gives place to another failure mode. The flow vectors are locally inconsistent and geometric coherence is lost. Figure~\ref{img:knncolapse} shows an example when we set $\gamma_{cc} = 0.0$. The Flow Extractor collapses the points into clusters that minimize the nearest neighbor loss without learning any flow estimation. 

There are options other than KNN for measuring distances between sets of points. Still, they are all nearest neighbor based distances. For instance, the chamfer and the, hungarian distance are possibilities \cite{PointPWCNet, hungariandistance} expected to decrease the clustering problem observed in Figure~\ref{img:knncolapse} because they perform bilateral assignments of points. We restrain from experimenting further with those distances, for we consider they are fundamentally ill-suited for giving feedback to the flow extractor. 

\begin{figure}
    \centering
    \includegraphics[width = 1\linewidth]{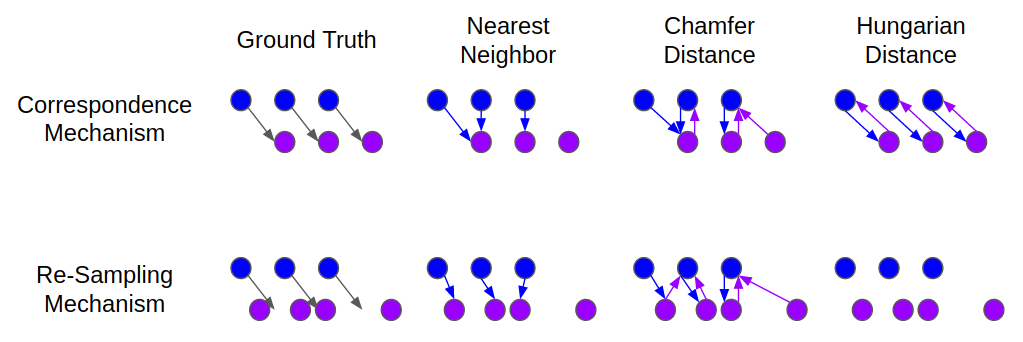}
    \caption{Illustrative example of the different set distances. Blue point belong to frame $t$, purple points belong to frame $t+1$. Gray arrows represent the ground truth flow. The blue arrows represent the pointwise distances with respect to $t$ and the purple the point wise distances with respect to $t+1$. The set distances are the sum of the arrows in each image. The image makes explicit how the set distances are ill-suited when the Re-sampling mechanism is used. Nearest neighbor tends to underestimate the total motion in both mechanisms. Chamfer distance has the same underestimation problem, however the bilateral assignment regularizes it to some extent. Hungarian distance may be ideal for the Correspondence mechanism, but it is not defined for sets with different number of points, as it the case for the Re-sampling mechanism.}
    \label{img:setdists}
\end{figure}

The fundamental issue with those distances is they assume correspondence between points. In both cases, the distance is zero when every point is correctly assigned to its new location. In particular, hungarian distance is not defined for sets with different number of points. The Re-sampling mechanism makes those distances at best a noisy estimate of motion, as illustrated in Figure~\ref{img:setdists}. Thus, even if we manage to bring the loss to zero, we still expect poor flow vector estimations. In order to improve the metrics, we have to improve how we measure distance between two point clouds.

\section{Correspondence vs Re-sampling mechanisms}
\label{subsec:assumption}

Troughout the experiments of this chapter, we used the Re-sampling mechanism as method for gathering data. In this section, we will show the impact of assuming the Correspondence or Re-sampling Mechanism during evaluation. We compare how models perform when using data from the different assumptions on synthetic and on real data.


We have used the publicly available code-base and set of hyper-parameters \cite{hplflow_repo, pointpwc_repo} to reproduce the work of \cite{HPLFlowNet, PointPWCNet}. Two models, HPLFlowNet \cite{HPLFlowNet} and PointPWC-net \cite{PointPWCNet}, were trained using FlyingThings3D with the Re-sampling Mechanism as defined in Chapter~\ref{sec:datasets}. Another two models were trained using FlyingThings3D with the Correspondence Mechanism as defined in the authors code-base \cite{hplflow_repo, pointpwc_repo}. All models were evaluated on both mechanisms.

Table~\ref{tab:samplevscorrespondence} shows the impact the different mechanisms make on performance. We report only the EPE for the sake of compactness. We include the performance of FlowNet3D \cite{FlowNet3D} for comparison. For all models, the metrics are lower when the Correspondence Mechanism is used for evaluation. The results are in line with what was reported by the authors \cite{HPLFlowNet, PointPWCNet}.

\begin{table}
    \centering
    \begin{tabular}{l|l|l|l|l}
        \toprule
        & \multicolumn{2}{l|}{Correspondence Training} & \multicolumn{2}{l}{Re-sampling Training} \\ \midrule
        Architecture &
        \begin{tabular}[l]{@{}l@{}}Correspondence\\ Eval\end{tabular} &
        \begin{tabular}[l]{@{}l@{}}Re-sampling\\ Eval\end{tabular} &
        \begin{tabular}[l]{@{}l@{}}Correspondence\\ Eval\end{tabular} &
        \begin{tabular}[l]{@{}l@{}}Re-sampling\\ Eval\end{tabular} \\ \midrule
        HPLFlow     & 0.0948 & 0.3997 & 0.1965 & 0.2598 \\
        PointPWC-net   & 0.0575 & 0.4747 & 0.1701 & 0.2644 \\
        \begin{tabular}[l]{@{}l@{}}PointPWC-net \\ (self-sup)\end{tabular}  & 0.0965 & 0.4888 & 0.3455 & 0.5502 \\
        FlowNet3D   & 0.1136 * & --     & --     & 0.1694 ** \\ \bottomrule      
    \end{tabular} 
    \caption{EPE comparison on the impact of the mechanism used for training in the evaluation performance, FlyingThings3D used as dataset. All results are on supervised training, except for Point-PWC (self-sup) where we use the same self-supervised losses as the authors. * taken from \cite{HPLFlowNet}, ** taken from \cite{FlowNet3D}.}
    \label{tab:samplevscorrespondence}
\end{table}


The Correspondence Mechanism simplifies the task of scene flow significantly. From Table~\ref{tab:samplevscorrespondence} we observe two reasons for such. All models report lower metrics on the correspondence evaluation regardless of the training dataset. Additionally, the re-sampling training generalizes to the correspondence evaluation, however, the opposite does not happen. Those findings are not surprising, the Correspondence Mechanism artificially carries over occlusions independent of the motion of the objects. Thus the consecutive frames are rather similar. The Re-sampling Mechanisms, on the other hand, allow for entire objects to disappear in-between frames. Which limits the information available for flow information.

We notice a curious outlier in Table~\ref{tab:samplevscorrespondence}. PointPWC-net self-supervised performs better on the re-sampling evaluation when trained using correspondence training rather than the re-sampling training. One possible reason is when using the correspondence training the loss is less noisy, as there are no occlusions nor objects leaving the field of view. This may be beneficial for the model to learn better features for flow estimation. We would like to point out, however, that the performance is one order of magnitude worse than when correspondence evaluation is used.

The goal of self-supervised Scene Flow estimation is to make use of real-world data. The results of this section added experimental weight to our understanding that the Re-sampling Mechanism is more representative of real data than the Correspondence counterpart.


\section{Reflection} 

The Section~\ref{sec:bait} was used as bait to capture the interest of the reader. The performance of our setup on real data is not the most important contribution of this work. We speculated about possible reasons that could justify the difference in the performance of the various methods. The fact is, we lacked the tools to perform any solid analysis. All we knew was, the supervised method works far better than any self-supervised method. Yet, it required annotated data. It is a trade-off between performance and flexibility. The supervised method requires a finetuning stage, which is not possible for datasets where flow targets are not available. However, this insight alone is somewhat underwhelming. It would be interesting to dissect our method and find out which are the critical test cases.

If the bait was successful, the reader went on to explore the Scene Flow Sandbox with a curious mindset. The individual datasets are not the most insightful part of the Sandbox, the failure modes are. The failure modes are a combination of data and flow models. We identified five failure modes that were useful to explore later sections. From failing to capture any motion from a dynamic scene to confusion caused by occluded objects.

The first and most obvious one was the failure to capture any motion. We expect a flow model to capture motion, but in our experiments, we ran into a large number of models and training setups that were ill-suited to perform scene flow estimation. We presented the example of the Segmenter trained with supervision, it was sufficient to learn flow from Single ShapeNet, but it was too simplistic to perform flow estimation from FlyingThings3D. Overcoming this failure mode may be less trivial than one would suspect.

The next failure mode is rather subtle. Underestimation of vector norms was seen when the Segmenter was trained with supervision and when FlowNet3D was trained via Adversarial Metric Learning on Multi ShapeNet. Even though the model grasps the motion of different objects, it is rather conservative in its estimations. Which points to low-quality features. A successful self-supervised setup must give clear feedback to the flow model so it learns high-quality features. 

The following two failure modes reflect the level of geometric understanding a flow model can abstract. Flow vectors of a neighborhood are expected to be similar in norm and direction. That is because the geometry of an object is not expected to suffer major deformations in between two frames. On the other hand, different objects have independent motion. Either failure mode suggests the model does not abstract the different objects in a scene and their trajectories. FlowNet3D failed to grasp this motion independence on KITTI without ground, both when trained supervised and with Adversarial Metric. 

The fifth and last failure mode we pointed out is confusion caused by partially visible scenes. Models must learn not only how the different objects move, but also how that changes the visibility of the scene. A model may guess the direction of an occluded object by gathering clues on the motion of objects around it. This was the most challenging failure mode we encountered. We expected the Flow Extractor and the Cloud Embedder to learn to deal with partially observable scenes. Our experiments did not reflect this expectation. As previously explained. We conjecture the Cloud Embedder identifies occlusions to differentiate between the real and the predicted point cloud. The best the Flow Extractor can do is merge two objects so to simulate occluded objects, which in turn leads to incorrect flow estimation. 

The sandbox allowed us to study the quality of the flow estimations of different models and training setups. The identified failure modes allowed us to bridge quantitative and qualitative results. Failure modes identified on simple datasets are often present on more complex datasets. This sandbox allowed us to experiment with various variations of our proposed setup and find the most promising faster than if we were experimenting directly with real data.

The proposed training setup was motivated primarily on devising a loss function that was suitable for scene flow estimation. Whereas \cite{SelfSupervisedFlow, PointPWCNet} proposed the use of nearest neighbor based losses in their self-supervised setups, we aimed to learn a distance metric between point clouds. More precisely, a distance between the predicted and the target point clouds. The flow models trained by the learned distance metric were able to perform flow estimation on fully visible scenes. Geometric consistencies and coherent movement were learned. However, precise motion estimation is still left for future work. 

Multi-Scale Triplet loss and Cycle Consistency are important components of our method. Our ablation study shows the positive aspects of using multiple activations from the Cloud Embedder instead of only the last one. It increases the amount of information that can be encoded in the latent space for comparing point clouds and possibly makes the training of the Cloud Embedder simpler. We also showed how the Cycle Consistency loss improved metrics and the stability of the training.

To properly compare what previous work has done, we coined the Correspondence and Re-sampling Mechanisms and made explicit their implications for the task of scene flow. We justify the need for the concept through the experiments of Section~\ref{subsec:assumption}. The difference in performance shows that the Correspondence Mechanism is not representative of our application. It is clear that if we want to work with real data from LiDAR scans or stereo videos we should stick our evaluation method using the Re-sampling Mechanism.

%% file: src/conclusion.tex
\chapter{Conclusion}
\label{sec:conclusion}

Self-supervised scene flow estimation is the task of estimating 3D flow vectors to individual points of a dynamic scene without having access to the flow targets at training time. We addressed the task on two ends. We proposed a Scene Flow Sandbox and the Adversarial Metric Learning setup for self-supervised training of flow models.  

We found that previous state-of-the-art models \cite{PointPWCNet, HPLFlowNet} assumed the Correspondence Mechanism in their evaluation. This makes the scene flow task simpler than it originally is. The flow targets are implicitly present in the evaluation data. We showed that the performance of those models degrade when the Re-sampling Mechanism is used in the evaluation. We argued the latter mechanism is the most representative of data collected by LiDAR sensors and stereo cameras.

The Scene Flow Sandbox is a benchmark of datasets designed to study individual aspects of flow estimation in progressive order of complexity. Using the sandbox and non-leaning baselines we identified five failure modes. Those failure modes were key to bridge the quantitative and qualitative results. 

We introduced the Adversarial Metric Learning for self-supervised flow estimation. The Cloud Embedder replaced nearest neighbor based losses and learns a metric that effectively differentiates between estimated and target point clouds. The Flow Extractor was trained to improve its flow estimations. We used our Scene Flow Sandbox to draw insights on the limitations of nearest neighbor based losses on the limitations of our proposed setup. We found that our setup is able to train a Flow Extractor to learn coarse flow estimation. It keeps motion coherence and preserves local geometries. The main open challenge is to improve its predictions in partially observable scenes.

\section{Future outlook}

The ultimate goal is to perform scene flow estimation on scenes collected via LiDAR scans and stereo videos. We focused on the setup for self-supervision. Yet, research that focuses on more effective models for the setup may be of great benefit. The most pressing issue is to improve the Flow Extractor predictions in partially visible scenes. Additionally, we focused on improving the informational capacity of the Cloud Embedder, but optimizing the use of the latent vectors may introduce a major leap in performance. More broadly, future work can profit from our proposed benchmark. It can help understand the limitations of different solutions and provide insights of how to overcome them.

%% file: src/appendix.tex
\chapter{Algorithms}

\label{apx:algos}

\begin{algorithm}[H]

\SetAlgoLined
  \KwInput{index}
 // Load point cloud of an random object of ShapeNet
 $nframe = 2$ \tcp*{number of frames}
 $npoints = 1024$ \tcp*{number of points on output}
 
 $obj$ = ShapeNet.load(index) \;
 // $obj.shape = [3, npoints]$ \\
 
 // Initialize parameters for initial transform \;
 $\theta_0 = [ U(-\pi, \pi), U(-\pi, \pi), U(-\pi, \pi)]^T $ \tcp*{initial angle}
 $\mu_0 = [ U(5, 6), U(5, 6), U(5, 6)]^T$ \tcp*{initial stretch factor}
 $T_0 =  [U(-1, 1), U(-1, 1), U(-1, 1)]^T$ \tcp*{initial position}
 $M_0 = \text{MakeRotationMatrix}(\theta_0, \mu_0) $ \\
 $obj_0 = M_0 \cdot obj + T_0$ \\ 
 
 // Initialize parameters for frame transform \;
 $\theta = [ U(-\frac{\pi}{15}, \frac{\pi}{15}), U(-\frac{\pi}{15}, \frac{\pi}{15}), U(-\frac{\pi}{15}, \frac{\pi}{15})]^T $ \\
 $\mu = [ U(0.9, 1.1), U(0.9, 1.1), U(0.9, 1.1)]^T$ \\
 $T =  [U(-0.25, 0.25), U(-0.25, 0.25), U(-0.25, 0.25)]^T$ \\
 $M = \text{MakeRotationMatrix}(\theta, \mu) $ \\
 
 $t = 1$ \\
 \While{$t \leq nframes$}{
  $obj_t = M \cdot obj_{t-1} + T$ \\
  $flow_{t-1} = obj_{t-1} - obj_{t}$ \\
  Draw $npoints$ random indices $idx$ without replacement from $obj_{t-1}$ \\
  $pc_{t-1} = obj_{t-1}[idx]$ \tcp*{select a random sample}
  $f_{t-1} = flow_{t-1}[idx]$ \tcp*{flow vectors of $pc_{t-1}$}
 }
 \KwOutput{$pc, f$}
 
 \caption{Single ShapeNet}
 \label{alg:shapenet}
\end{algorithm}

\newpage

There are minor implementation and computational differences between the algorithms \ref{alg:shapenet} and \ref{alg:multishapenet}. Those modifications are in place to allow for the batch computation of matrix transformations. 

\begin{algorithm}[H]

\SetAlgoLined
  \KwInput{index}
 // Load point cloud of an random object of ShapeNet
 $nframe = 2$ \tcp*{number of frames}
 $npoints = 1024$ \tcp*{number of points on output}
 $nobjs =  U(2, 20)$ \tcp*{number of objects}
 
 $n = 1$ \\
 \While{$n \leq nobjs$}{
    $idx$ is randomly drawn;
    $obj_n$ = ShapeNet.load(idx) \;
    $n += 1$
 }
 // $obj.shape = [nbojs, npoints, 3]$ \\
 
 // Initialize parameters for initial transform \;
 $\theta_0 = [ U(-\pi, \pi), U(-\pi, \pi), U(-\pi, \pi)]^T $ \tcp*{initial angle}
 $\mu_0 = [ U(3, 8), U(3, 8), U(3, 8)]^T$ \tcp*{initial stretch factor}
 $T_0 =  [U(-10, 10), U(-10, 10), U(-10, 10)]^T$ \tcp*{initial position}
 // $theta_0.shape, \mu_0.shape, T_0.shape = [nobjs, 3]$
 $M_0 = \text{MakeRotationMatrix}(\theta_0, \mu_0)^T $ \\
 // $M_0.shape = [nobjs, 3, 3]$
 $obj_0 = obj \cdot M_0 + T_0$ \\ 
 
 // Initialize parameters for frame transform \;
 $\theta = [ U(-\frac{\pi}{15}, \frac{\pi}{15}), U(-\frac{\pi}{15}, \frac{\pi}{15}), U(-\frac{\pi}{15}, \frac{\pi}{15})]^T $ \\
 $\mu = [ U(0.9, 1.1), U(0.9, 1.1), U(0.9, 1.1)]^T$ \\
 $T =  [U(-0.25, 0.25), U(-0.25, 0.25), U(-0.25, 0.25)]^T$ \\
 $M = \text{MakeRotationMatrix}(\theta, \mu)^T $ \\
 
 $t = 1$ \\
 \While{$t \leq nframes$}{
  $obj_t = obj_{t-1} \cdot M + T$ \\
  $flow_{t-1} = obj_{t-1} - obj_{t}$ \\
  Draw $npoints$ random indices $idx$ without replacement from $obj_{t-1}$ \\
  $pc_{t-1} = obj_{t-1}[idx]$ \tcp*{select a random sample}
  $f_{t-1} = flow_{t-1}[idx]$ \tcp*{flow vectors of $pc_{t-1}$}
  $t += 1$
 }
 \KwOutput{$pc, f$}
 
 \caption{Multi ShapeNet}
 \label{alg:multishapenet}
\end{algorithm}

\newpage

The original FlyingThings3D \cite{flyingthings3d} was designed to emulate stereo cameras. The authors use a rendering engine to render diverse objects with independent movement in a scene. Each scene was made of ten frames. For each frame the following data is available:

\begin{itemize}
    \item RGB image of left and right camera;
    \item Disparity map, which can be converted to a depth map via the focal length that is provided;
    \item Forward and backward optical flow;
    \item Forward and backward disparity change;
\end{itemize}

The authors of \cite{FlowNet3D} post-processed the original dataset into point clouds. We use their post-processed dataset. The authors of \cite{FlowNet3D} did not publish the algorithm used for the post-processing, however, code was made available. We summarize it in the algorithm \ref{alg:ft3d}. 

\newpage

\begin{algorithm}[H]

\SetAlgoLined
  \KwInput{index}
 $npoints = 1024$ \tcp*{number of points on output}
 $focalLength = 1050$ \\
 $imgWidth, imgHeight = 540, 960$ \\
 Load $img_0$, $img_1$, $disparityMap_0$, $disparityMap_1$, $disparityChange_0$, $opticalFlow_0$ from files \\
 // The data is in grid format $[imgWidth, imgHeight, C]$ \\
 // $C$ is the number of channels (3 for RGB, 1 for disparity, 2 for optical flow) \\

 $depthMap_0 = \frac{focalLength}{disparityMap_0} $ \\
 $xyGrid = \text{meshgrid}(\text{range}(imgWidth), \text{range}(imgHeight)$ \\
 $xyzMap_0[x] = (xyGrid[x] - (\frac{imgWidth}{2}))  * \frac{focalLength}{depthMap_0} $ \\
 $xyzMap_0[y] = (xyGrid[y] - (\frac{imgHeight}{2}))  * \frac{focalLength}{depthMap_0} $ \\
 $xyzMap_0[z] = depthMap_0$ \\
 
 $depthChangeMap = \frac{focalLength}{disparityMap_0}$ \\
 $xyNextGrid = xyGrid + opticalFlow_0$ \\
 $xyzChangeMap[x] = (xyNextGrid[x] - (\frac{imgWidth}{2}))  * \frac{focalLength}{depthChangeMap} $ \\
 $xyzChangeMap[y] = (xyGrid[y] - (\frac{imgHeight}{2}))  * \frac{focalLength}{depthChangeMap} $ \\
 $xyzChangeMap[z] = depthChangeMap $ \\
 $flowMap_0 = xyzChangeMap - xyzChangeMap $ \\
 
 // Draw points for the first frame
 Draw random indices $idx$ from $img_0$ \\
 $rgb_0 = img_0[idx]$ \\
 $pos_0 = xyzMap_0[idx] $ \\
 $flow_0 = flowMap_0[idx] $ \\
 
 $depthMap_1 = \frac{focalLength}{disparityMap_1} $ \\
 $xyzMap_1[x] = (xyGrid[x] - (\frac{imgWidth}{2}))  * \frac{focalLength}{depthMap_1} $ \\
 $xyzMap_1[y] = (xyGrid[y] - (\frac{imgHeight}{2}))  * \frac{focalLength}{depthMap_1} $ \\
 $xyzMap_1[z] = depthMap_1$ \\
 
 Draw random indices $idx$ from $img_1$ \\
 $rgb_1 = img_1[idx]$ \\
 $pos_1 = xyzMap_1[idx] $ \\
 
 \KwOutput{$pos_0, pos_1, rgb_0, rgb_1, flow_0$}
 
 \caption{Point Cloud FlyingThings3D}
 \label{alg:ft3d}
\end{algorithm}

\chapter{Experimental Setup}
\label{apx:expsetup}

In this section we describe the experimental setup of individual experiments. 

Tables \ref{tab:sumresults} and \ref{tab:samplevscorrespondence}: we used the same setup as made publicly available by the authors of \cite{HPLFlowNet, PointPWCNet} on \cite{hplflow_repo, pointpwc_repo}.

Table \ref{tab:nlbaselines}: non-leaning baselines have no training stage. The KNN estimator uses $k = 1$. We set $\gamma_{\knn} = 1.0$ and $\gamma_{\cc} = 1.0$. 

Table \ref{tab:nlbaselines}, supervised baselines:

\begin{itemize}
    \item Training Type: Supervised;
    \item Flow Extractor training details: \begin{itemize}
        \item Optimizer: Adam;
        \item Learning Rate: $lr = 0.0001$;
        \item Betas: $\beta_1 = 0.0, \beta_2 = 0.99$;
        \item Schedule: $0.5 * lr$ every 5 epochs the evalaluation EPE does not decrease;
        \item Weight Decay: $4 \cdot 10^{-4}$;
        \item Loss: l2.
    \end{itemize}
    \item Batch Size: $64$ for ShapeNet and Multi ShapeNet and $32$ for FlyingThings3D;
    \item Stopping criteria: Training for at least 50 epochs, train concludes if no improvement on the EPE is seen for consecutive 20 epochs or 500 epochs is reached.
\end{itemize}

Table \ref{tab:supervisedkitti}: finetunning perfomed with Kitti, used the following setup:

\begin{itemize}
    \item Training Type: Supervised, warm start;
    \item Flow Extractor pretrained on FlyingThings3D. Training details: \begin{itemize}
        \item Optimizer: Adam;
        \item Learning Rate: $lr = 0.0001$;
        \item Betas: $\beta_1 = 0.0, \beta_2 = 0.99$;
        \item Schedule: $0.5 * lr$ every 5 epochs the evalaluation EPE does not decrease;
        \item Weight Decay: $4 \cdot 10^{-4}$;
        \item Loss: l2.
    \end{itemize}
    \item Batch Size: $16$;
    \item Stopping criteria: Training for at least 50 epochs, train concludes if no improvement on the EPE is seen for consecutive 20 epochs or 500 epochs is reached.
\end{itemize}

Table \ref{tab:knnloss}:

\begin{itemize}
    \item Training Type: Self-Supervised via KNN;
    \item Flow Extractor training details: \begin{itemize}
        \item Optimizer: Adam;
        \item Learning Rate: $lr = 0.0001$;
        \item Betas: $\beta_1 = 0.0, \beta_2 = 0.99$;
        \item Schedule: $0.5 * lr$ every 5 epochs the evalaluation EPE does not decrease;
        \item Weight Decay: $4 \cdot 10^{-4}$;
        \item Loss: $\mathcal{L} = \mathcal{L}_{knn} + \mathcal{L}_{cc}$
    \end{itemize}
    \item Batch Size: Batch Size: $64$ for ShapeNet and Multi ShapeNet and $32$ for FlyingThings3D;
    \item Stopping criteria: Training for at least 50 epochs, train concludes if no improvement on the EPE is seen for consecutive 20 epochs or 500 epochs is reached.
\end{itemize}

Tables \ref{tab:advresults}, \ref{tab:ablation_cycleloss}, \ref{tab:mstloss}:

\begin{itemize}
    \item Training Type: Self-Supervised via Adversarial Metric Learning;
    \item Flow Extractor training details: \begin{itemize}
        \item Optimizer: Adam;
        \item Learning Rate: $lr = 0.0005$;
        \item Betas: $\beta_1 = 0.0, \beta_2 = 0.99$;
        \item Schedule: $0.75 * lr$ every 10 epochs the evalaluation EPE does not decrease;
        \item Weight Decay: $4 \cdot 10^{-4}$;
        \item Loss: $\mathcal{L} = \mathcal{L}_{FE}$, defined by equation \ref{eq:fe};
        \item Cycle Consistency uses cosine similarity and l2 distance unless explicitly stated.
    \end{itemize}
    \item Cloud Embedder training details: \begin{itemize}
        \item FlowNet3D and PointPWC-net were trained with PointNet++ Cloud Embedder described on Table~\ref{tab:CEpointnetpp}. The Segmenter was trained with PointNet Cloud Embedder described on Table~\ref{tab:CEpointnet}.
        \item Optimizer: Adam;
        \item Learning Rate: $lr = 0.00005$;
        \item Betas: $\beta_1 = 0.0, \beta_2 = 0.99$;
        \item Schedule: $0.75 * lr$ every 10 epochs the evalaluation EPE does not decrease;
        \item Weight Decay: $4 \cdot 10^{-4}$;
        \item Loss: $\mathcal{L} = \mathcal{L}_{CE}$, defined by equation \ref{eq:ce}.
    \end{itemize}
    \item Batch Size: Batch Size: $64$ for ShapeNet and Multi ShapeNet and $32$ for FlyingThings3D;
    \item Stopping criteria: Training for at least 50 epochs, train concludes if no improvement on the EPE is seen for consecutive 40 epochs or 500 epochs is reached.
\end{itemize}

\begin{table}
\centering
\begin{tabular}{@{}ll@{}}
\toprule
\multicolumn{2}{l}{PointNet Cloud Embedder}                                                                                                        \\ \midrule
Module & Details                                                                                                                                   \\ \midrule
PointNet Feature Extractor 1 & \begin{tabular}[c]{@{}l@{}}layers: [64, 128, 256]\\ activation: ReLU\\ normalization: BatchNorm\\ output: point wise features\end{tabular} \\ \midrule
PointNet Feature Extractor 2 & \begin{tabular}[c]{@{}l@{}}layers: [1024, 1024, 512]\\  activation: ReLU\\ normalization: BatchNorm\\ output: global feature\end{tabular}   \\ \midrule
MLP 1  & \begin{tabular}[c]{@{}l@{}}layers: [1024, 2048, 2048]\\ activation: ReLU\\  normalization: BatchNorm\\ output: global feature\end{tabular} \\ \midrule
MLP 2  & \begin{tabular}[c]{@{}l@{}}layers: [2048, 4096]\\ activation: ReLU\\ normalization: BatchNorm\\ output: global feature\end{tabular}       \\ \bottomrule
\end{tabular}
\caption{The architecture used for the Cloud Embedder. The output of every module was used on the Multi-Scale Triplet loss. Pointwise features are converted to global feature vectors by max pooling.}
\label{tab:CEpointnet}
\end{table}

\begin{table}
\centering
\begin{tabular}{@{}ll@{}}
\toprule
\multicolumn{2}{l}{PointNet++ Cloud Embedder}                                                                                                  \\ \midrule
Module & Details                                                                                                                             \\ \midrule
PointNet++ Set Abstraction 1 &
  \begin{tabular}[c]{@{}l@{}}number of samples: 16\\ layers: [32, 64, 64]\\ activation: ReLU\\ normalization: BatchNorm\\ output: point wise features\end{tabular} \\ \midrule
PointNet++ Set Abstraction 2 &
  \begin{tabular}[c]{@{}l@{}}number of samples: 16\\ layers: [64, 128, 256]\\ activation: ReLU\\ normalization: BatchNorm\\ output: point wise features\end{tabular} \\ \midrule
PointNet++ Set Abstraction 3 &
  \begin{tabular}[c]{@{}l@{}}number of samples: 8\\ layers: [256, 512, 1024]\\ activation: ReLU\\ normalization: BatchNorm\\ output: point wise features\end{tabular} \\ \midrule
MLP 1  & \begin{tabular}[c]{@{}l@{}}layers: [128, 256]\\ activation: ReLU\\ normalization: BatchNorm\\ output: global feature\end{tabular}   \\ \midrule
MLP 2  & \begin{tabular}[c]{@{}l@{}}layers: [512, 1024]\\ activation: ReLU\\ normalization: BatchNorm\\ output: global feature\end{tabular}  \\ \midrule
MLP3   & \begin{tabular}[c]{@{}l@{}}layers: [2048, 4096]\\ activation: ReLU\\ normalization: BatchNorm\\ output: global feature\end{tabular} \\ \bottomrule
\end{tabular}
\caption{The architecture used for the Cloud Embedder. The output of every module was used on the Multi-Scale Triplet loss. Point wise features are converted to global feature vectors by mean pooling.}
\label{tab:CEpointnetpp}
\end{table}


\chapter{Multi-Scale Triplet Loss}
\label{apx:mstl}

The first promising results came from experiments for flow extraction on point clouds. The Multi-Scale Triplet Loss helps with the stability of the training and boosts performance. The table \ref{tab:deeploss} gives a taste of those experiments. We also performed classification and clustering experiments.

\begin{table}
    \centering
    \begin{tabular}{llllll}
        \toprule
        Loss type                       & EPE    & Acc 01  & Acc 005 & wall time & epochs \\ \midrule
        \textit{Shallow Triplet L2}     & 0.1622 & 26.57\% & 6.37\%  & 4h        & 250 \\
        \textit{Multi-Scale Triplet L2} & 0.1574 & 32.09\% & 7.17\% & 2h50min    & 179 \\ \bottomrule
    \end{tabular}
    \caption{Difference in performance between Multi-Scale and shallow losses. The metrics are all better for the Multi-Scale loss. Even though the Multi-Scale loss adds some overhead to the calculations, the wall time is still lower.}
    \label{tab:deeploss}
\end{table}

Setup: One network learns to map MNIST digits to a 2D space. We call it the Cluster Network, it is trained using Hard Triplet Mining. The classification is done via a linear Classifier, the classifier maps the 2D vector to a one-hot encoded vector with 10 classes. The two models are trained together. For evaluation it is considered:

\begin{itemize}
    \item Triplet Accuracy: the positive samples are mapped closer to the anchor than to randomly chosen negative samples.
    \item Classifier Accuracy: The Classifier trains for one epoch with the Cluster Net frozen. I measure the overall accuracy of the classifier.
    \item Time to 50\% Classifier Accuracy: The number of epochs needed to reach 50\% of classification accuracy, that is to proxy how the first stage of training evolves.
\end{itemize}

The motivation behind using a linear classifier is as follows. The Cluster Net will ideally map the digits to be linearly separable. If that is the case, the Classifier should get near 100\% accuracy. However, if the Cluster Net is not distinctive enough, it will make overlapping clusters. The Classifier Accuracy will then drop.

The first experiment compares the same Cluster Network trained using deep and shallow loss at difference capacities. The Network has 7 layers, 4 conv and 3 fully connected. The capacity is defined as the layer with the most number of hidden units, all others are proportionally adjusted, a minimum of 4 hidden units per layer is ensured. 

Table \ref{tab:capacity} shows how using Multi-Scale triplet loss helps to make training robust to networks with different capacity.

\begin{table}
    \centering
    \begin{tabular}{lllll}
        \toprule
        Loss type & Capacity & Time to 50\% & Classifier acc & Triplet Acc \\ \midrule
        Shallow  & 512 & 35 & 73\% & 96\% \\
        Multi-Scale (6) & 512 & 20 & 75\% & 96\% \\
        Shallow  & 32  & 30 & 71\% & 93\% \\
        Multi-Scale (6) & 32  & 25 & 70\% & 95\% \\
        Shallow  & 4   & -- & 44\% & 86\% \\
        Multi-Scale (6) & 4   & 45 & 64\% & 90\% \\ \bottomrule
    \end{tabular}
    \caption{Difference in performance between Multi-Scale and shallow losses. Multi-Scale loss makes the training much more robust to reduced capacity of the Cluster Net.}
    \label{tab:capacity}
\end{table}

Another experiment shows the impact of the number of layers used for Multi-Scale triplet loss. The number in the brackets indicates the number of layers receiving triplet feedback. 6 means every layer receives a feedback, 3 means every other layer and so on. For networks with limited capacity, dense feedback seems useful too keep training progressing. If the network has enough capacity, then it seems to be beneficial to allow for a few not-linear transformations in between feedback points.

\begin{table}
    \centering
    \begin{tabular}{lllll}
        \toprule
        Loss type       & Capacity  & Time to 50\%  & Classifier acc & Triplet Acc \\ \midrule
        Shallow         & 512       & 35            & 73\%           & 96\% \\
        Shallow         & 4         & --            & 44\%           & 86\% \\
        Multi-Scale (6) & 512       & 20            & 75\%           & 96\% \\
        Multi-Scale (6) & 4         & 45            & 64\%           & 90\% \\
        Multi-Scale (3) & 512       & 30            & 75\%           & 96\% \\
        Multi-Scale (3) & 4         & 50            & 61\%           & 89\% \\
        Multi-Scale (2) & 512       & 40            & 82\%           & 96\% \\
        Multi-Scale (2) & 4         & --            & 46\%           & 84\% \\
        Multi-Scale (1) & 512       & 30            & 79\%           & 96\% \\
        Multi-Scale (1) & 4         & --            & 44\%           & 83\% \\ \bottomrule
    \end{tabular}
    \caption{Difference in performance between Multi-Scale and shallow losses. Multi-Scale loss makes the training much more robust to reduced capacity of the Cluster Net.}
    \label{tab:skips}
\end{table}